%% file: main.tex
\documentclass[11pt]{article}
\input{math_commands.tex}

\usepackage{amsmath,amssymb,amsfonts,amsthm,bbm}
\usepackage[skins]{tcolorbox}
\usepackage{listings,lstautogobble}
\tcbuselibrary{skins,breakable,xparse,raster}

\usepackage{times}
\RequirePackage[colorlinks,citecolor=blue,urlcolor=blue,linkcolor=blue,linktocpage=true]{hyperref}
\usepackage{url}
\usepackage{graphicx}
\usepackage{multicol}
\usepackage{multirow}
\usepackage{array}
\usepackage{adjustbox}
\usepackage{tabularx}
\usepackage{hhline}
\usepackage{wrapfig,lipsum}
\usepackage{soul}
\usepackage{booktabs}
\usepackage{verbatim}
\usepackage{setspace}
\usepackage{graphicx}
\usepackage{subcaption}
\usepackage[shortlabels]{enumitem}
\usepackage{bbm}
\usepackage[top=1in, bottom=1in, left=1in, right=1in]{geometry}
\usepackage[colorinlistoftodos]{todonotes}
\usepackage{longtable}
\usepackage{rotating}

\newcommand{\head}{\mathrm{head}}
\newcommand{\Ave}{\mathrm{Ave}}
\newcommand{\Std}{\mathrm{Std}}

\newcommand{\SA}{\mathrm{MSA}}
\newcommand{\QK}{\mathrm{QK}}
\newcommand{\OV}{\mathrm{OV}}

\definecolor{python-green}{rgb}{0.3333333333333333, 0.6588235294117647, 0.40784313725490196}

\usepackage[toc,page,header]{appendix}
\usepackage{minitoc}

\title{Out-of-distribution generalization via composition: a lens through induction heads in Transformers}
\author{Jiajun Song, Zhuoyan Xu, Yiqiao Zhong}
\author{Jiajun Song\thanks{National Key Laboratory of General Artificial Intelligence, BIGAI,
Beijing 100080, China,
\texttt{songjiajun@bigai.ai}} \and Zhuoyan Xu\thanks{Department of Statistics, University of Wisconsin--Madison, Madison, WI, 53706, USA. Emails: \texttt{zhuoyan.xu@wisc.edu}, \texttt{yiqiao.zhong@wisc.edu}} \and Yiqiao Zhong\footnotemark[2]}
\date{}

\begin{document}

\maketitle
\begin{abstract}

Large language models (LLMs) such as GPT-4 sometimes appear to be creative, solving novel tasks often with a few demonstrations in the prompt. These tasks require the models to generalize on distributions different from those from training data---which is known as out-of-distribution (OOD) generalization. Despite the tremendous success of LLMs, how they approach OOD generalization remains an open and underexplored question. We examine OOD generalization in settings where instances are generated according to hidden rules, including in-context learning with symbolic reasoning. Models are required to infer the hidden rules behind input prompts without any fine-tuning. We empirically examined the training dynamics of Transformers on a synthetic example and conducted extensive experiments on a variety of pretrained LLMs, focusing on a type of components known as induction heads. We found that OOD generalization and composition are tied together---models can learn rules by composing two self-attention layers, thereby achieving OOD generalization. Furthermore, a shared latent subspace in the embedding (or feature) space acts as a bridge for composition by aligning early layers and later layers, which we refer to as the  \textit{common bridge representation hypothesis}.

\end{abstract}
\doparttoc 
\faketableofcontents 
\part{} 

\input{src/intro}

\input{src/background}

\input{src/synthetic}

\input{src/llm}

\input{src/related}

\input{src/limit}

\input{src/availability_acknowledge}

\bibliography{refs}
\bibliographystyle{plain}

\newpage

\appendix
\addcontentsline{toc}{section}{Appendix} 
\part{Appendix} 
\parttoc 

\newpage 

\input{src/appendix/notations}

\newpage 

\section{Experiment details for the synthetic example}\label{sec:append-synthetic}

\input{src/appendix/synthetic-model}
\input{src/appendix/computing-err}
\input{src/appendix/one-layer-tf}
\input{src/appendix/larger_models}
\input{src/appendix/model_variants}
\input{src/appendix/measurements}

\newpage

\section{Additional results for the synthetic example}\label{sec:append-add-synthetic}
\input{src/appendix/larger_models_results}
\input{src/appendix/synthetic}
\input{src/appendix/synthetic-rank}
\input{src/appendix/synthetic-rope}
\input{src/appendix/spike}

\newpage

\section{Model and task details for experiments with LLMs}\label{sec:append-llm}

\input{src/appendix/implementation}

\newpage

\section{Experiment details for the Common bridge representation hypothesis}

\input{src/appendix/IH}
\input{src/appendix/induction_head_details}
\input{src/appendix/cbrh}

\input{src/appendix/histogram}
\input{src/appendix/shuffle-histogram}
\input{src/appendix/projection-curve}

\newpage

\section{Additional results for the Common bridge representation hypothesis}\label{sec:append-cbrh}
\input{src/appendix/add-cbrh}

\newpage

\section{Experiment details for the Scaling Experiment}
\input{src/appendix/scaling}

\end{document}

%% file: math_commands.tex
\usepackage{amsmath,amssymb,amsfonts,amsthm, bm}

\def\eqref#1{equation~\ref{#1}}

\def\1{\bm{1}}

\def\ve{{\bm{e}}}

\def\vh{{\bm{h}}}

\def\vs{{\bm{s}}}

\def\vx{{\bm{x}}}
\def\vy{{\bm{y}}}

\def\mA{{\bm{A}}}

\def\mI{{\bm{I}}}

\def\mM{{\bm{M}}}

\def\mO{{\bm{O}}}

\def\mR{{\bm{R}}}

\def\mU{{\bm{U}}}
\def\mV{{\bm{V}}}
\def\mW{{\bm{W}}}
\def\mX{{\bm{X}}}

\def\mZ{{\bm{Z}}}

\DeclareMathAlphabet{\mathsfit}{\encodingdefault}{\sfdefault}{m}{sl}
\SetMathAlphabet{\mathsfit}{bold}{\encodingdefault}{\sfdefault}{bx}{n}

\def\gA{{\mathcal{A}}}

\def\gI{{\mathcal{I}}}

\def\gP{{\mathcal{P}}}

\def\gS{{\mathcal{S}}}

\newcommand{\R}{\mathbb{R}}

\DeclareMathOperator*{\argmax}{arg\,max}


%% file: src/intro.tex
\section{Introduction}\label{sec:intro}

Large language models (LLMs) are sometimes able to solve complex tasks that appear novel or require reasoning abilities. The appearance of creativity in task-solving has sparked recent discussions about artificial general intelligence \cite{bubeck2023sparks, liang2023holistic, Donoho2024Data, bengio2024machine}.

The classical notion of statistical generalization does not fully explain the progress observed in LLMs. Traditionally, both training instances and test instances are drawn from the same distribution, and it is generally not expected that a model will generalize well on a different test distribution without explicit domain adaptation involving model updates.

The recent success of LLMs suggests a different story: if test data involve compositional structures, LLMs can generalize across different distributions with just a few demonstrations in the prompt (few-shot learning) or even without any demonstrations (zero-shot learning) \cite{brown2020language} without updating the model parameters. Indeed, the apparent ability of models to infer rules from the context of a prompt---known as in-context learning (ICL)---is a hallmark of LLMs~\cite{dong2022survey}. Moreover, a growing body of literature on chain-of-thought prompting explicitly exploits the compositional structures of reasoning tasks. For example, phrases like ``let's think step by step'' are prepended to input prompts to elicit reasoning \cite{reynolds2021prompt, kojima2022large}, and prompts with intermediate steps are used to improve accuracy on mathematical tasks \cite{wei2022emergent}. 

The ability to generalize on distributions different from the training distribution---known as out-of-distribution (OOD) generalization---is well documented empirically, ranging from mathematical tasks that involve arithmetic or algebraic structures \cite{kazemnejad2023the, zhang2022unveiling, abbe2023generalization}, to language reasoning problems requiring multistep inference \cite{wu-etal-2024-reasoning, ood-language-23}. Yet, less is known about \textit{when} a model achieves OOD generalization and \textit{how} it solves a compositional task with OOD data.

Our empirical investigations are motivated by the pioneering work on the \textit{induction head} \cite{elhage2021mathematical, inductionhead22}, which is a component within the Transformer architecture.
Our main contributions are as follows.


    First, on the synthetic task of copying sequences of arbitrary patterns, a 2-layer Transformer exhibits an abrupt emergence of subspace matching that accompanies OOD generalization between two Transformer layers, a phenomenon that echoes \textit{emergent abilities} \cite{wei2022emergent}.
    
    Second, on language reasoning tasks where LLMs infer the meanings of planted symbols, including examples of in-context learning, OOD generalization requires a similar compositional structure. Extensive experiments on LLMs suggest the presence of a latent subspace for compositions in multilayer and multihead models, which we propose as the \textit{common bridge representation hypothesis}.

\subsection{An exemplar: copying}\label{sec:copying}

\begin{figure*}[t]
\centering
\includegraphics[width=0.8\textwidth]{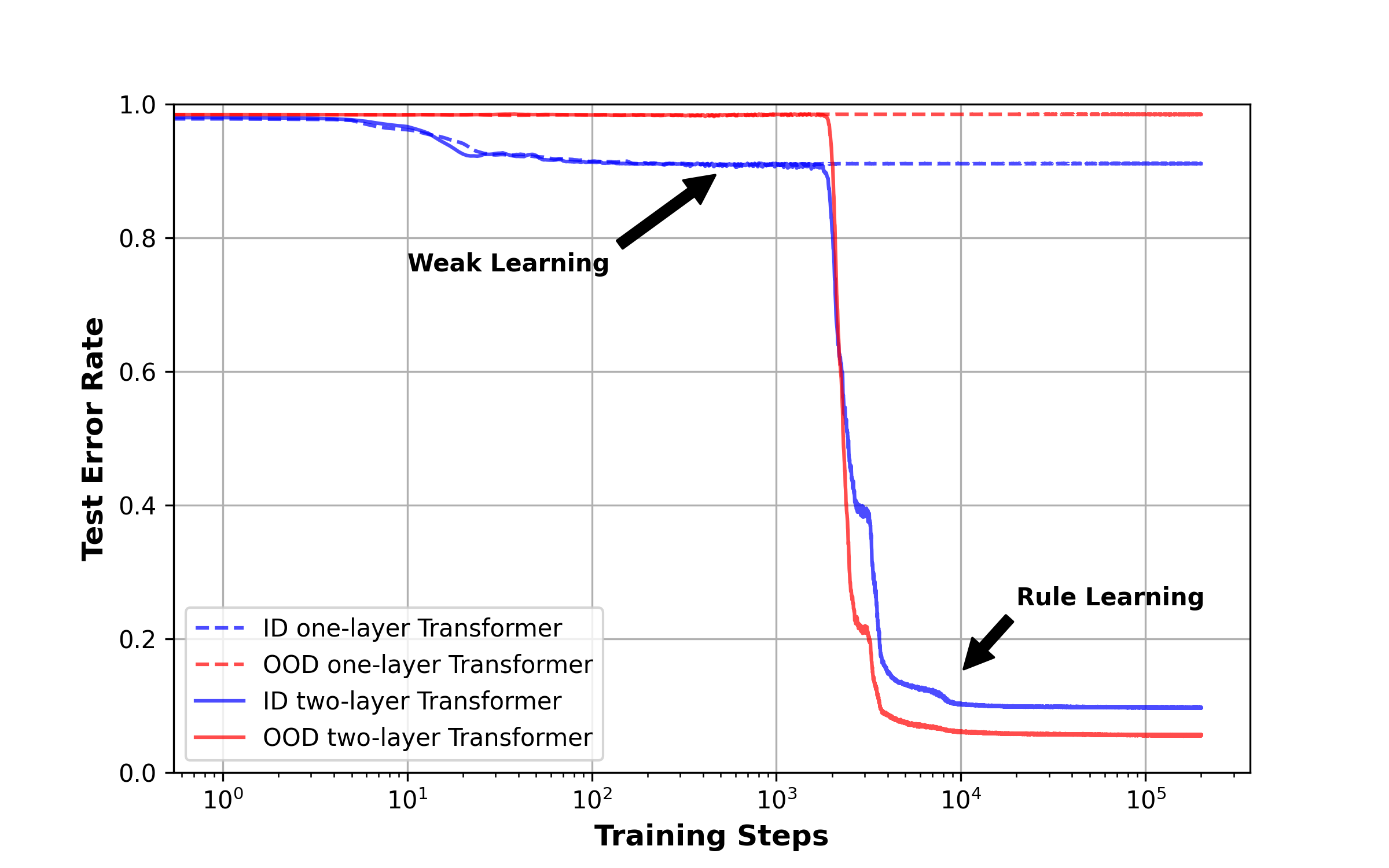}
\caption{Training a two-layer Transformer (TF) and a one-layer TF for copying task using \textbf{fresh samples} of the format $(*, \vs^\#, *, \vs^\#, *)$. The models are evaluated on an in-distribution ({\color{blue}ID}) test dataset and an out-of-distribution ({\color{red}OOD}) test dataset. \textbf{Weak learning phase}: the models rely on simple statistics of ID data and fail to generalize on OOD data; \textbf{Rule-learning phase:} two-layer TF learns the rule of copying from ID data and generalize well on ID/OOD data.}\label{fig:main}
\end{figure*}

\textit{Copying} is a simple yet nontrivial task that exemplifies OOD generalization. Briefly speaking, given a sequence that contains several consecutive tokens such as $[A]$, $[B]$, $[C]$, a model predicts the next token as $[C]$ upon receiving $[A], [B]$: 
\begin{align*}
    &\ldots [A], [B], [C] \ldots [A], [B] \qquad \xrightarrow{\text{next-token prediction}}   \\
    &
  \ldots [A], [B], [C]\ldots [A], [B],{\color{blue} [C]}
 \end{align*}
Formally, consider a sequence of tokens $\mathbf{s} = (s_1,\ldots,s_T)$ where each $s_t$ is in a discrete vocabulary set $\mathcal{A}$. Suppose that a segment of length $L$ is repeated in this sequence: $\vs_{T_0:(T_0+L-1)} = \vs_{(T-L+1):T}$ where $L < (T-T_0)/2+1$. Upon receiving the sequence $\vs_{1:(T-1)}$, copying requires a model to output\footnote{Since Transformers output probabilities at test time, a predicted token $s$ is the one that maximizes the conditional probability mass function $p_T(s| \vs_{1:{(T-1)}})$.} token $s_T$.

Copying represents a primitive form of abstract reasoning. 
While humans can code the copying rule into a model, 
classical statistical models have difficulty learning this rule purely based on instances of such sequences. For instance, hidden Markov models and $n$-gram models \cite{brants2007large} require estimating a large transition matrix or high-order conditional probability, which scales exponentially in $L$.

As a simple experiment, we fix $|\mathcal{A}| = 64$ and consider a power law distribution $\mathcal{P}$ on $\mathcal{A}$. We sample each sequence $\vs$ of length $T_{\max} = 64$ independently by planting repeated segments in a ``noisy background'':
\begin{enumerate}
    \item Sample $L$ uniformly from $\{10,11,\ldots,19\}$, sample $s^\#_t$ from $\mathcal{P}$ independently for $t=1,\ldots,L$, and form a segment $\vs^\# = (s_1^\#,\ldots, s^\#_L)$; 
    \item Denote $r_L = T_{\max} - 2L$. Sample two integers uniformly from $\{1,2,\ldots,r_L\}$ and denote the smaller/larger ones by $T_0, T_1$;
    \item Form a sequence $(*, \vs^\#, *, \vs^\#, *)$ where $*$ is filled by random tokens of lengths $T_0, T_1-T_0, r_L - T_1$ respectively drawn from $\mathcal{P}$ independently.
\end{enumerate}
We train a 2-layer attention-only Transformer on batches of fresh samples, namely each training step uses independently drawn sequences. Model architecture and training follow the standard practice; see Appendix~\ref{sec:append-synthetic} for details. We report both in-distribution (ID) test errors and OOD test errors by computing the average token-wise prediction accuracy based on 
the second segment. Here the OOD error is evaluated on sequences of a similar format but $\mathcal{P}$ is replaced by the uniform distribution $\mathcal{P}_{\mathrm{ood}}$, and $L$ is replaced by $L_{\mathrm{ood}} = 25$. As a comparison, we train a 1-layer attention-only Transformer using the same data.
 
\subsection{Compositional structure is integral to OOD generalization}

In Figure~\ref{fig:main}, we observe that the 2-layer Transformer experiences two phases. (i) In the weak learning phase, the model learns the marginal token distribution $\mathcal{P}$ and predicts the most probable token irrespective of the structure of the entire sequence $\vs$. 
This results in a decrease of the ID error but not the OOD error.
(ii) In the rule-learning phase, the model learns the copying rule and generalize reasonably well on both ID and OOD data. In contrast, one-layer Transformer only achieves weak learning. Note that the OOD error is smaller after training because the longer repetition segment means a larger signal strength. 

Learning the copying rule requires the model to generalize on OOD instances in two aspects.
\begin{enumerate}
    \item Generalize to a larger repetition length $L_{\mathrm{ood}}$ (aka length generalization).
    \item Generalize to a different token distribution $\mathcal{P}_{\mathrm{ood}}$.
\end{enumerate}
Our analysis of training dynamics later suggests that the two layers of the Transformer play complementary roles: one layer specializes in processing positional information and another in token information. Composing the two layers yields OOD generalization.

%% file: src/background.tex
\section{Background}

We briefly introduce Transformers, model interpretability, and OOD generalization. For illustrative purposes, we focus on core model modules and selectively present key concepts, deferring further discussion of related work to Section~\ref{sec:related} and implementation details to Appendix.

\subsection{Transformers}
Transformers are standard neural network architectures for processing text and other sequential data. While many variants of Transformers have been developed, their core components align with the original Transformer model proposed by \cite{vaswani2017attention}. The first step of processing is tokenization, where a string $\vs$ is converted into tokens. Each token is an element of a vocabulary $\mathcal{A}$, typically representing words, subwords, special symbols, and similar units. Then, these tokens are mapped to  numerical vectors known as token embeddings. In the original formulation \cite{vaswani2017attention}, positional information is encoded by a set of positional embedding vectors, and the sum of token and positional embeddings produce 
input vectors $\vx_1,\ldots,\vx_T \in \R^d$ to a Transformer.

\paragraph{Transformer architecture.} The input vectors are processed by multiple Transformer layers, which yield hidden states $\vx_1^{(\ell)},\ldots, \vx_T^{(\ell)} \in \R^d$ after each layer $\ell$. Let $\mX = [\vx_1^{(\ell)},\ldots,\vx_T^{(\ell)}]^\top \in \R^{T \times d}$ be the input vectors or the hidden states at a layer $\ell$ ($\ell=0$ means the input vectors).
The computations at each layer follow the generic formulas: for each $\ell=1,2,\ldots,L$,
\begin{align}
\begin{split}\label{eq:update}
    & \mX \longleftarrow \mX + \mathrm{MSA}(\mX; \mW^{(\ell)})\\
    & \mX \longleftarrow \mX + \mathrm{FFN}(\mX;\widetilde \mW^{(\ell)})
\end{split}
\end{align}
where $\mW^{(\ell)}$ and $\widetilde \mW^{(\ell)}$ contain all trainable parameters. Here, a crucial component is the multihead self-attention (MSA), which is a mapping $\SA(\mX; \mW): \R^{T \times d} \to \R^{T \times d}$ given by 


\begin{equation}\label{def:msa}
    \SA(\mX; \mW) = \sum_{j=1}^H 
    \overbrace{\mathrm{Softmax} \hspace{1mm} \big( \mX \mW_{\QK,j} \mX^\top \big)}^{\text{attention matrix}}
    \underbrace{\phantom{\big(} \mX \mW_{\OV,j}^\top \phantom{\big)}}_{\substack{\text{OV circuit writes and}\\\text{adds info to stream}}}
\end{equation}

where $\mW_{\QK,j}, \mW_{\OV,j} \in \R^{d \times d}$, and $\mW$ denotes the collection of all such matrices. The softmax operation applies to each row of $\mX \mW_{\QK,j}  \mX^\top$ and yields attention weights that are positive and sum to $1$ by definition. The feedforward network $\mathrm{FFN}$ applies to each column of $\mX$ individually, and one common form reads $\mathrm{FFN}(\vx_t; \widetilde \mW) = \widetilde \mW_1 \mathrm{ReLU}(\widetilde \mW_2 \vx_t)$ for each position $t$.

At the final layer $L$, there is a simple classification layer $\mathrm{Softmax}(\mW^{(\mathrm{out})} \vh_T^{(L)})$ where $\mW^{(\mathrm{out})} \in \R^{|\mathcal{A}| \times d}$, to convert the last-position hidden states $\vh_T^{(L)}$ to a probability vector, which is used to predict the next token. During pretraining, the entire model (weights in all layers and token embeddings) is trained with backpropagation.



\paragraph{Historical development.}
Word embedding \cite{mikolov2013linguistic, Mikolov2013EfficientEO, pennington2014glove}, the precursor to Transformers, aims to represent words as numerical vectors, similar to the input vectors described earlier. To find contextualized representation of text strings, recurrent neural networks (RNNs) and long short-term memory networks (LSTMs) were initially employed, but it was soon found that attention mechanism \cite{vinyals2015show, bahdanau2014neural} handles long-range dependencies more effectively, laying the foundation for Transformers. As the scale of model sizes and pretraining corpora increased, Transformers started to exhibit surprising capabilities. Notably, they can generalize on many downstream tasks with crafted prompts without even updating the model parameters \cite{reynolds2021prompt}. For example, in in-context learning, a string of instructions or examples are added as a prefix to the test string, and the model seemingly ``learns'' from the context and generalizes on new tasks.

\subsection{Interpreting Transformers} 
How do Transformers process and represent text data? While Transformers are designed and often used as black-box models, recent analyses offer partial insights into this question. The hidden states $\vx_t$ are believed to encode the semantic meaning of input texts as a linear combination of base concept vectors, e.g., \cite{yun2021transformer}
\begin{align}
    \text{apple} &= 0.09 \text{``dessert''} + 0.11 \text{``organism''} + 0.16 \text{``fruit''} \notag \\
    &+ 0.22\text{``mobile\&IT''} + 0.42\text{``other''},
\end{align}
which means that the hidden states representing the word ``apple'' is decomposed into a linear combination of vectors of interpretable base concepts/features. A further refined decomposition of ``other'' vector may reveal syntactic features, positional features, and other contextual features. 

This approach of interpreting Transformers is strongly supported by empirical evidence. In word embedding, word vectors show interpretable factor structure, such as ``$\text{King} - \text{Man} + \text{Woman} = \text{Queen}$'' \cite{mikolov2013linguistic}; the FFN component is found to memorize concepts \cite{geva2020transformer}; and a dictionary learning algorithm applied to hidden states identifies a substantial number of features on large-scale language models \cite{templeton2024scaling, gao2024scaling}. These insights have inspired practice in model editing \cite{li2024inference} and task arithmetic \cite{ilharco2022editing}.

Below we present a list of key concepts often used for interpreting Transformers.
\begin{itemize}
\item Linear representation hypothesis (LRH) and feature superposition. The LRH posits that concepts are encoded as linear subspaces within the embedding or hidden states space. Feature superposition posits that hidden states are sparse linear combinations of base concept vectors from a potentially large dictionary.
\item Attention matrix. It is the output of $\mathrm{Softmax}$ operation in each of $H$ attention heads in Eqn.~\ref{def:msa}. An attention matrix $\mA \in \R^{ T \times T}$ indicates how an attention head aggregates information from past tokens via the weighted sum, where the weights correspond to the entries of $\mA$.
\item Residual stream. It is the output of the identity map (namely a copy of $\mX$) in Eqn.~\ref{eq:update}, which is updated by the outputs of both $\mathrm{MSA}$ and $\mathrm{FFN}$ in each Transformer layer.
\item QK circuit. It is a component of an attention head that computes $\mZ = \mX \mW_{\QK,j}  \mX^\top \in \R^{T \times T}$. A QK circuit provides a similarity measure between two hidden states, and $\mZ$ decides at each token how much an attention head ``reads'' and ``processes'' information from past tokens. 
\item OV circuit. It is a component of an attention head that computes $\mA \mX \mW_{\OV,j}^\top$, whose output is added to the residual stream. The OV circuit decides how information is ``written'' to the stream.
\item Previous-token head (PTH). It refers to a type of attention heads in a Transformer that attends to the previous token, namely $A_{t,t-1} \approx 1$, for a typical input text.
\item Induction head (IH). It refers to a type of attention heads that attends to the token next to an identical token on an input string with repetition patterns.
\end{itemize}
In the literature, PTHs and IHs are usually qualitative characterization of attention patterns as they are input-dependent. In this paper, we identify PTHs and IHs from all attention heads by calculating and ranking scores based on desired attention patterns, cf.~Eqn.~\ref{eq:PTH-score}--\ref{eq:IH-score}.

\subsection{OOD generalization} 
Since GPT-3 and GPT-4, many research communities have showed significant interest in understanding why trained models generalize well on novel tasks. Broadly speaking, OOD generalization refers to the generalization behavior of models when training data have a different distribution compared with the test data. Unlike classical statistics topics such as extrapolation, distribution shift, and domain adaptation, pretrained LLMs can solve compositional tasks and seemingly learn from novel context. While mathematical characterization of OOD data for natural languages remains elusive, recent experimental investigations have offered insights. We highlight several representative approaches for generating OOD data.
\begin{enumerate}
    \item Changing the length or the distribution of task strings. This is particularly useful for experiments with synthetic data, where the distribution of task strings can be easily modified \cite{lee2023teaching, kazemnejad2024impact, wang2024grokked}.
    \item Replacing parts of a text string with counterfactual statements or unnatural symbols. It is used for probing LLMs to test reliability or reveal the internal mechanism \cite{rong2021extrapolating, wu2023reasoning, pan-etal-2023-context, halawi2024overthinking}.
    \item Constructing compositional tasks based on deductive rules. It is used for testing models on complex reasoning tasks to examine whether models have learned the latent rules \cite{ood-language-23}.
\end{enumerate}
However, prior studies either fail to probe the inner workings of Transformer models or lack systematic measurements across many models. Compared with existing papers, we provide in-depth measurements to probe model behavior during training dynamics, and comprehensive experiments across many pretrained models, ranging from small Transformer such as GPT-2 (124M parameters) to larger LLMs such as Llama-3 (70B parameters). Moreover, our findings about subspace matching are novel, especially in the context of 
compositionality of LLMs.

%% file: src/synthetic.tex
\begin{figure*}[t]
\centering
\includegraphics[width=0.95\textwidth]{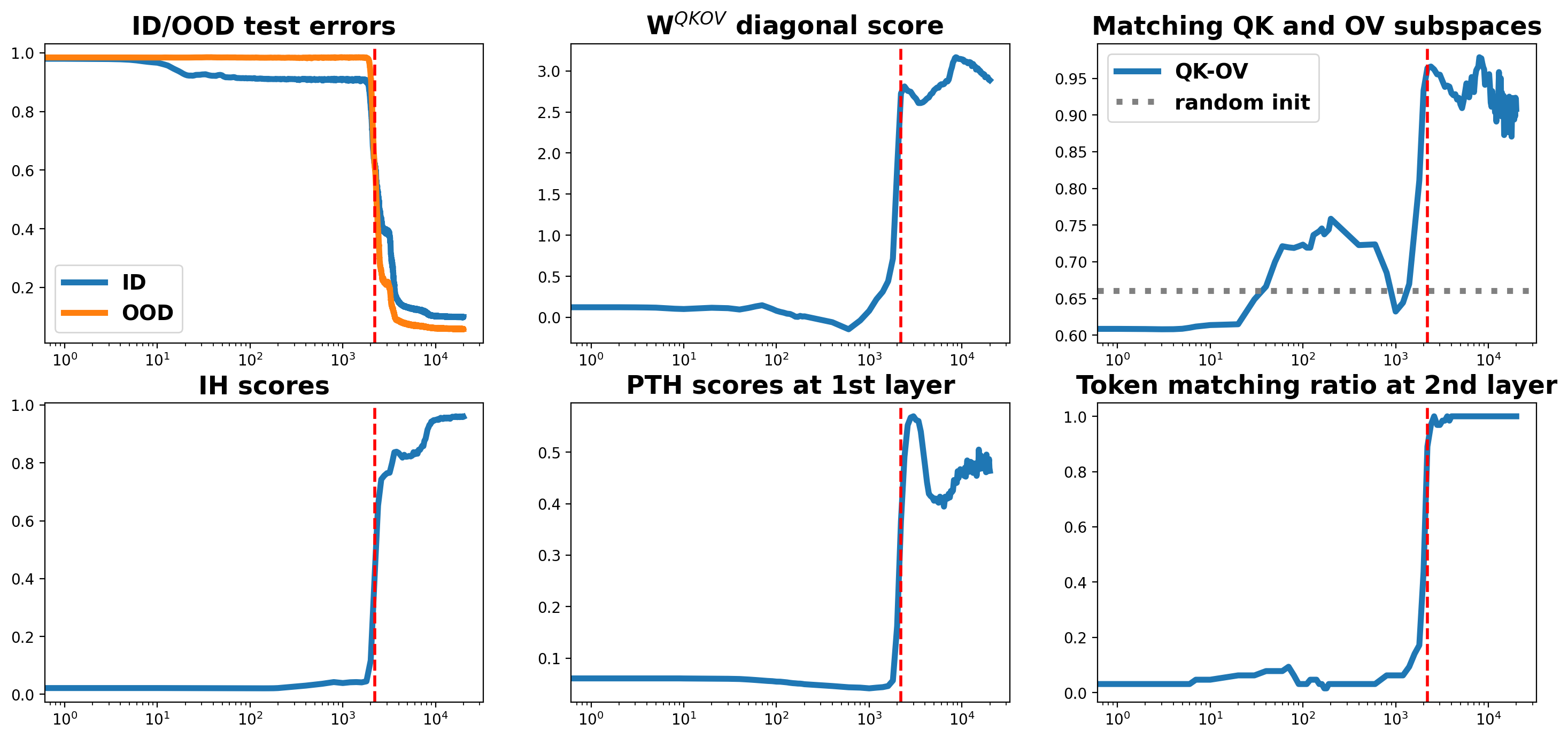}
\caption{Measuring training dynamics for 2-layer 1-head Transformers on the copying synthetic data. \textbf{First row:} Test errors drop abruptly as structural matching occurs. Middle and right plots measure the matching between 1st-layer output circuit (OV) and 2nd-layer input circuit (QK). \textbf{Second row:} Model achieves OOD generalization by learning to compose two functionally distinct components (position matching vs.~token matching). Left plot shows the formation of the IH on OOD data. Middle shows PTH scores on completely random tokens devoid of token info. Right shows token matching stripped of positional info. }\label{fig:progress}
\end{figure*}

\section{Dissecting sharp transition in synthetic example}\label{sec:synthetic}

As in Section~\ref{sec:copying}, we train a \textbf{minimally working} Transformer for the copying task as a clean synthetic example, though we find similar results on larger models. In particular, the model has 2 self-attention layers with a single head and no MLPs. The architecture and training are standard, including residual connection, LayerNorm, RoPE, dropout, autoregressive training with the AdamW optimizer, and weight decay. In the Appendix, we provide a list of notations, experimental details, and comparison with model variants. 

\subsection{Progress measures} \label{sec:prog}

Under the circuits perspective in Eq.~\ref{def:msa}, each of the $H$ attention head consists of a ``reading'' QK circuit, a ``writing'' OV circuit, and the softmax operation. Since $H=1$ in our experiment, our goal is to study how the 1st-layer attention head interact with the 2nd-layer head. To this end, we define several measurements.


\begin{figure*}[t]
\centering
\includegraphics[width=0.95\textwidth]{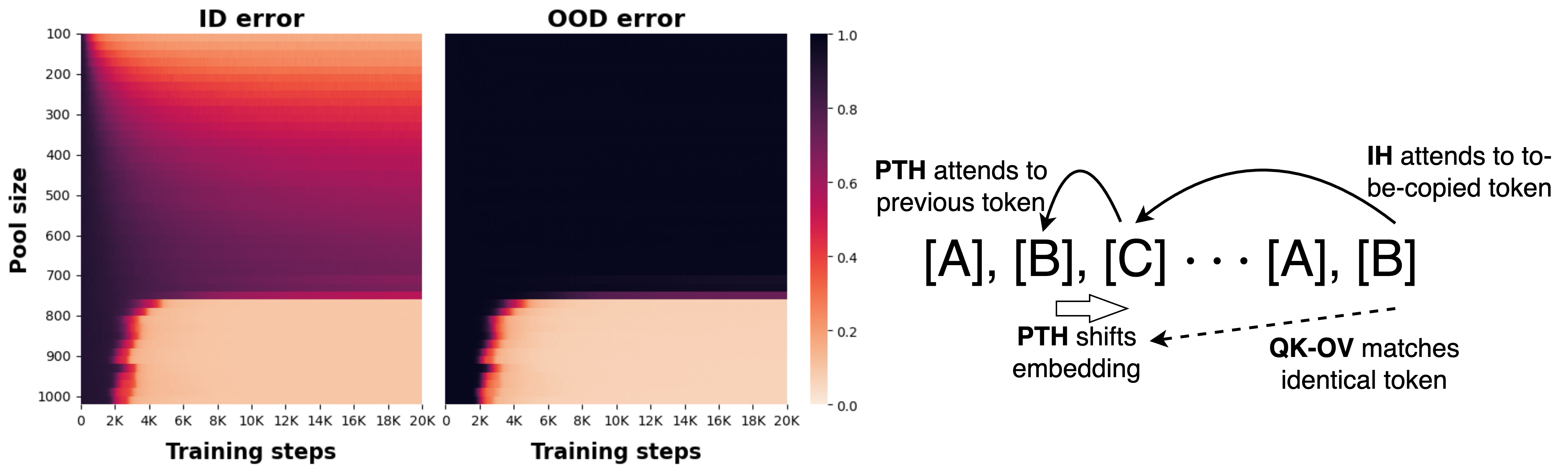}
\caption{\textbf{Left:} \textbf{Memorization vs.~generalization: more varied repetition patterns help models to learn the copying rule}. When the set $\mathcal{S}$ of allowable $\vs^\#$ during training has a size smaller than 740, models fail to learn the rule and generalize OOD under 20K steps, yet they can still memorize the patterns if the pool size is small. 
\textbf{Right:} 
\textbf{Composition of two layers expresses the rule of copying}. 1st-layer head shifts the embedding at [B] to [C]. Through the QK-OV circuits, the embedding at [C] then matches the last token [B] in the 2nd-layer attention calculation, resulting in attention to [C] and completes the copying task.
}\label{fig:heatmap}
\end{figure*}

\begin{enumerate}
    \item [1.] Diagonal score. For 1st-layer $\mW_{\OV}$ and 2nd-layer $\mW_{\QK}$, we calculate $\mW^{\mathrm{QKOV}} = \mW_{\QK} \mW_{\OV} \in \R^{d \times d}$ and define $z=z(\mW_{\QK}, \mW_{\OV})$ as
    \begin{equation*}
        z = \frac{\Ave\Big( (W_{ii}^{\mathrm{QKOV}})_{i\le d} \Big) - \Ave\Big( (W_{ij}^{\mathrm{QKOV}})_{i, j \le d} \Big)}{\Std\Big( (W_{ij}^{\mathrm{QKOV}})_{i, j \le d}\Big)} \, 
    \end{equation*}
    where $\Ave, \Std$ mean taking average and standard deviation respectively. 
\end{enumerate}
This score measures the strength of diagonal entries. 
Roughly speaking, $z$ is interpreted as the signal-to-noise ratio under the ``$\lambda \mI_d $ + noise'' assumption on $\mW^{\mathrm{QKOV}}$.

\begin{enumerate}
    \item [2.] Subspace matching score. Fix rank parameter $r = 10$. We calculate the top-$r$ right singular vectors $\mU \in \R^{d \times r}$ of 2nd-layer $\mW_{\QK}$ and top-$r$ left singular vectors $\mV \in \R^{d \times r}$ of 1st-layer $\mW_{\OV}$. The column linear span $\mathcal{P}_{\QK} :=\mathrm{span}(\mU)$ (or similarly for $\mV$) represents the principal subspace for reading (or writing) information.
    We calculate a generalized cosine similarity between the two subspaces.
    \begin{equation}\label{def:sim}
        \mathrm{sim}\big(\mathcal{P}_{\QK}, \mathcal{P}_{\OV}\big) := \sigma_{\max} \big( \mU^\top \mV  \big)\, ,
    \end{equation}
    where $\sigma_{\max}$ denotes the largest singular value.
\end{enumerate}
This score is equivalent to the regular cosine similarity between two optimally chosen unit vectors within the two subspaces. Results are analogous under a similar average similarity between $\mathcal{P}_{\QK}, \mathcal{P}_{\OV}$.

In Figure~\ref{fig:progress} (top row), we discovered simultaneous sharp transitions in generalization and also in the two scores. To investigate why the structural matching yields OOD generalization, we consider more measurements. An \textit{attention matrix} $\mA = (A_{t,t'})_{t, t' \le T}$ is the output of the softmax in Eq.~\ref{def:msa}. The weight $A_{t,t'} \in [0,1]$ represents the coefficient (aka attention) from position $t$ to $t'$ in a sequence and satisfies $\sum_{t'} A_{t,t'}=1$.
\begin{enumerate}
    \item [3.] PTH (previous-token head) and IH (induction-head) attention scores. Given an attention head and $N$ input sequences, we first calculate the attention matrices $(\mA_i)_{i \le N}$. The PTH attention score measures the average attention to the previous adjacent token, and IH attention score measures the average attention to the to-be-copied token:
\begin{align}
    &\mathrm{score}^{\mathrm{PTH}} = \mathrm{Ave}_{i \le N} \Big( \frac{1}{T-1} \sum_{T \ge t \ge 2} (\mA_i)_{t, t-1} \Big) \, , \label{eq:PTH-score} \\   
    &\mathrm{score}^{\mathrm{IH}} = \mathrm{Ave}_{i \le N} \Big(\frac{1}{|\gI_i|} \sum_{t \in \gI_i} (\mA_i)_{t, t-L_i+1} \Big)\,. \label{eq:IH-score}
\end{align}
where $\gI_i$ is the index set of repeating tokens (second segment $\vs^\#$), and $L_i$ is the relative distance between two segments. 
\end{enumerate}
To emphasize the OOD performance, we calculate IH scores by using the OOD dataset of format $(*, \vs^\#, *, \vs^\#, *)$ described in Section~\ref{sec:copying}. Moreover, PTH scores are based on random tokens uniformly drawn from the vocabulary as total removal of token-specific information.


\begin{enumerate}
    \item [4.] Token matching ratio. Let $\ve_j \in \R^d$ be the token embedding for the $j$-th token in the vocabulary. 
    \begin{equation*}
    \text{Matching ratio} = \Ave_{j \in \mathcal{A}} \Big[ \argmax_{j' \le j}\big( \ve_j^\top \mW^{QKOV} \ve_{j'}\big) = j \Big].
    \end{equation*}
\end{enumerate}
This ratio isolates pure token components $\ve_j, \ve_{j'}$ from embeddings and examines whether functionally, 
an identical token maximally activates the 2nd-layer attention through the path of OV--QK circuits.

Additionally, we consider the same copying task with slightly generalized data generation: we restrict the repeating patterns $\vs^\#$ to a subset $\gS$. To be more specific, given length $L$ and size $S$, first the subset $\gS \subset \mathcal{A}^L$ is determined by sampling the pattern $S$ times independently according to Step~1 in Section~\ref{sec:copying}; then each $\vs^\#$ is drawn uniformly from $\gS$. We call $\gS$ the \textit{repetition pool}, and $S=|\gS|$ the \textit{pool size}.

\subsection{Results}


\paragraph{OOD generalization is accompanied by abrupt emergence of subspace matching.} The top row of Figure~\ref{fig:progress} shows that the sharp transition of generalization occurs at training steps around $2000$. In the meantime, the weight matrices in the two layers of the model exhibit a sudden increase of alignment: a large diagonal component in $\mW^{QKOV}$ appears, and the two principal subspaces of $\mW_{\QK}, \mW_{\OV}$ change from being relatively random (similar to random initialization) to highly synchronized.

The structure of $\mW^{QKOV}$ helps to match similar embeddings. Indeed, if we believe that $\mW_{\QK} \mW_{\OV}$ has the ``$\lambda \mI_d $ + noise'' structure, then the embedding $\vx_t$ at position $t$ satisfies $\vx_t^\top \mW_{\QK} \mW_{\OV} \vy \approx \vx_t^\top \vy$. To maximize this inner product, $\vy$ of fixed length must be approximately aligned with $\vx_t$, so embeddings similar to $\vx_t$ tend to receive large attentions in the 2nd layer. Further, subspace matching between QK and OV 
shows that aligning the embeddings depends on the low-dimensional principal subspaces.

\paragraph{Two layers have complementary specialties: position shifting and token matching.} The bottom row of Figure~\ref{fig:progress} provides an explanation for OOD generalization.
The 1st-layer attention head has a large PTH score after training, even for sequences of completely random tokens. The high PTH score indicates that the 1st-layer head specializes in position shifting. In fact, in the ideal case where $A_{t, t-1}=1$, the map $\mX \mapsto \mA \mX$ is simply the shifting operator. Complementarily, the 2nd-layer QK matches the OV circuit and serves as token matching. So upon accepting the shifted tokens as inputs, the 2nd-layer head attends to the next position after the repeated token. Collectively, they yield an IH head, attending correctly to the to-be-copied token; see Figure~\ref{fig:heatmap} (right).

\paragraph{Memorization vs. generalization: pattern diversity matters.} Figure~\ref{fig:heatmap} shows that a smaller pool size $S$ (decreasing from $1000$ to $750$) requires more training steps to achieve good generalization. Moreover, when $S$ drops below $740$, models fail to generalize well under 20K steps; instead, they may memorize the repetition patterns, yielding small ID errors especially for small $S$. We also find that memorizing models exhibit very different attention matrices (Figure~\ref{fig:heatmap-S-1000}--\ref{fig:heatmap-S-100}), suggesting the failure of learning the copying rule.

%% file: src/llm.tex
\section{Intervention experiments in LLMs} \label{sec:llm}

How is the synthetic example relevant to realistic reasoning tasks for LLMs? 
In this section, we address this question by presenting two types of realistic scenarios: out-of-distribution (OOD) prompts (Section~\ref{sec:symbol} and~\ref{sec:cot}) and realistic compositional structures (Section~\ref{sec:cbr}). Through these examples, we aim to demonstrate two key points:
\begin{enumerate}
    \item 
    Prompts (natural language inputs) planted with arbitrarily chosen symbols can be inferred by LLMs in certain tasks without fine-tuning. This reasoning abilities depend crucially on IHs.
    \item 
    Subspace matching as the compositional mechanism takes a more general form  in multilayer and multihead models,
    where a shared latent subspace matches many PTHs and IHs simultaneously.
\end{enumerate}

\paragraph{Pretrained LLMs.} We consider a variety of LLMs in our experiments: (1) Llama2-7B \cite{touvron2023llama}, (2) Llama3-8B \cite{dubey2024llama}, (3) Mistral-7B \cite{jiang2023mistral}, (4) Falcon-7B \cite{almazrouei2023falcon}, (5) Falcon2-11B \cite{malartic2024falcon2}, (6) OlMo-7B \cite{groeneveld2024olmo}, (7) Gemma-7B \cite{team2024gemma}, (8) Gemma2-8B \cite{team2024gemma2}. See Appendix~\ref{sec:append-llm} for details about models and implementations.

\paragraph{Defining induction heads and previous-token heads.}

We sample $N=100$ test prompts with a simple repetition pattern $(\vs^\#, \vs^\#)$: a block of $25$ uniformly random tokens followed by a replica of the block, totaling $T=50$ tokens. For any layer $\ell$ and head $j$ of a Transformer, we denote by $\mA^{\ell,j}_i \in \R^{T \times T}$ the attention matrix defined in Eq.~\ref{def:msa} on a test prompt $i$. By definition $\sum_{t'} (\mA_i)_{t,t'}^{\ell,j} = 1$. This definition of IHs and PTHs only depend on the model, irrespective of downstream tasks.

For each model, we score all attention heads according to Eq.~\ref{eq:PTH-score} and~\ref{eq:IH-score} based on the test prompts, yielding $\mathrm{score}_{\ell, j}^{\mathrm{PTH}}$ and $\mathrm{score}_{\ell, j}^{\mathrm{IH}}$. Then we rank the PTH scores and IH scores in descending order respectively. For a pre-specified $K$, we define PTHs (or IHs) as the top-$K$ attention heads according to $\mathrm{score}_{\ell, j}^{\mathrm{PTH}}$ (or $\mathrm{score}_{\ell, j}^{\mathrm{IH}}$). 

See Appendix~\ref{sec:append-llm} for a set of complete examples we introduce below.

\subsection{Symbolized language reasoning}\label{sec:symbol}
In each task, we sample prompts based on a specified rule and use LLMs directly to predict next tokens. Tokens in blue are the target outputs for prediction. 

\begin{figure*}[t]
\centering
\includegraphics[width=0.85\textwidth]{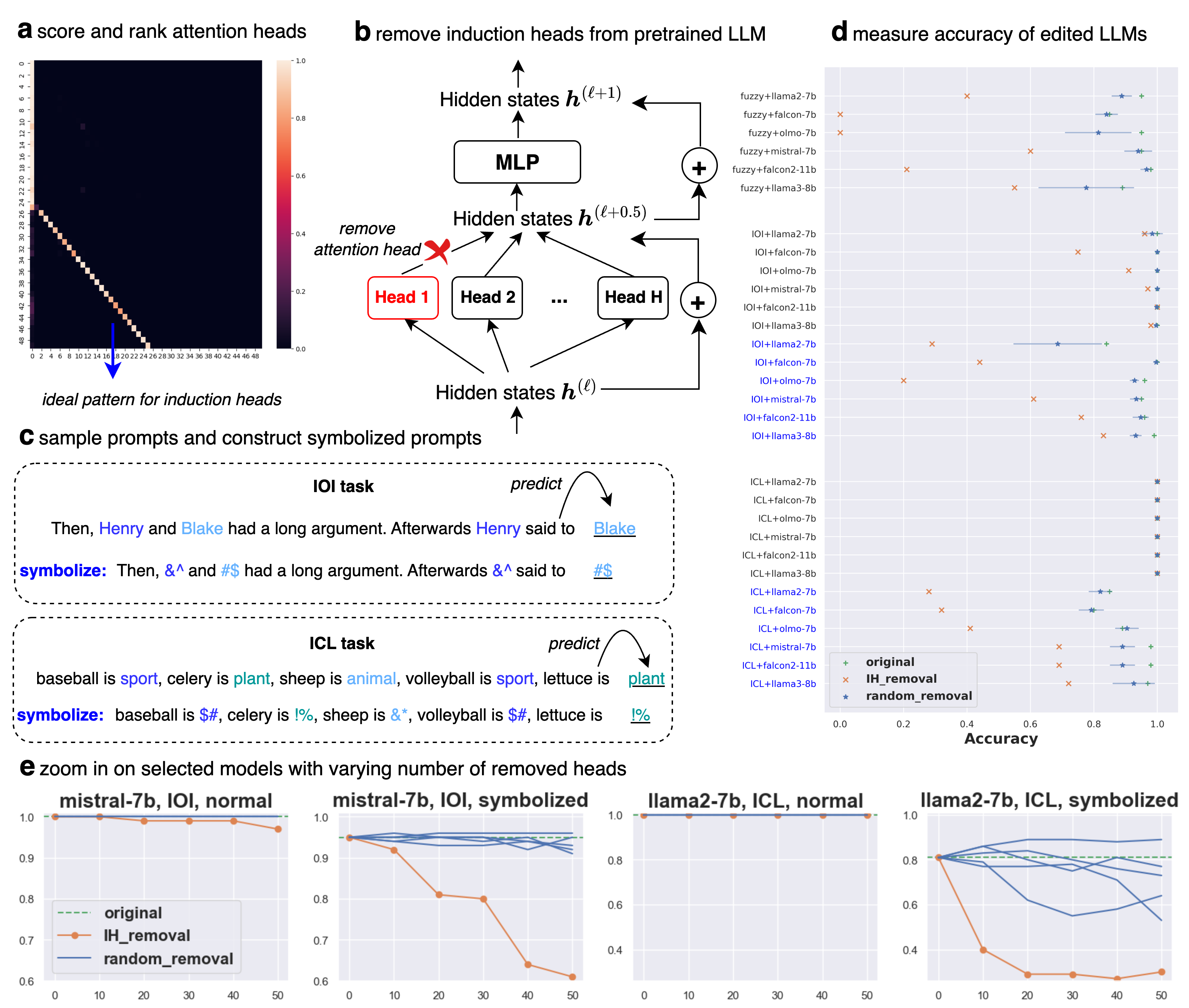}
\caption{
\textbf{LLMs depend on induction heads (IHs) for symbolized language reasoning}. \textbf{(a)} 
We rank attention heads and determine IHs as $K=50$ top-scoring heads.
\textbf{(b)} We remove IHs by manually setting attention matrices to zero. \textbf{(c)} We sample instances according to the rule of each task and then construct OOD instances by symbolizing names/labels. \textbf{(d)} We measure the accuracy of various LLMs under IH removal. Symbolized tasks are indicated by {\color{blue} blue} names. We also report random baseline (deleting $50$ randomly  
selected heads) using $5$ random seeds, and report the variability using a segment showing $\pm$ one standard deviation. \textbf{(e)} We show the accuracy vs.~varying $K$, where a smaller $K$ means deleting fewer heads.}\label{fig:IH}
\end{figure*}

\begin{enumerate}
\item [1.] Fuzzy copying: $[A], [B], [C] \ldots [A'], [B'], {\color{blue} [C']}$
\end{enumerate}
We consider conversion from lower-cased words to upper-cased words. For example, the correct completion of ``bear snake fox poppy plate BEAR SNAKE FOX POPPY'' is ``PLATE''.

The IOI and ICL tasks below were proposed by \cite{wang2023interpretability} and \cite{rong2021extrapolating}, and recently analyzed by \cite{wei2024larger, pan-etal-2023-context, agarwal2024manyshot, halawi2024overthinking,shi2024why}. We extended the method in \cite{rong2021extrapolating} to construct symbolized prompts.

\begin{enumerate}
\item [2.] Indirect object identification (IOI): 
\begin{center} 
$[\text{Subject}] \ldots [\text{Object}] \ldots [\text{Subject}] \ldots {\color{blue} [\text{Object}]}$
\end{center}
\end{enumerate}
We sample input prompts where [Subject] and [Object] are common English names; see Figure~\ref{fig:IH}(c) for an example.  Moreover, we consider a \textit{symbolized} variant where the names of [Subject] and [Object] are replaced by arbitrary special symbols. We pick arbitrary symbols so that the prompt instances are unlikely seen during (thus OOD). 

\begin{enumerate}
\item [3.] In-context learning (ICL): for $f$ that maps an object to its category,
    $$
    x_1, f(x_1), x_2, f(x_2) \ldots x_n, {\color{blue}f(x_n)}.
    $$
\end{enumerate}
We consider mapping an object name $x_i$ to one of the three categories $\{ \text{sport}, \text{plant}, \text{animal}\}$. We concatenate instances that follow the format ``object is category''; see Figure~\ref{fig:IH}(c). Similarly, in the symbolized prompt, categories are replaced by special symbols. Again, the symbolized prompts are OOD because the three class labels are replaced by ``unnatural'' ones, requiring LLMs to infer their meanings at test time.



\paragraph{Experiments with removal of IHs.} Given a prompt, we remove $K=50$ IHs from a pretrained LLM by manually setting its attention matrix $\mA_i^{\ell, j}$ to a zero matrix. This effectively deletes the paths containing the targeted IHs from the architecture. As a baseline for comparison, we randomly select $K$ pairs $(\ell, j)$ from all possible attention heads for deletion, repeated on $5$ random seeds.

\begin{figure*}[t]
\centering
\includegraphics[width=0.7\textwidth]{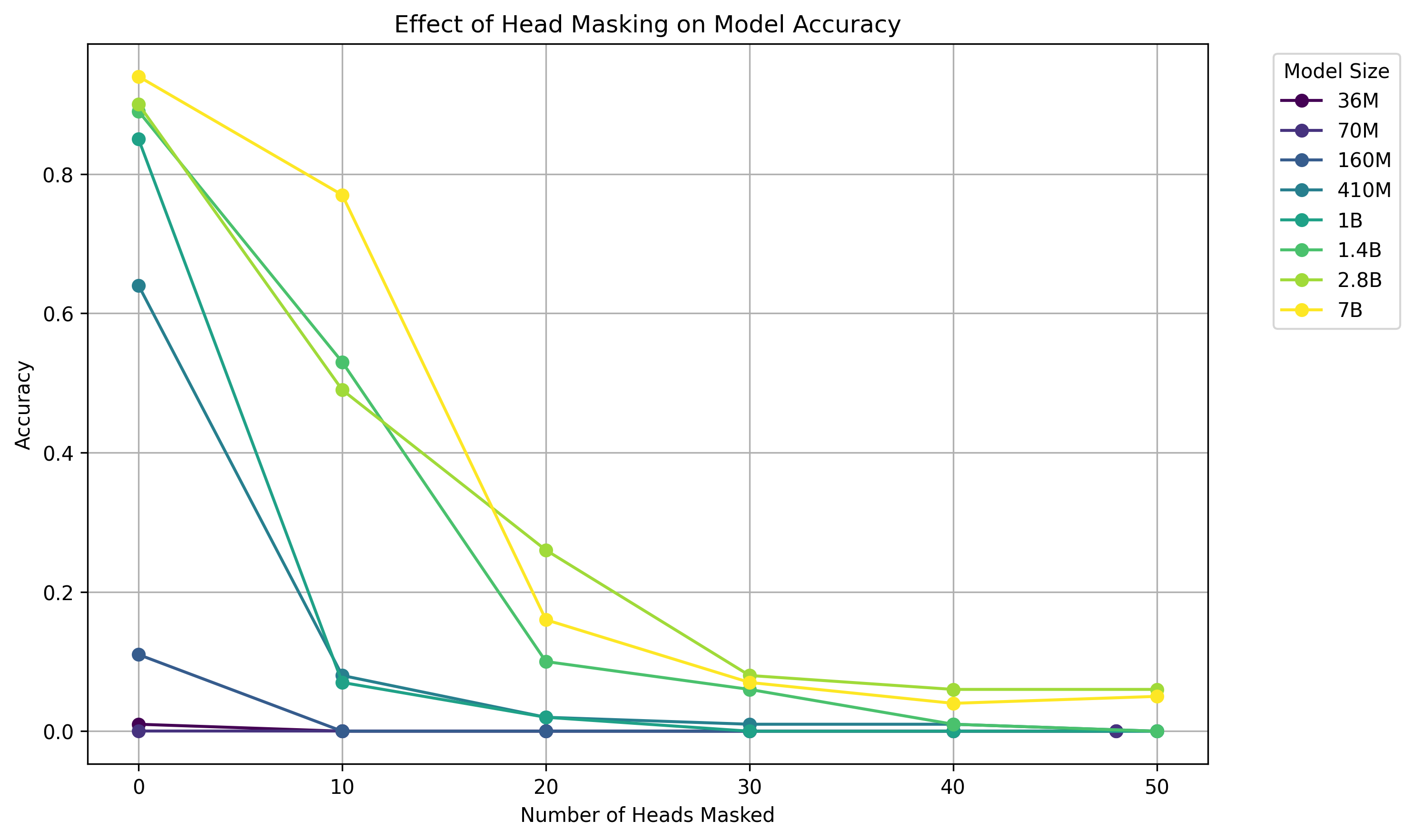}
\caption{\textbf{Scaling experiment on fuzzy copying: how removal of induction heads impact Pythia models}. We use the family of Pythia models of varying sizes, ranging from 36M parameters to 7B parameters. As we increase the number of removed induction heads, there is a consistent accuracy drop of all models on fuzzy copying.}\label{fig:scaling}
\end{figure*}

We measure the prediction accuracy for IOI and ICL in the multiple-choice form, namely we pick the solution with the highest prediction probability among candidate solutions: [Subject] and [Object] for IOI, and the three categories for ICL. We report prediction accuracy based on $100$ sampled prompts for each task. Accuracy by random guess is $1/2$ and $1/3$ respectively. 

In addition, we examine how the accuracy changes across different scales of models. We use the Pythia model suite \cite{biderman2023pythia}, which is a collection of Transformers of varying sizes pretrained on the same data.

\paragraph{Results: inferring from symbolized OOD prompts relies critically on IHs.} Symbolized IOI/ICL instances represent a form of language reasoning, as they require combining semantic understanding with an extra step of converting names/labels to symbols. 
A key finding from Figure~\ref{fig:IH}(d)(e) is that while LLMs generalize reasonably well on OOD prompts, removal of IHs significantly reduces prediction accuracy compared with removal of random heads. 

Interestingly, LLMs on normal IOI and ICL prompts are very robust to IH removal. One explanation for the discrepancy between ID and OOD is that LLMs rely on memorized facts for ID prompts, and use combined abilities (both memorized facts and IHs) to solve OOD prompts. 

\subsection{Mathematical reasoning with chain-of-thought}\label{sec:cot}

\begin{figure}[t]
\centering
\noindent
\begin{minipage}{0.43\textwidth}
\begin{tcolorbox}[nobeforeafter,
title={GSM8K}, height=13cm]
\begin{lstlisting}
"Jerry is cutting up wood for his wood-burning stove. Each pine tree makes 80 logs, each maple tree makes 60 logs, and each walnut tree makes 100 logs. If Jerry cuts up 8 pine trees, 3 maple  trees, and 4 walnut trees, how many logs does he get?"
\end{lstlisting} 
\tcblower
\begin{lstlisting}
"First find the total number of 
pine logs by multiplying the number of trees by the number of logs per tree: 80 logs/pine * 8 pines = <<80*8=640>>640 logs

Then do the same thing for the maple trees: 60 logs/maple * 3 maples = <<60*3=180>>180 logs

And do the same thing for the walnut trees: 100 logs/walnut * 4 walnuts = <<100*4=400>>400 logs

Finally, add up the number of logs from each type of tree to find the total number: 640 logs + 180 logs + 400 logs = <<640+180+400=1220>>1220 logs

#### 1220"
\end{lstlisting}
\end{tcolorbox}
\end{minipage}
\hfill
\begin{minipage}{0.56\textwidth}
\begin{tcolorbox}[nobeforeafter, title=GSM8K-rand Template, height=13cm]
\begin{lstlisting}
"{name} is cutting up wood for his wood-burning stove. Each {pine} tree makes {pine_logs} logs, each {maple} tree makes {maple_logs} logs, and each {walnut} tree makes {walnut_logs} logs. If {name} cuts up {pine_count} {pine} trees, {maple_count} {maple} trees, and {walnut_count} {walnut} trees, how many logs does he get?
\end{lstlisting}
\vspace{0.265cm}
\tcblower
\begin{lstlisting}
"First find the total number of {pine} logs by multiplying the number of trees by the number of logs per tree: {pine_logs} logs/{pine} * {pine_count} {pine} = <<{pine_logs}*{pine_count}={total_pine}>>{total_pine} logs

Then do the same thing for the maple trees: {maple_logs} logs/{maple} * {maple_count} {maple} = <<{maple_logs}*{maple_count}={total_maple}>>{total_maple} logs

And do the same thing for the walnut trees: {walnut_logs} logs/{walnut} * {walnut_count} {walnut} = <<{walnut_logs}*{walnut_count}={total_walnut}>>{total_walnut} logs

Finally, add up the number of logs from each type of tree to find the total number: {total_pine} logs + {total_maple} logs + {total_walnut} logs = <<{total_pine}+{total_maple}+{total_walnut}={total}>>{total} logs

#### {total}"
\end{lstlisting}
\end{tcolorbox}
\end{minipage}
\caption{\textbf{An example from the GSM8K benchmark, and a template from GSM8K-rand that we constructed}. \textbf{Left:} a sample question and its solution with CoT reasoning. The final solution is the number $1220$ after \#\#\#\#. \textbf{Right:} we sample variables \texttt{name}, \texttt{pine}, \texttt{maple}, \texttt{walnut} independently from a pool of $40$ symbol strings. Then we sample other variables such as \texttt{pine\_logs} and \texttt{pine\_count} independently from a range of integers. We ensure the reasoning logic leads to the correct solution \texttt{total}.
}\label{fig:gsm}
\end{figure}

To further examine the impact of induction heads in a more realistic setting, we consider a popular benchmark dataset GSM8K \cite{cobbe2021gsm8k}, which contains a collection of K-12 mathematical questions. Solving the questions requires basic arithmetic abilities---usually a combination of addition, subtraction, multiplication, and division---on top of natural language understanding.

Figure~\ref{fig:gsm} (left) shows an example of a typical question and a solution with chain-of-thought (CoT) reasoning \cite{wei2022chain}. CoT is a standard technique for LLMs to solve reasoning tasks, where intermediate steps are explicitly shown in in-context examples. Given the prompt that contains these in-context examples, LLMs are observed to be significantly better at solving reasoning tasks. We consider 10-shot learning, where $10$ question-solution pairs are added as a prefix to a test question.

To generate explicit OOD test data, we constructed a variant of GSM8K which we call GSM8K-rand. This dataset comprises $12$ templates derived from maintaining the core structures of $12$ question-solution examples from GSM8K and randomizing names and numbers, as shown in Figure~\ref{fig:gsm} (right). Inspired by \cite{mirzadeh2024gsm}, we sample the numbers from a range of integers with 1 to 3 digits, and sample the names and items from a predefined pool. A key difference from \cite{mirzadeh2024gsm} is that the pool for names and items consists of $40$ tuples of special symbols such as ``\$\#'', ``$@$\%'', ``!\&'', ``*\#'', ``\#$@$'', so the test examples were unlikely seen by the LLMs during training. Similar to the previous three tasks, we measure the accuracy of models under the removal of induction heads.  Appendix~\ref{sec:append-llm} details the template construction and measurements.

\paragraph{Results: CoT reasoning relies heavily on IHs.} Similar to Figure~\ref{fig:IH}, we remove induction heads and measure the accuracy of three LLMs on GSM8K and GSM8K-rand.  Table~\ref{tab:gsm} shows that induction heads have a large impact on model accuracy on both datasets, as their removal significantly decrease the model accuracy compared with the removal of the same number of random heads. One possible explanation is that CoT reasoning depends on copying abilities to reuse intermediate outputs in its subsequent calculation. We leave a comprehensive investigation to future work.

\begin{table*}[h!]
\centering
\begin{tabular}{c|ccccccc}
\textbf{Accuracy}                    & \multicolumn{7}{c}{\textbf{Number of IHs / random heads removed}} \\ \hline
& \textbf{} & 0    & 10    & 20     & 30     & 40     & 50     \\ \hline
\multirow{2}{*}{\textbf{Llama2-7B}}  & GSM8K     &   0.11    &  0.06/0.12     &   0.01/0.11     &   0.01/0.09     &   0.02/0.07     &   0.01/0.07     \\
& GSM8K-rand &  0.29     &   0.23 /0.27   &   0.10/0.25     &  0.08/0.19      &   0.06/0.20     &  0.08/0.16      \\ \hline
&           & 0     & 50    & 100    & 150    & 200    & 250    \\ \hline
\multirow{2}{*}{\textbf{Llama2-70B}} & GSM8K     &  0.59     &   0.41/0.55    &  0.21/0.52      &   0.21/0.52   &   0.09/0.53     &   0.01/0.50     \\
& GSM8K-rand &   0.87    &   0.68/0.81    &   0.53/0.80     &   0.51/0.80     &  0.29/0.78      &   0.10/0.76     \\
\multirow{2}{*}{\textbf{Llama3-70B}} & GSM8K     &   0.89    &   0.88/0.90    &   0.92/0.90     &  0.86/0.89      &  0.60/0.88     &  0.51/0.87      \\
& GSM8K-rand &   0.96    &   0.96/0.96    &   0.95/0.97     &   0.61/0.98     &   0.40/0.97     &   0.51/0.94    
\end{tabular}
\caption{CoT reasoning depends on IHs on mathematical tasks. Accuracy comparison under removal of IHs vs.~random heads (averaged over 5 random seeds)
}\label{tab:gsm}
\end{table*}




\subsection{Common bridge representation hypothesis}\label{sec:cbr}

How do compositions work in LLMs beyond the synthetic example? 
We consider four tasks: (i) Copying, (ii) Fuzzy Copying, (iii) IOI, and (iv) ICL. For illustrative purposes, we report results for the copying task. The results for the other three tasks are similar, which can be found in Appendix~\ref{sec:append-cbrh}. With a microscopic analysis,
we posit the following \textit{Common Bridge Representation} (CBR) hypothesis. 

\begin{center}
\textit{For compositional tasks, a latent subspace stores intermediate representations from the outputs of relevant attention heads and then matches later heads.}
\end{center}
The CBR hypothesis is an extension of linear representation hypothesis to compositional tasks. We give a mathematical statement of the CBR hypothesis in the ideal scenario. For a matrix $\mW_0 \in \R^{d \times d}$, we denote by $\mathrm{span}(\mW_0) := \{ \mW_0 \vx: \vx \in \R^d\}$ the column span of $\mW_0$. Let $\mW_{\OV,j}, \mW_{\QK,k} \in \R^{d \times d}$ be the matrices in a task-relevant OV/QK circuit in an earlier/later layer respectively. Then, for a compositional task, there exists a low-dimensional subspace $\mathcal{V} \subset \R^d$ such that
\begin{equation}\label{def:CBR}
\mathcal{V} = \mathrm{span}(\mW_{\OV,j}) = \mathrm{span}(\mW_{\QK,k}^\top).
\end{equation}
Intuitively, the output of an OV circuit has the form $\mW_{\OV,j}\vx$, which lies in $\mathcal{V}$. It needs to be processed by a later QK circuit through $\mW_{\QK,k}\vx$ to match relevant hidden states.

LLMs in practice are trained with stochastic gradient descent, so we do not expect Eqn.~\ref{def:CBR} to hold exactly. Nevertheless, it motivates us in our experiment to estimate $\mathcal{V}$ from a collection of relevant QK circuits: given $(\mW_{\QK,k})_k^K$, we apply singular value decomposition to the stacked matrix $[\mW_{\QK,1}, \ldots, \mW_{\QK,K}]$ and use the principal right singular subspace as an estimate of $\mathcal{V}$. 

\begin{figure*}[t]
\centering
\includegraphics[width=0.95\textwidth]{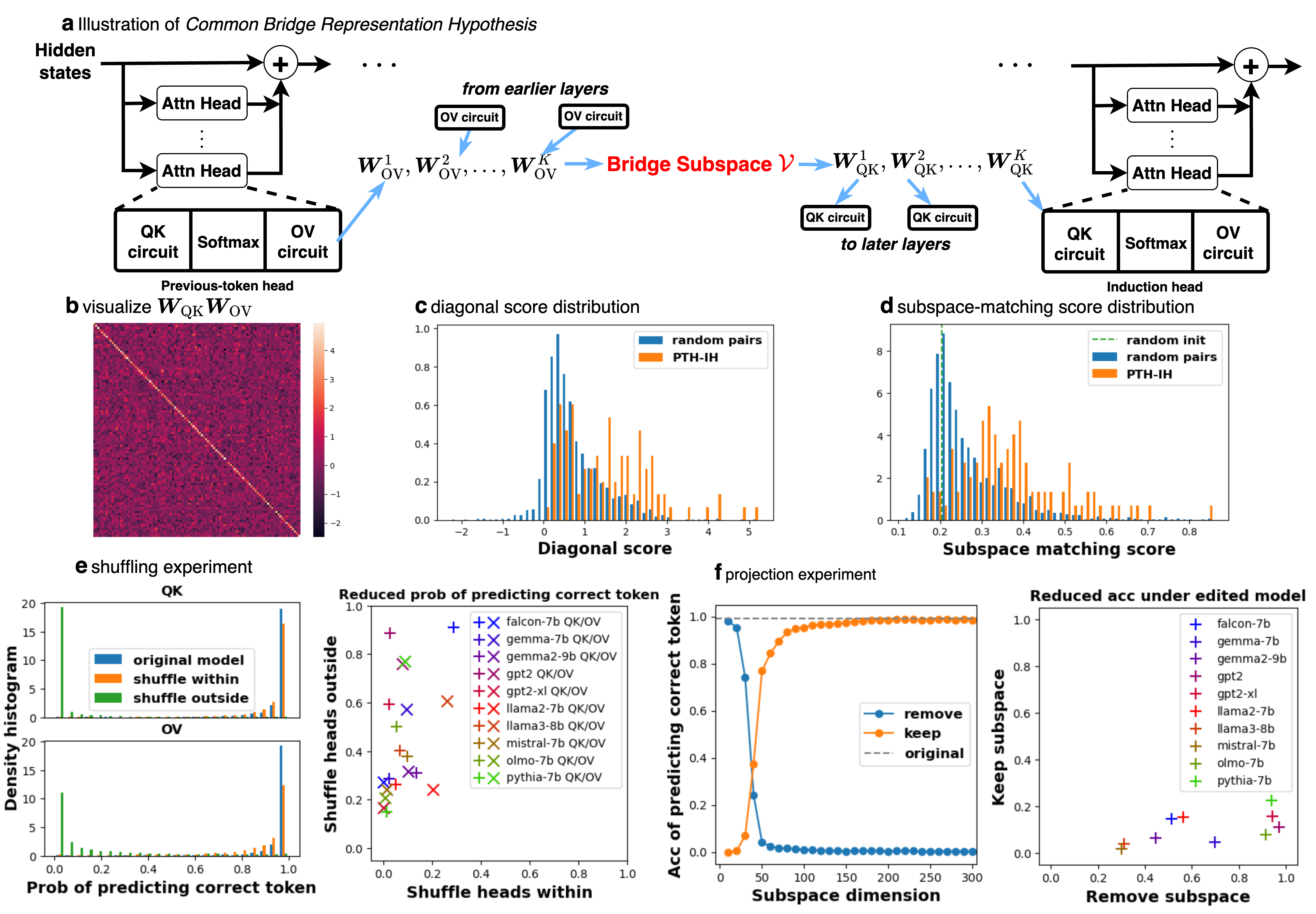}
\caption{\textbf{Common bridge representation hypothesis as the mechanism of composition in LLMs}. (a) Linear maps from circuits in relevant attention heads are connected by an intermediate \textit{bridge subspace}. (b) We visualize $\mW_{\QK} \mW_{\OV}$ using highest-ranking PTH and IH. (c) Distributions of diagonal scores, comparing pairs from top-$10$ PTHs/IHs and pairs from random QK/OV circuits. (d) Distributions of subspace matching scores. (e) Shuffling $\mW_{\QK}$ (or $\mW_{\OV}$) among top-scoring attention heads mildly changes probabilities of correct prediction, whereas replacing these QK or OV matrices with random heads significantly reduces the probabilities. (f) Removing a low-rank ($r=50$ for GPT-2, $r=\frac{1}{20}d$ for others) bridge subspace from attention calculation ('remove') heavily reduces prediction accuracy, whereas only using this subspace for attention calculation ('keep')  yields mild loss of accuracy.}\label{fig:CBRH}
\end{figure*}

We experiment on a variety of LLMs and highlight key findings in Figure~\ref{fig:CBRH}(b)--(f), where we use GPT-2 as a recurring example and summarize results of other models. 
Top-scoring PTHs and IHs are distributed across different layers of LLMs, though more PTHs appear in early layers. We sample sequences of the format $(\vs^\#, \vs^\#, \vs^\#)$ and calculate the average token-wise probability/accuracy for predicting the third segment $\vs^\#$. See experiment details in the  Appendix.




\paragraph{Experiments with two interventions.} 
For both experiments, we apply screening to top PTHs and IHs based on diagonal scores, resulting in an average of $9$ effective PTHs and $7$ effective IHs. The first intervention experiment involves shuffling heads. We randomly permute matrices  $\mW_{\mathrm{QK}}$ within the list of IHs, yielding an edited model (``shuffle within''). As comparison, we replace each $\mW_{\mathrm{QK}}$ within the list by a random $\mW_{\mathrm{QK}}$ outside the list, yielding another edited model (``shuffle outside''). In a parallel experiment, we shuffle $\mW_{\mathrm{OV}}$ circuits within PTHs similarly. We evaluate the original model and edited models by calculating the average probability of predicting correct tokens. 

The second intervention experiment involves projection of weight matrices based on the discussion on Eqn.~\ref{def:CBR}. First, we stack the QK matrices from IHs into a matrix $[ \mW_{\QK}^1, \ldots, \mW_{\QK}^K]$ and extract the top right singular vectors $\mV \in \R^{d \times r}$ for a pre-specified $r$. 
We call the column linear span of $\mV$ as the \textit{bridge subspace}. Then, we edit $25\%$ of all attention heads by weight projection $\mW_{\QK} \leftarrow \mW_{\QK} \mV \mV^\top$. After the edit (`keep'), the attention calculation can only use the component of embeddings within the bridge subspace. In a parallel experiment, we make a projection edit (`remove') with an orthogonal complement $\mW_{\QK} \leftarrow \mW_{\QK} (\mI_d - \mV \mV^\top)$ to force attention calculation not to use component in the bridge subspace. We evaluate the edited models by calculating the average accuracy of predicting correct tokens.

\paragraph{Results: Subspace matching is a pervasive mechanism of compositions.} 

Fig.~\ref{fig:CBRH}(b)(c) show that $\mW_{\QK} \mW_{\OV}$ contains large diagonal entries when the circuits are selected from PTHs and IHs. Further, Fig.~\ref{fig:CBRH}(d) shows that leading (left) singular spaces of $\mW_{\OV}$ and (right) singular spaces of $\mW_{\QK}$ are highly aligned. Among many PTHs and IHs across different layers, subspace matching is a pervasive mechanism, which extends the findings from the synthetic example to realistic models.



\paragraph{Results: relevant attention heads share a common latent  
subspace.} We further show that the pairwise matching between circuits is not a coincidence; instead, a global latent subspace provides a ``bridge'' connecting relevant OV/QK circuits. The result of the shuffling experiment in Fig.~\ref{fig:CBRH}(e) indicates that $\mW_{\QK}$ (or $\mW_{\OV}$) from IHs (or PTHs) are approximately exchangeable, since permuting $\mW_{\QK}$ (or $\mW_{\OV}$) does not reduce prediction probabilities significantly. 
Moreover, Fig.~\ref{fig:CBRH}(f) further supports that $\mathcal{V}$ contains nearly complete information about copying, because removing/keeping this low-rank subspace from attention calculation drastically changes prediction accuracy in opposite directions.

The common bridge subspace is connected to the statistical literature on spiked matrices \cite{spiked}. Roughly speaking, the principal subspaces of $\mW_{\OV}^j, \mW_{\QK}^j$ correspond to a shared spike, which contains relevant information for the compositional task.



%% file: src/related.tex
\section{Related work}\label{sec:related}

There are many recent papers on analyzing and interpreting Transformers \cite{zhang2024trained, bai2024transformers, song2023uncovering, hsu2024mechanistic}; see \cite{anwar2024foundational} for a comprehensive survey. We highlight a few key threads of research.

\paragraph{Mechanistic interpretability and induction heads.} Mechanistic interpretability (MI) aims to provide microscopic interpretability of inner workings of Transformers \cite{olah2020zoom}. In particular, \cite{elhage2021mathematical, inductionhead22} proposed IHs as crucial components of Transformers. In particular, they suggested that matching OV circuits and QK circuits is crucial for ICL. A line of empirical research extends IHs in various aspects \cite{wang2023interpretability, reddy2023mechanistic, merullo2023circuit, singh2024needs, ferrando2024primer, akyurek2024context,gould2024successor,Akyrek2024InContextLL}, and \cite{nichani2024transformers} provides theoretical analysis of IHs assuming a latent causal graph. Compared with the existing literature, we conducted extensive experiments on both small Transformers and a variety of LLMs, demonstrating that IHs are crucial components for many reasoning tasks. Further, we propose that subspace matching---and more broadly common bridge representation hypothesis---as the compositional mechanism.

\paragraph{OOD generalization and compositions.} Historically, studies of generalization on novel domains focus on extrapolation, distribution shift, domain adaptation, etc. Since GPT-3 \cite{brown2020language}, 
recent studies of OOD generalization focus on arithmetic and algebraic tasks \cite{lee2023teaching, zhang2022unveiling}, formal language and deductive reasoning \cite{saparov2024testing, reizinger2024understanding,tang2023large, Grokked}, learning boolean functions \cite{abbe2023generalization}, etc. Our copying task is an example of length generalization \cite{anil2022exploring, zhou2023algorithms}. Compositional tasks are strongly related to reasoning abilities of LLMs, such as arithmetic tasks \cite{dziri2024faith, kudo2023deep}, formal language \cite{hahn2023theory}, logical rules~\cite{boix-adsera2024when,xu2024do}, and so on. Despite recent insights \cite{arora2023theory, dziri2024faith, 2405-15302, chen2024can}, there is a lack of systematic and quantitative analysis. Our work is a first step toward understanding compositional capabilities of LLMs in a systematic way.

\paragraph{Linear Representation Hypothesis.} 
Linear Representation Hypothesis (LRH) states that monosemantic (meaning basic or atomic) concepts are represented by linear subspaces (or half-spaces) in embeddings \cite{arora-etal-2018-linear,  elhage2022superposition, Park2023TheLR}. This hypothesis is empirically verified not only in simpler statistical models but also in LLMs \cite{Mikolov2013EfficientEO, pennington-etal-2014-glove}.
The LRH treats subspaces as fundamental units for representing linguistic concepts, but it does explain how models solve reasoning tasks or achieve OOD generalization. Our CBR hypothesis furthers this view by linking intermediate outputs in compositions to interpretable subspaces. 


%% file: src/limit.tex
\section{Limitations and future work}\label{sec:limit}



First, our hypothesis is a general conjecture on the mechanism of compositions in Transformers, based on our analysis of IHs. While IHs are pervasive in LLMs, other components or mechanisms for compositions may exist. Additionally, our interpretations are based on a simplified form of self-attention. It would be interesting to explore alternative mechanisms for compositions, and examine variants or practical techniques in LLMs that may impact our hypothesis.

Second, we did not develop insights to explain the emergence of the bridge subspace during training. The sharp transition in prediction accuracy is related to the \textit{emergent abilities} of LLMs observed in broader contexts \cite{wei2022emergent}. In simpler settings, feedforward networks learning algebraic rules also exhibit phase transitions from memorization to generalization, a phenomenon known as \textit{Grokking} \cite{power2022grokking, nanda2023progress, zhong2024clock, lyu2023dichotomy, liu2023omnigrok, mallinar2024emergence}. Analyzing the training dynamics of Transformers for copying and other compositional tasks would be an interesting avenue for further research.






%% file: src/availability_acknowledge.tex
\section*{Code and data availability}

The code for replicating the experiments and analyses can be found in GitHub page: 
\begin{center}
\url{https://github.com/JiajunSong629/ood-generalization-via-composition} 
\end{center}
\section*{Acknowledgement}

This work was partially supported by NSF-DMS grant 2412052 and by the Office of the Vice Chancellor for Research and Graduate Education at the UW Madison with funding from the Wisconsin Alumni Research Foundation. We are grateful for the feedback from Yingyu Liang, Haolin Yang, Junjie Hu, and Robert Nowak.

%% file: src/appendix/notations.tex
\section{Notations}\label{sec:append-notations}

\begin{itemize}
\item $\mathcal{A}$ is the vocabulary which is a discrete set. A token $s$ is an element in $\mathcal{A}$. $\mathcal{A}^L$ denotes the product of $L$ copies of $\mathcal{A}$.
\item $\vs = (s_1,s_2,\ldots,s_T)$ denotes a sequence of tokens where $s_t \in \mathcal{A}$ for $t \le T$, and $\vs_{<t} = (s_1,\ldots,s_{t-1})$.
\item $p_t(s|\vs_{<t})$ is a conditional probability mass function given by a pretrained language model, where $t$ is up to a maximum sequence length.
\item $\mathcal{S} \subset \mathcal{A}^L$ is the set of all possible repeating pattern $\vs^\# = (s_1,s_2,\ldots,s_L)$. We use $S$ to denote the cardinality of $\mathcal{S}$, which we call the pool size.
\item $\mA \in \R^{T \times T}$ is the attention matrix. The attention weight satisfies $A_{t,t'} \in [0,1]$ and $\sum_{t'} A_{t,t'} = 1$ for each $t$. LLMs are GPT-styled Transformers, which apply a mask to attention matrices that effectively sets $A_{t,t'} = 0$ for $t' > t$ (``no look into the future'').
\item $\mI_d \in \R^{d \times d}$ is the identity matrix.
\item $\mathrm{span}(\mV)$ is a $r$-dimensional subspace in $\R^d$ representing the column linear span of a matrix $\mV \in \R^{d \times r}$.
\item $\sigma_{\max}(\mM)$ is the largest singular value of a matrix $\mM$, and $\sigma_j(\mM)$ is the $j$-th largest singular value.
\item $\mathrm{Ave}_{i \in \gI}(a_i)$ denotes the average of a collection of numbers $a_i$ indexed by $i \in \gI$, namely $\frac{1}{|\gI|} \sum_{i \in \gI} a_i$. Similarly, $\mathrm{Std}_{i \in \gI}(a_i)$ denotes the standard deviation of a collection of numbers $a_i$.
\end{itemize}


%% file: src/appendix/synthetic-model.tex
\subsection{Two-layer Transformer model}\label{sec:append-synthetic-model}

We detail the model hyperparameters in our synthetic example. The embedding dimension (or model dimension) is $64$, maximum sequence length is $64$. We sample both $5000$ sequences for calculating ID errors and OOD errors. The $\theta$ parameter in RoPE is $10000$. We apply layer normalization before the self-attention in each layer. We use Rotary positional embedding in each of the two layers.

Optimization related hyperparameters are listed below.
\begin{itemize}
\item Learning rate: 0.001
\item Weight decay: 0.0005
\item Batch size: 50
\item Dropout: 0.1
\end{itemize}

%% file: src/appendix/computing-err.tex
\subsection{ID error and OOD error}\label{sec:append-err}

We provide explanations for the ID error and the OOD error for Figure~\ref{fig:main}. At any give training step, the Transformer model gives the conditional probability $p_t(s|s_{<t})$ and makes prediction based on the token that maximizes the probability, namely $\hat s_{t} = \argmax_{s\in \mathcal{A}} p_t(s|s_{<t})$ for input sequence $s_{<t}$.

Given the $i$-th ID/OOD test sequence of the format $\vs_i = (*, \vs_i^\#, *, \vs_i^\#, *)$, we are interested in how well the model predicts the second segment $\vs_i^\#$. Let the length of $\vs_i^\#$ be $L_i$ and the starting index of the second $\vs_i^\#$ be $m_i$. Note that $L_i \in \{10,11,\ldots,19\}$ for ID and $L_i = 25$ for OOD. We calculate ID/OOD errors:
\begin{equation*}
    \mathrm{Err} = \Ave_{i \le n, m_i+5 \le t \le m_i + L_i}\big( \mathbf{1}((\hat{\vs}_i)_t \neq (\vs_i)_t) \big)\,.
\end{equation*}
We choose $m_i+5$ instead of $m_i+1$ as the starting position for calculating errors, because the model needs a burn-in period before starting repeating patterns due to the presence of `noise' tokens.

%% file: src/appendix/one-layer-tf.tex
\subsection{One-layer Transformer}\label{sec:append-one-layer}

Figure~\ref{fig:main} shows that one-layer Transformers do not learn the rule of copying under 20K training steps. We used the same model architecture and hyperparameters for training as detailed in Section~\ref{sec:append-models}.

\begin{figure}[h!]
\centering
\includegraphics[width=0.7\textwidth]{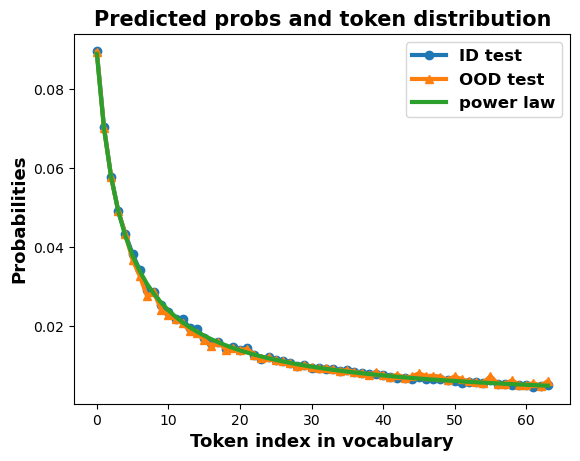}
\caption{Prediction probabilities of one-layer Transformers on the copying task after 20K training steps.
}\label{fig:append-distr}
\end{figure}

Here we measure the prediction probabilities of the one-layer Transformer after 20K training steps, namely $p_t(s_t|s_{<t})$. Recall that we used a power law distribution $\gP$ to generate training sequences and ID test sequences, and a uniform distribution $\gP_{\mathrm{OOD}}$ to generate OOD test sequences. Figure~\ref{fig:append-distr} shows the probabilities $p_t(s_t|s_{<t})$ gathered over different positions $t$ and sequences the ID test dataset, OOD test dataset respectively. We also plot the power law distribution $\gP$.

The three curves are almost identical, which suggests that one-layer Transformer only learns the marginal distribution of input sequences---which is the token distribution $\gP$---for predicting the next token, and that it fails to learn the rule of copying. 

%% file: src/appendix/larger_models.tex
\subsection{Larger Transformers}\label{sec:append-larger}

We believe that the 2-layer 1-head Transformer with no MLP is a minimal yet representative model for analyzing the sharp transition. To demonstrate this, we consider training other Transformers using the same dataset:
\begin{enumerate}
    \item 2-layer 1-head Transformer with MLP,
    \item 4-layer 4-head Transformer without MLP,
    \item 8-layer 8-head Transformer without MLP.
\end{enumerate}
The results of the three models are similar to the minimal model (2-layer 1-head Transformer with no MLP) we analyzed in the paper. See plots in Section~\ref{sec:append-larger-tfs}.

%% file: src/appendix/model_variants.tex
\subsection{Details about models}\label{sec:append-model-variants}


We make connections to the original Transformer \cite{vaswani2017attention}. Let $\vx_1, \ldots, \vx_T \in \R^d$ be a sequence of input embedding vectors or hidden states in the intermediate layers, and denote $\mX = [\vx_1,\ldots,\vx_T]^\top \in \R^{T \times d}$. Let $d_{\head} \le d$ be the head dimension. Given the query/key/value weights $\mW^q, \mW^k, \mW^v \in \R^{d \times d_{\head}}$ respectively, a self-attention head calculates
\begin{equation}\label{def:sa}
    \mathrm{AttnHead}(\mX; \mW^q, \mW^k, \mW^v) = \mathrm{softmax} \left( \frac{\mX \mW^q (\mW^k)^\top \mX^\top }{\sqrt{d_\head}}\right) \mX \mW^v
\end{equation}
where $\mathrm{softmax}(\mZ)_{ij} = \exp(Z_{ij}) / \exp(\sum_{t}Z_{it})$, 
Now let the number of attention heads be $H$ where $H d_{\head} = d$, and suppose that there are $H$ sets of matrices $(\mW^{q,j}, \mW^{k,j}, \mW^{v,j})$ where $j \le H$. Given an output matrix $\mW^o \in \R^{d \times d}$, we partition it into $H$ blocks of equal sizes $\mW^o = [\mW^{o,1}, \ldots, \mW^{o,H}]$. Let $\mW$ be the collection $\mW = (\mW^{q,j}, \mW^{k,j}, \mW^{v,j}, \mW^{o,j})_{j \le H}$. Then, the $\mathrm{MultiHead}$ attention calculates
\begin{equation}\label{def:mh}
    \mathrm{MuliHead}(\mX; \mW) = \sum_{j=1}^H \mathrm{softmax} \left( \frac{\mX \mW^{q,j} (\mW^{k,j})^\top \mX^\top }{\sqrt{d_\head}}\right) \mX \mW^{v,j} (\mW^{o,j})^\top
\end{equation}
Comparing with Eq.~\ref{def:msa}, we identify $\mW^{q,j} (\mW^{k,j})^\top / \sqrt{d_{\head}}$ with $\mW_{\QK,j}$ and $\mW^{v,j} (\mW^{o,j})^\top$ with $\mW_{\OV,j}^\top$. 

Now we explain the major differences between our simplified Transformer in Eq.~\ref{def:msa} and Transformers in practice. We note the following differences.
\begin{itemize}
    \item Practical Transformers usually contain layer normalization and sometimes bias terms.
    \item Practical Transformers contain an MLP sublayer following every multihead self-attention.
    \item Early Transformers use the absolute positional embedding whereas more recent Transformers use rotary positional embedding.
\end{itemize}
The justification for the simplified Transformer is discussed in \cite{elhage2021mathematical}. We provide some comments about positional embeddings.

\paragraph{Positional embedding.} There are two major variants of positional embedding: absolute positional embedding (APE) and rotary embedding (RoPE). APE is proposed in the original Transformer paper \cite{vaswani2017attention} and used widely in early LLMs such as GPT-2. The input vector to a Transformer is obtained by adding a token-specific embedding vector and a position-specific  embedding vector, thus encoding both token information and positional information. RoPE \cite{su2024roformer} is often used in recent LLMs such as Llama-2. The positional information is not encoded at the input layer; instead, RoPE directly modifies the attention by replacing $\mW^q (\mW^k)^\top$ in Eq.~\ref{def:sa} with $\mW^q \mR_{\Delta t} (\mW^k)^\top$ where $\mR_{\Delta t} \in \R^{d_{\head} \times d_{\head}}$ is a matrix that depends on relative distance of two tokens. In particular, if $\Delta t=0$ then $\mR_{\Delta t} = \mI_d$. Our analysis in Section~\ref{sec:synthetic} and \ref{sec:llm} essentially treats $\Delta t=0$.

Section~\ref{sec:append-synthetic-rope} explores the impact of $\mR_{\Delta t}$ on our results. We did not find significant differences when we repeat our measurements with a nonzero $\Delta t$.


%% file: src/appendix/measurements.tex
\subsection{Details about measurements}\label{sec:append-measure}

\paragraph{Subspace matching score.} Note that the above definition is invariant to the choice of bases $\mU, \mV$. Indeed, for different orthonormal matrices $\mU' = \mU \mO_1$, $\mV' = \mV \mO_2$ where $\mO_1, \mO_2$ are two orthogonal matrices, we have $\sigma_{\max}(\mU^\top \mV) = \sigma_{\max}((\mU')^\top (\mV'))$.

This score is a generalization of vector cosine similarity since
\begin{equation*}
    \sigma_{\max}(\mU^\top \mV) = \max_{\| \vy_1\|_2 = \| \vy_2 \| = 1} \langle \mU \vy_1, \mV \vy_2 \rangle \,.
\end{equation*}
which is equivalent to the inner product between two optimally chosen unit vectors in the two subspaces. In the special case $r=1$, $\mU, \mV$ are two vectors, and this definition reduces to the regular cosine similarity between two vectors (after taking the absolute values).

%% file: src/appendix/larger_models_results.tex
\subsection{Larger models show similar sharp transitions}\label{sec:append-larger-tfs}

\begin{figure}[h!]
\centering
\includegraphics[width=0.7\textwidth]{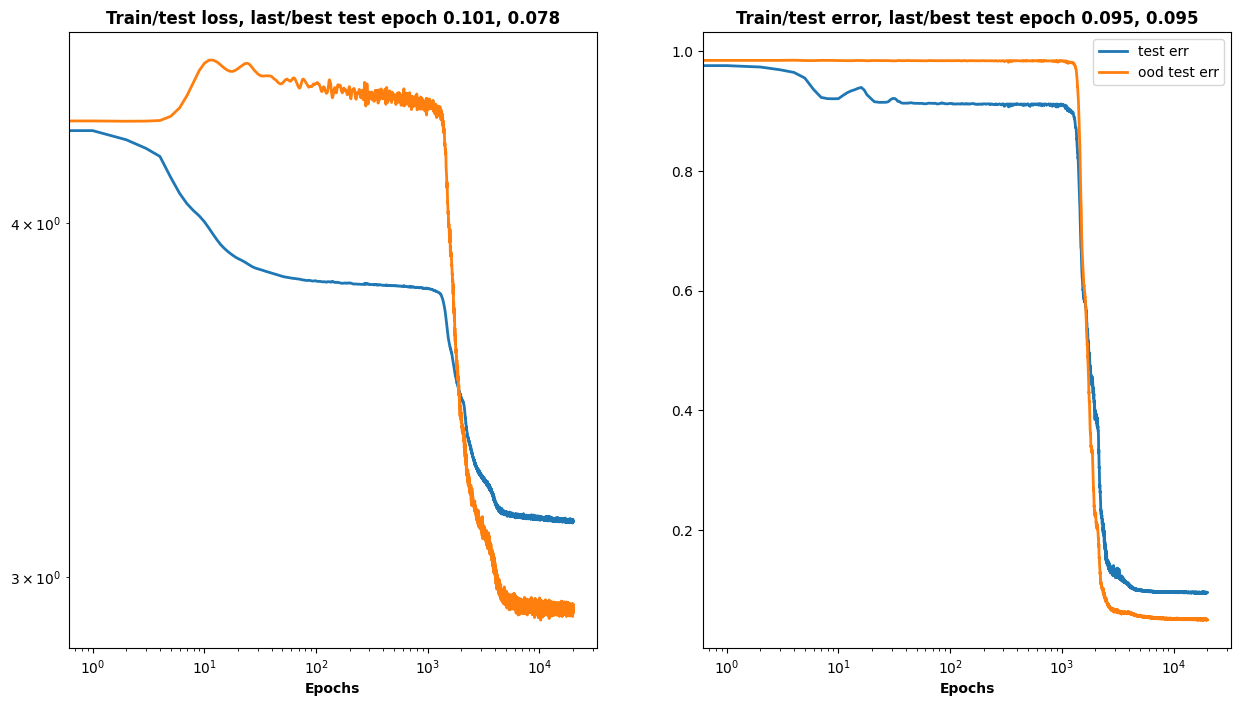}
\caption{2-layer 1-head Transformer \textbf{with MLP}. ID/OOD test errors vs.~training steps.
}\label{fig:append-model-mlp}
\end{figure}

\begin{figure}[h!]
\centering
\includegraphics[width=0.7\textwidth]{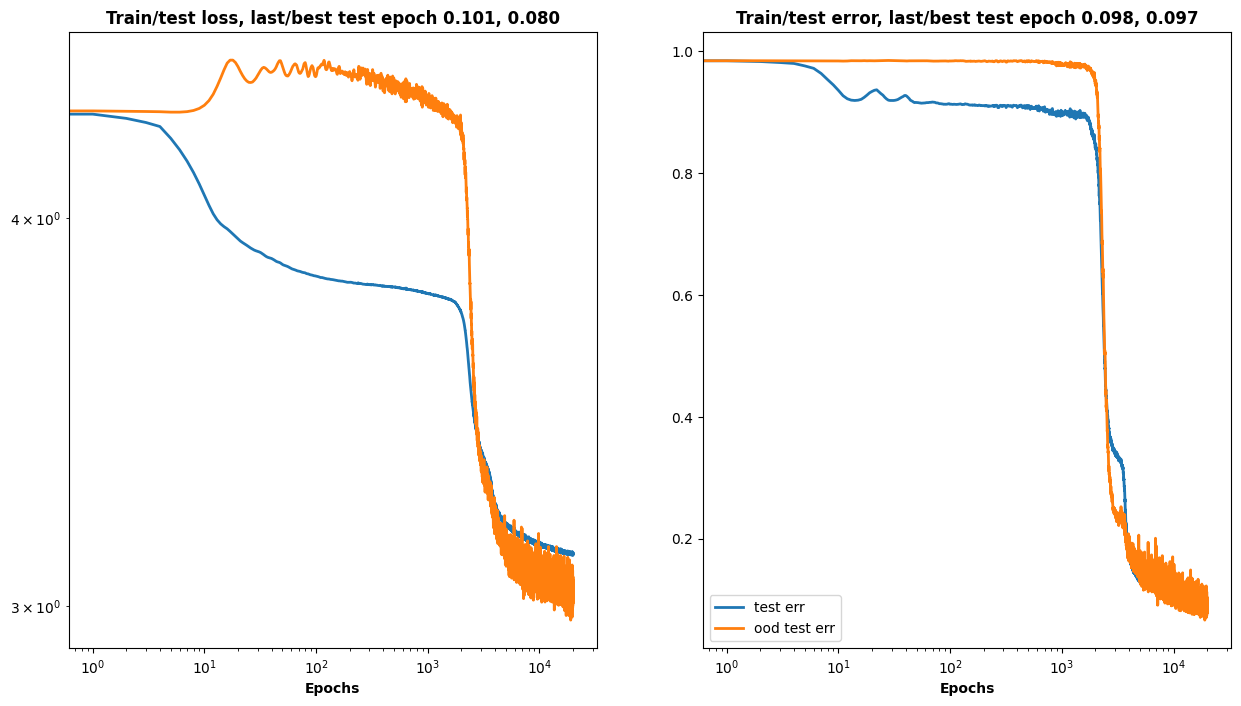}
\caption{4-layer 4-head Transformer without MLP. ID/OOD test errors vs.~training steps.
}\label{fig:append-model-4L4H}
\end{figure}

\begin{figure}[h!]
\centering
\includegraphics[width=0.7\textwidth]{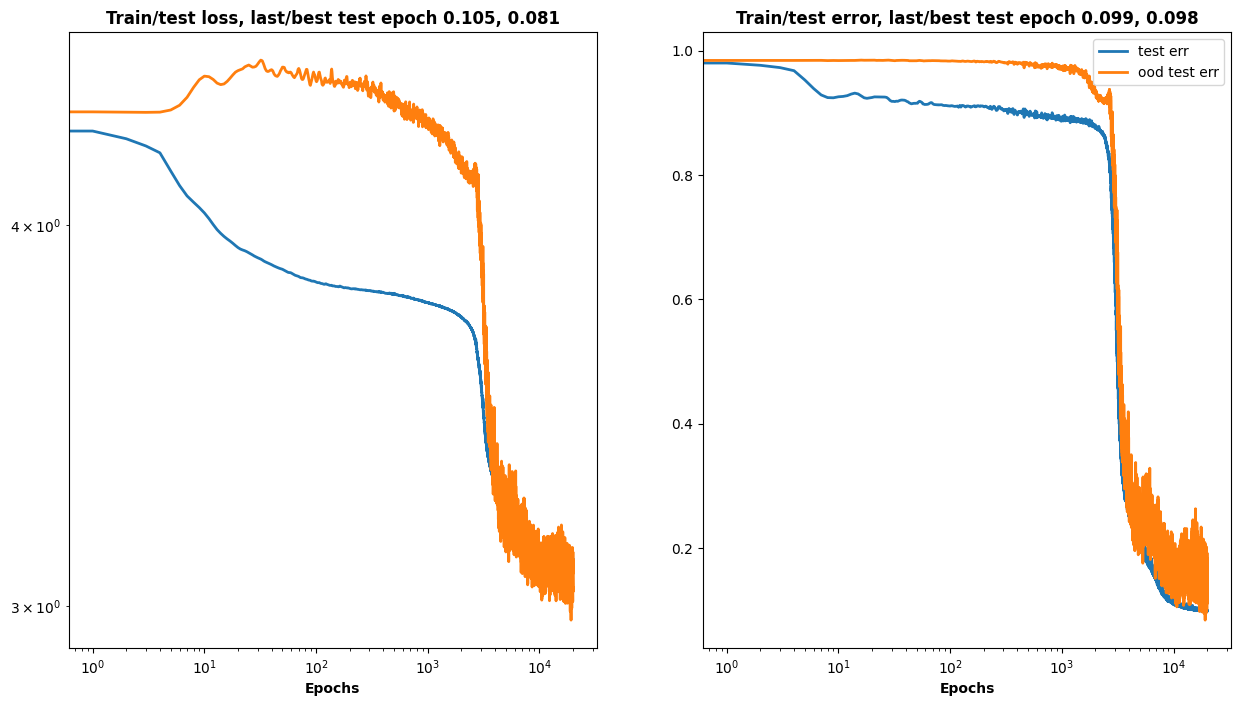}
\caption{8-layer 8-head Transformer without MLP. ID/OOD test errors vs.~training steps.
}\label{fig:append-model-8L8H}
\end{figure}

\begin{figure}[h!]
\centering
\includegraphics[width=0.9\textwidth]{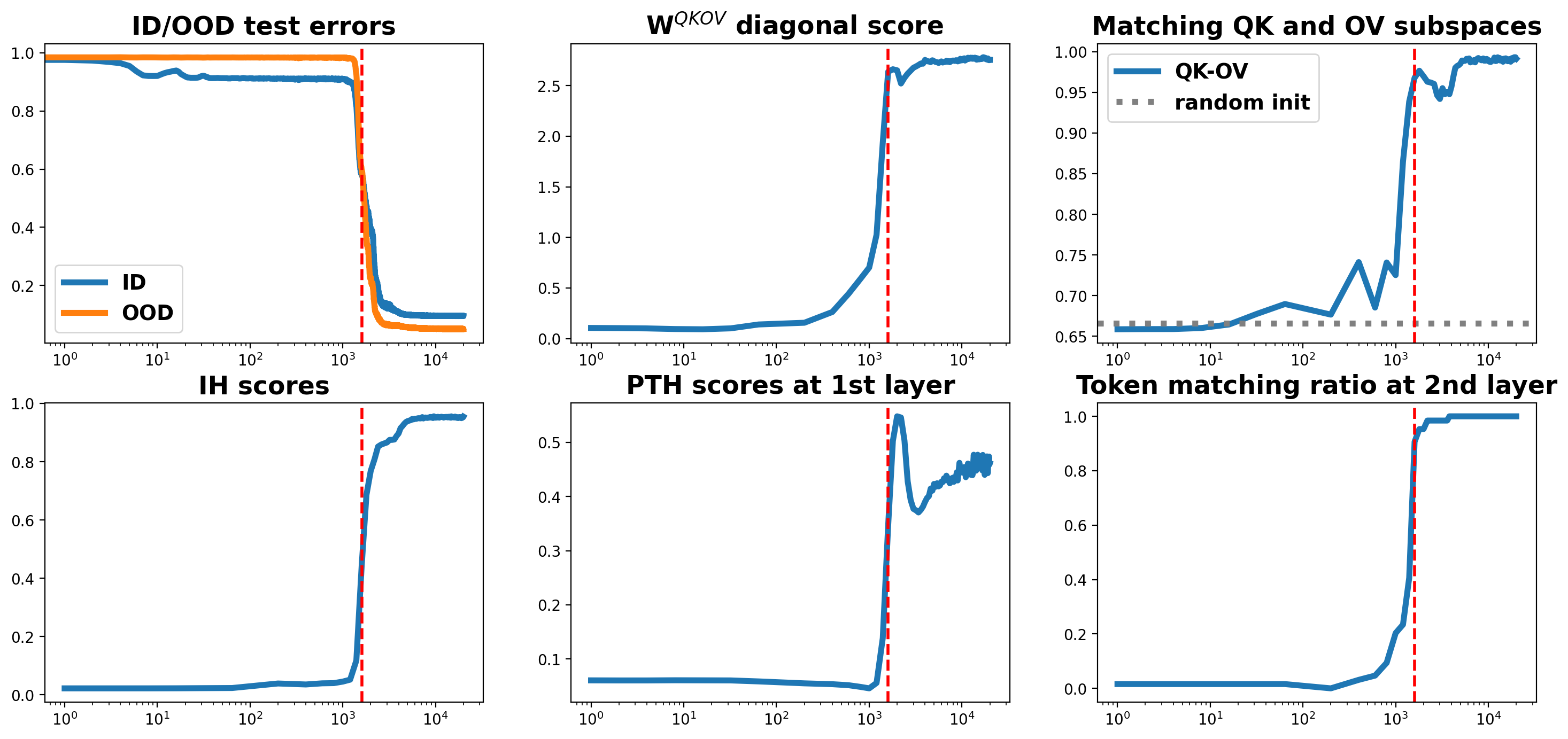}
\caption{Progress measures for 2-layer 1-head Transformer \textbf{with MLP}.
}\label{fig:append-model-mlp-measures}
\end{figure}

We believe that the 2-layer 1-head Transformer with no MLP is a minimal yet representative model for analyzing the sharp transition. We show similar results on three larger Transformer models in Figure~\ref{fig:append-model-mlp} to~\ref{fig:append-model-mlp-measures}.

%% file: src/appendix/synthetic.tex
\subsection{Progress measures under varying pool sizes}

\begin{figure}[h!]
\centering
\includegraphics[width=0.95\textwidth]{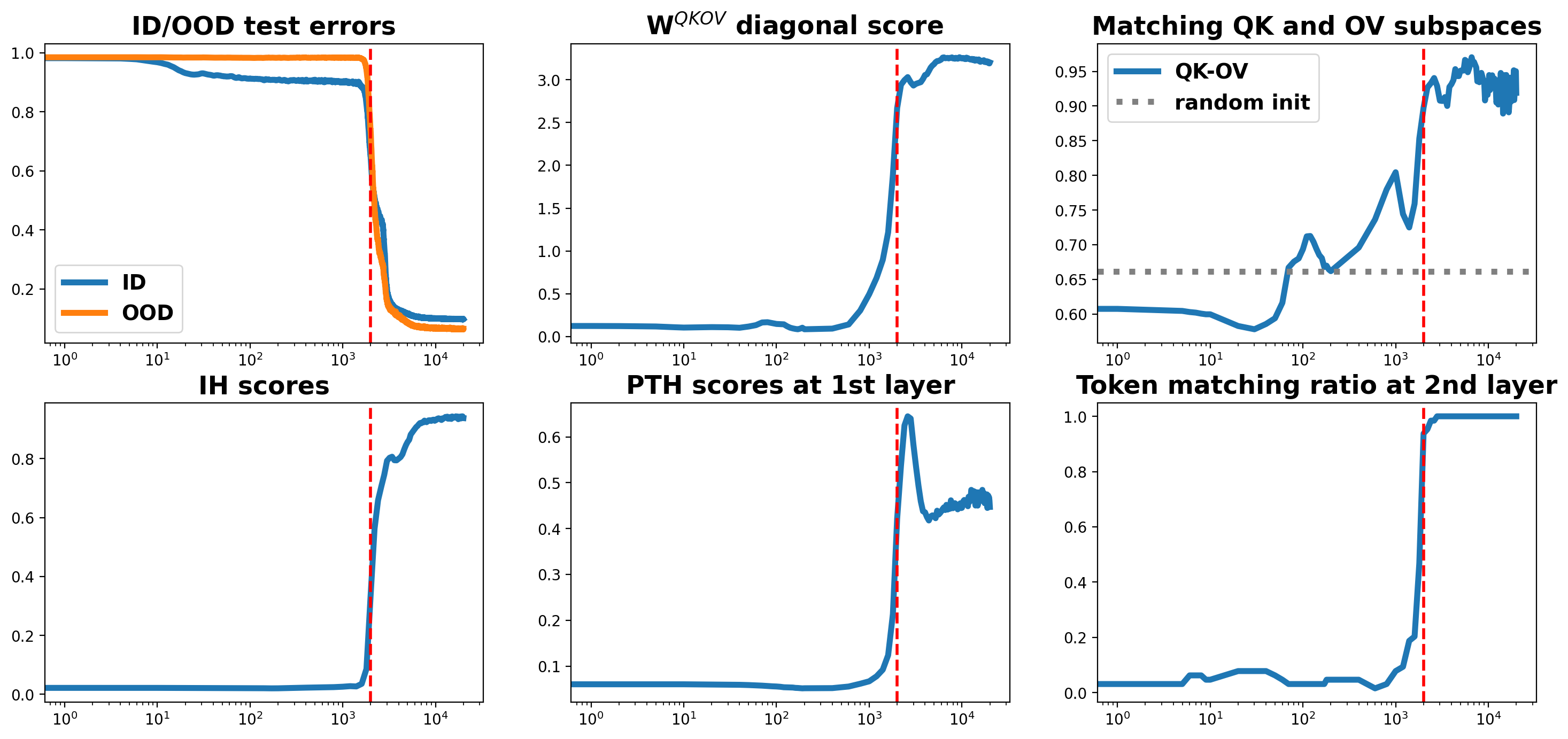}
\caption{Progress measures for synthetic data with a large pool size $S=1000$. Plots are similar to Figure~\ref{fig:progress}.
}\label{fig:progress-large}
\end{figure}

\begin{figure}[h!]
\centering
\includegraphics[width=0.95\textwidth]{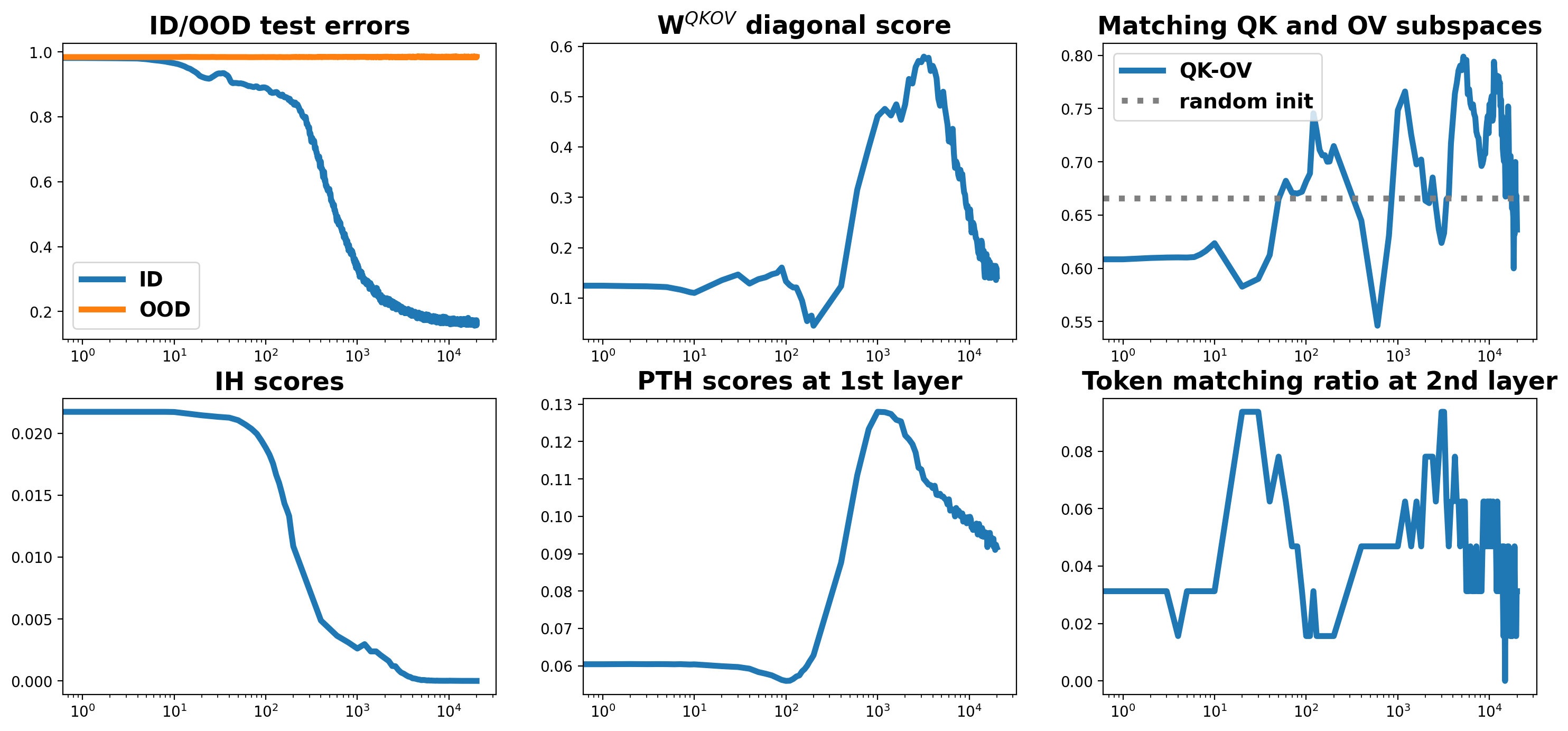}
\caption{Progress measures for synthetic data with a small pool size $S=100$. Plots are dissimilar to Figure~\ref{fig:progress}.
}\label{fig:progress-small}
\end{figure}

We consider the same model/training settings as in Section~\ref{sec:synthetic} but we choose two different pool sizes $S=1000$ and $S=100$. In Figure~\ref{fig:progress-large} and~\ref{fig:progress-small}, the progress measures under $S=1000$ are similar to Figure~\ref{fig:progress} but very different under $S=100$.

\subsection{Attention matrices}\label{sec:append-memorize}

\begin{figure}[h!]
\centering
\includegraphics[width=0.8\textwidth]{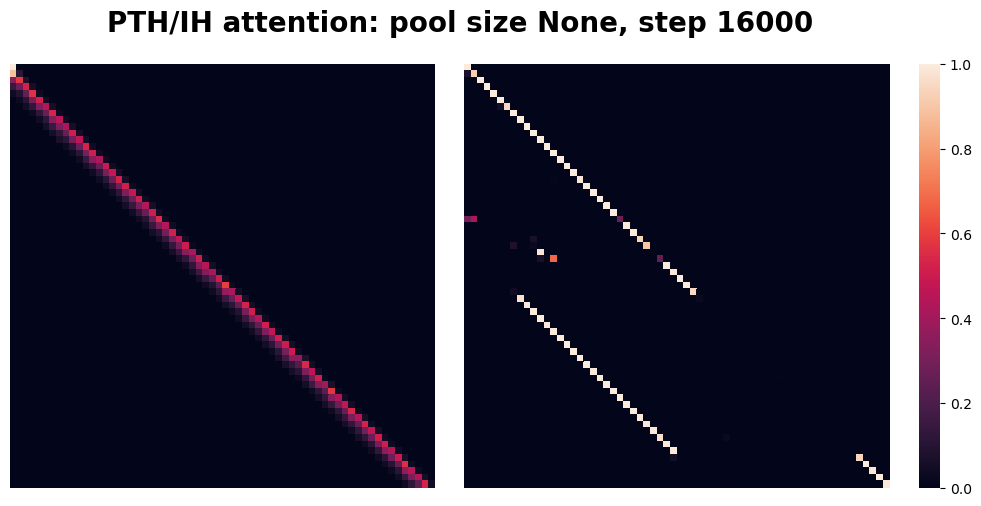}
\caption{Attention matrices of the generalizing model (pool size is 1000). As claimed, the PTH (left) largely attends to the previous position (diagonal line shifted down by 1), and IH (right) attends to to-be-copied positions of the repeated segment.
}\label{fig:heatmap-S-1000}
\end{figure}

\begin{figure}[h!]
\centering
\includegraphics[width=0.8\textwidth]{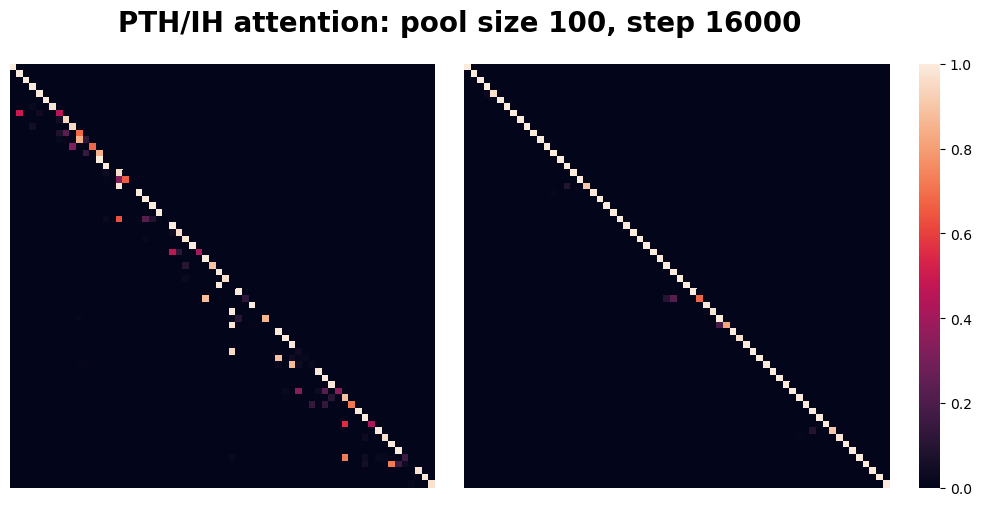}
\caption{Attention matrices of memorizing model (pool size is 100). Both PTH (left) and IH (right) exhibit diagonal lines, which do not reflect the repetition pattern in the OOD instance.
}\label{fig:heatmap-S-100}
\end{figure}

The heatmaps in Figure~\ref{fig:heatmap} indicate that the memorizing model ($S=100$) are qualitatively different from that generalizing model ($S=1000$). In Figure~\ref{fig:heatmap-S-1000} and~\ref{fig:heatmap-S-100}, we further plot attention matrices $\mA$ based on one OOD sequence at training step 16K to support our claim that the memorizing model does not learn compositional structure for OOD generalization.

%% file: src/appendix/synthetic-rank.tex
\subsection{Measurements for subspace matching}\label{synthetic-rank}

We explore two factors that may impact the subspace matching score. Recall that we defined the score as
\begin{equation*}
    \sigma_{\max}(\mU^\top \mV)
\end{equation*}
where $\mU, \mV \in \R^{d \times r}$ are the principal subspaces of $\mW_{\QK}$ and $\mW_{\OV}$. We consider alternative measurements.
\begin{itemize}
    \item Replace the largest singular value by an average quantity $\big(r^{-1} \sum_{j\le r}\sigma^2_{j}((\mU^\top \mV)\big)^{1/2}$ where $\sigma_j$ denotes the $j$-th largest singular value.
    \item Vary the rank parameter $r$.
\end{itemize}

First, the average score reflects the alignment of two subspaces by picking two vectors randomly from the subspaces.  Second, we investigate whether the rank parameter $r$ has impact on the subspace matching measurement on the synthetic example. We consider the same setting as in Section~\ref{sec:synthetic} but use different $r \in \{5, 10, 15, 20\}$ when measuring the subspace matching scores.

\begin{figure}[h!]
\centering
\includegraphics[width=0.5\textwidth]{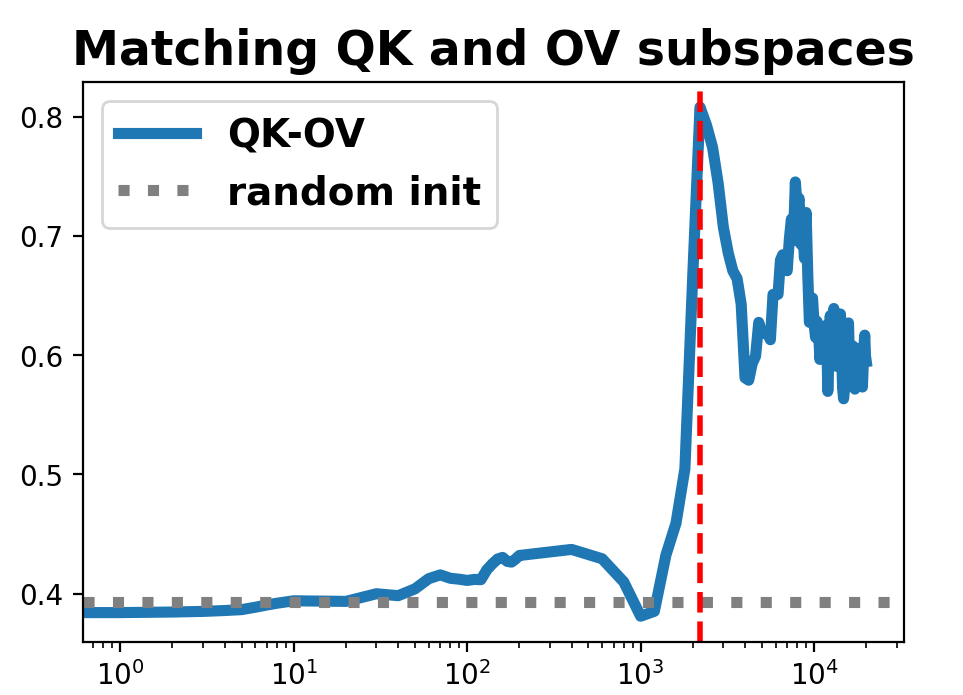}
\caption{Subspace matching on the synthetic example with an average score.
}\label{fig:progress-mean}
\end{figure}

\begin{figure}
     \centering
     \begin{subfigure}[b]{0.22\textwidth}
         \centering
         \includegraphics[width=\textwidth]{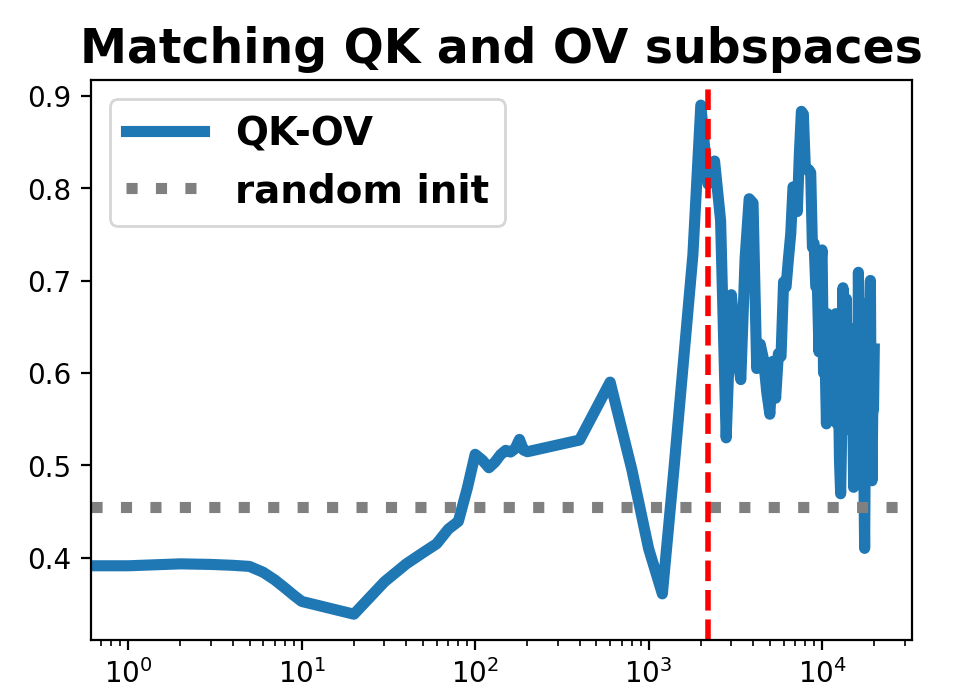}
     \end{subfigure}
     \hfill
     \begin{subfigure}[b]{0.22\textwidth}
         \centering
         \includegraphics[width=\textwidth]{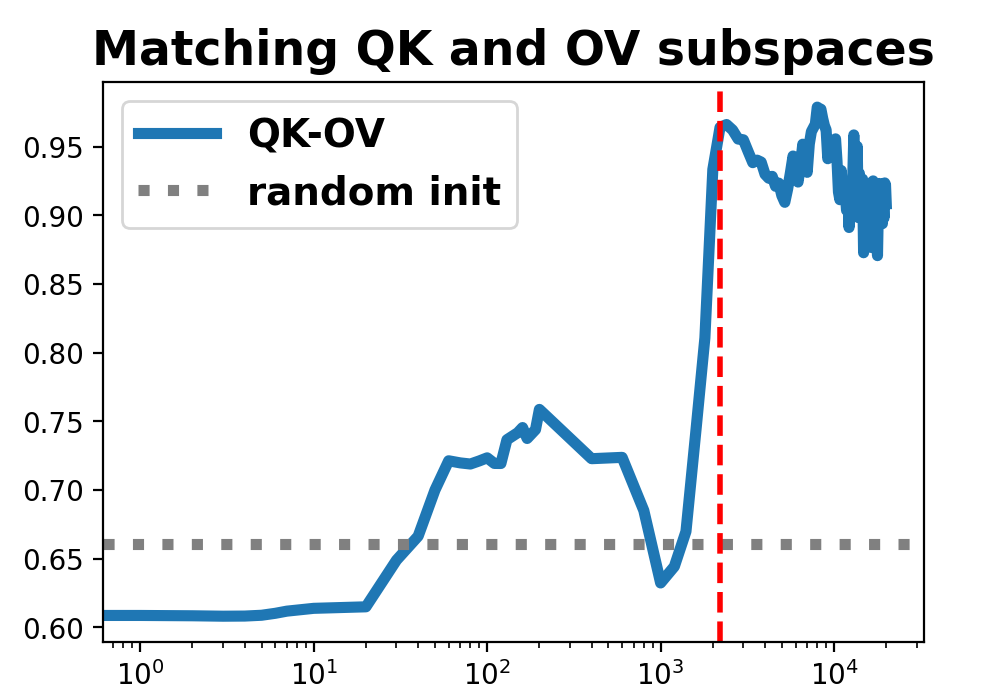}
     \end{subfigure}
     \hfill
     \begin{subfigure}[b]{0.22\textwidth}
         \centering
         \includegraphics[width=\textwidth]{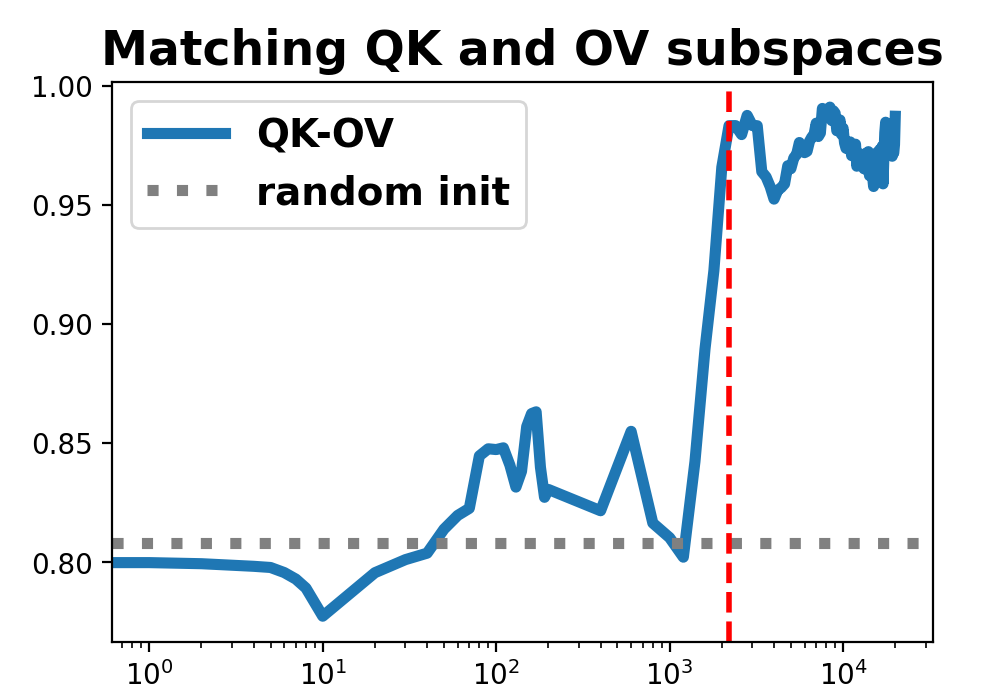}
     \end{subfigure}
          \hfill
     \begin{subfigure}[b]{0.22\textwidth}
         \centering
         \includegraphics[width=\textwidth]{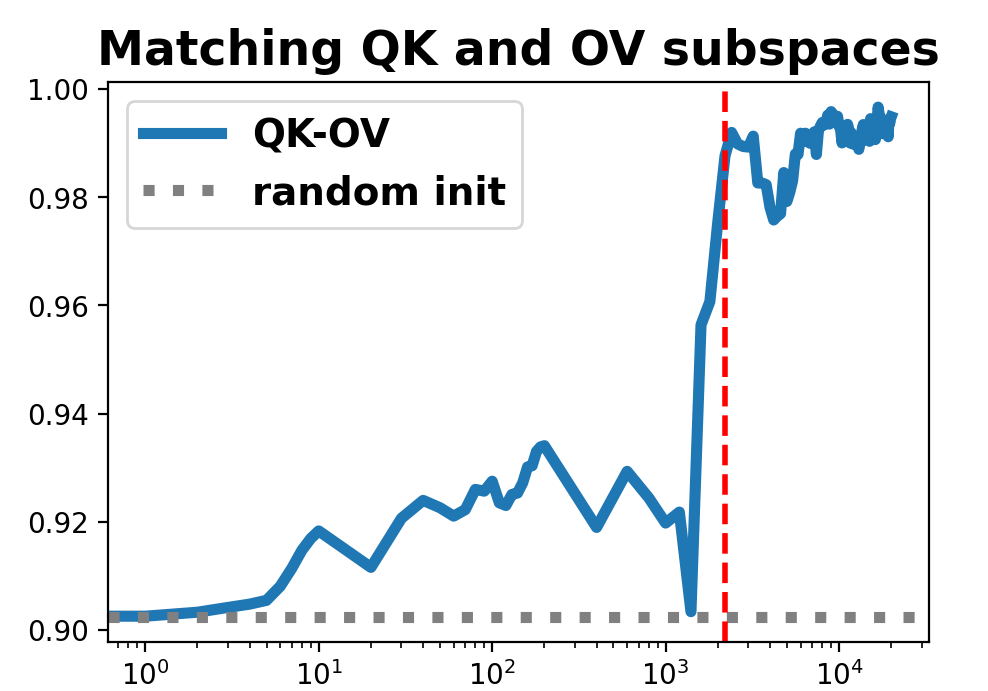}
     \end{subfigure}
        \caption{Subspace matching on the synthetic example with different rank $r \in \{5, 10, 15, 20\}$.}
        \label{fig:synthetic-rank}
\end{figure}

We find that the two alternative measurement yield qualitatively similar results; see Figure~\ref{fig:progress-mean} and~\ref{fig:synthetic-rank}.

%% file: src/appendix/synthetic-rope.tex
\subsection{Rotary positional embedding}\label{sec:append-synthetic-rope}

As we explained in Section~\ref{sec:append-model-variants}, there is a nuance with rotary positional embedding (RoPE) in the attention calculation. To calculate the attention in a head, we need a set of QK matrices $(\mW_{QK, \Delta t})_{\Delta t \ge 0}$ where
\begin{equation*}
    \mW_{\QK, \Delta t} = \mW^q \mR_{\Delta t} (\mW^k)^\top 
\end{equation*}
The matrix $\mZ \in \R^{T \times T}$ before softmax operation is given by $Z_{t, t'} = \vx_{t}^\top \mW_{\QK, |t-t'|} \vx_{t'}$. In particular, if $\Delta t = 0$, then $\mR_{0} = \mI_d$ and thus $\mW_{\QK, 0} = \mW^q (\mW^k)^\top$. We used exactly $\mW_{\QK, 0}$ for Figure~\ref{fig:progress}.

\begin{figure}[ht!]
\centering
\includegraphics[width=0.95\textwidth]{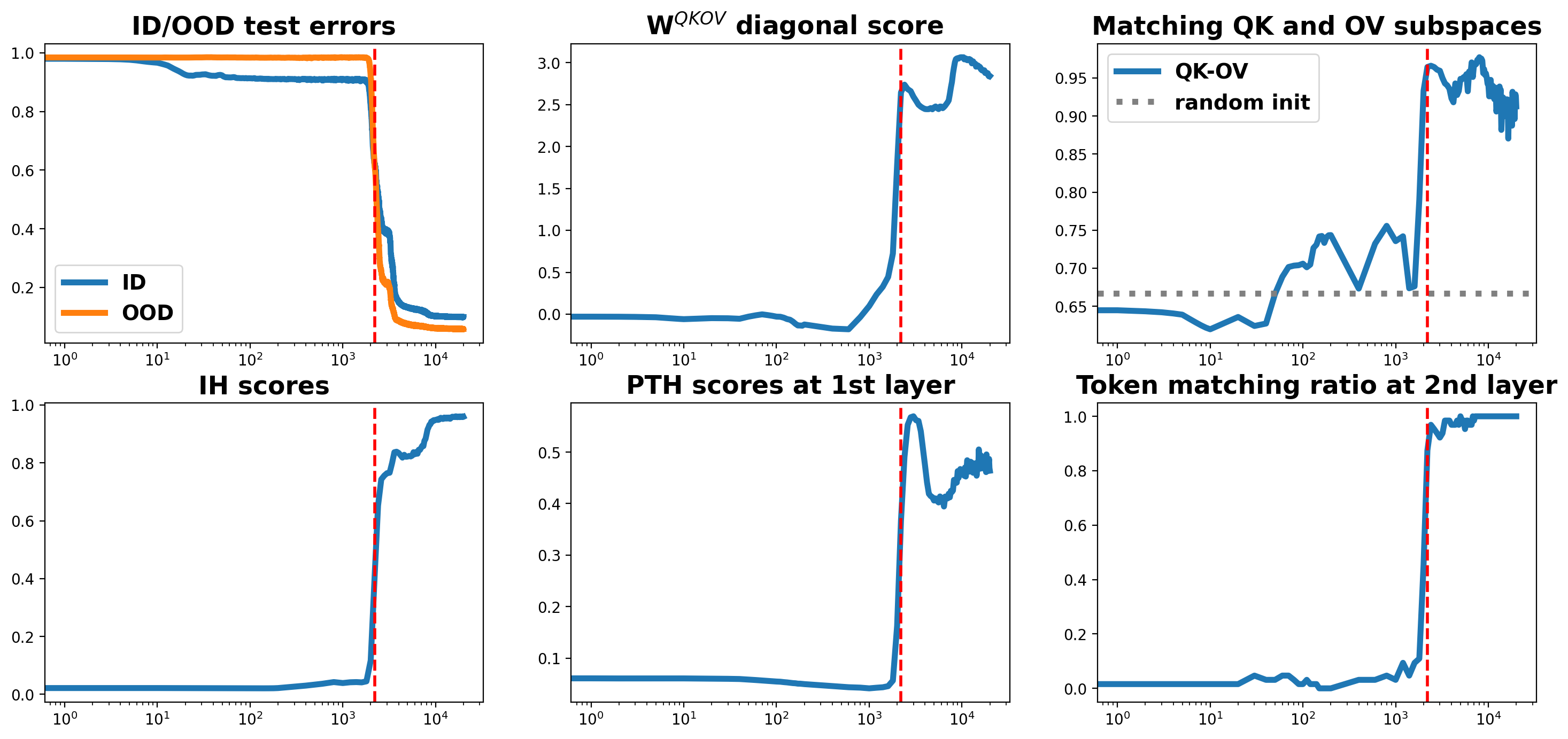}
\caption{Progress measures using $\mW_{\QK, \Delta t}$ where $\Delta t = 5$.
}\label{fig:progress-delta-5}
\end{figure}

\begin{figure}[ht!]
\centering
\includegraphics[width=0.95\textwidth]{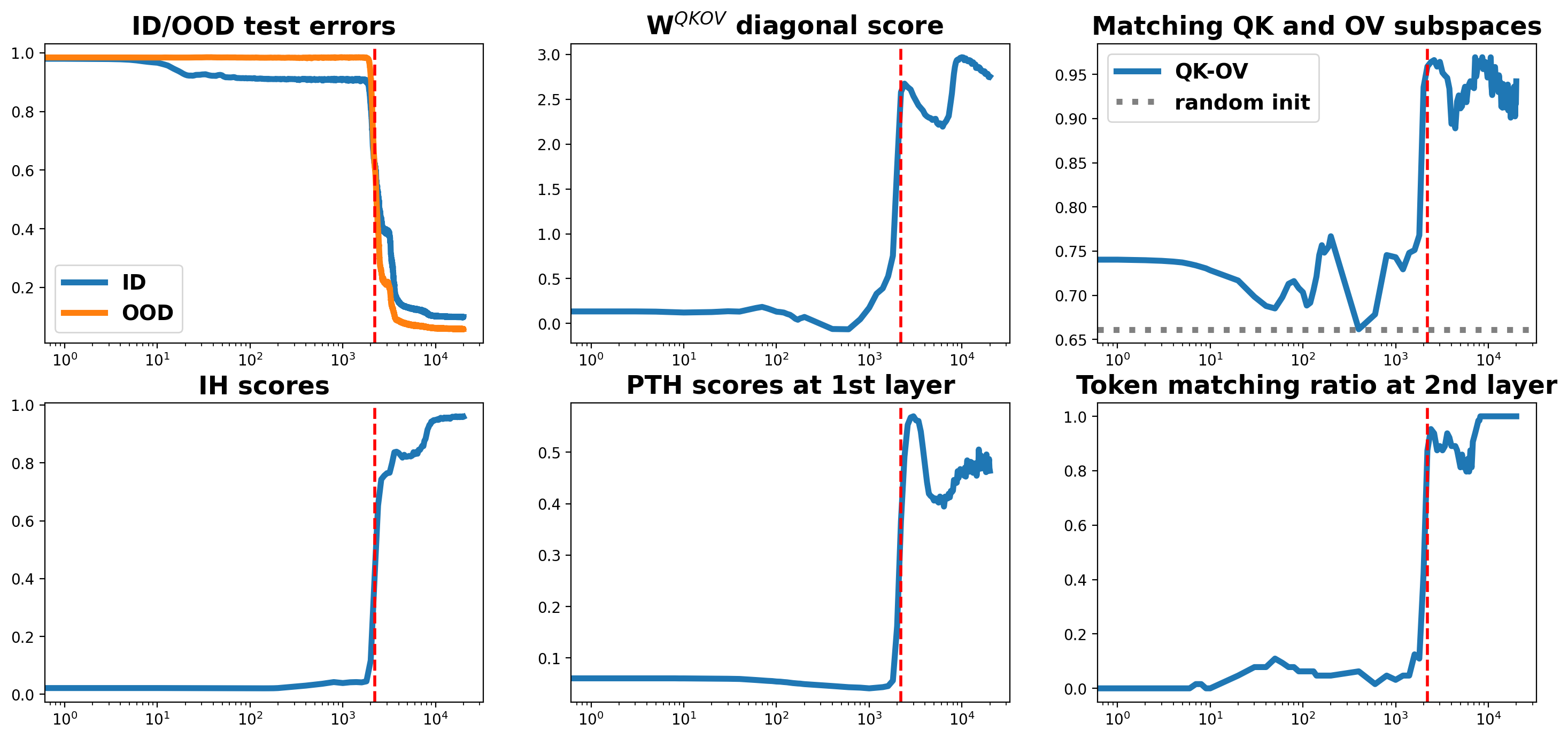}
\caption{Progress measures using $\mW_{\QK, \Delta t}$ where $\Delta t = 10$.
}\label{fig:progress-delta-10}
\end{figure}

\begin{figure}[ht!]
\centering
\includegraphics[width=0.95\textwidth]{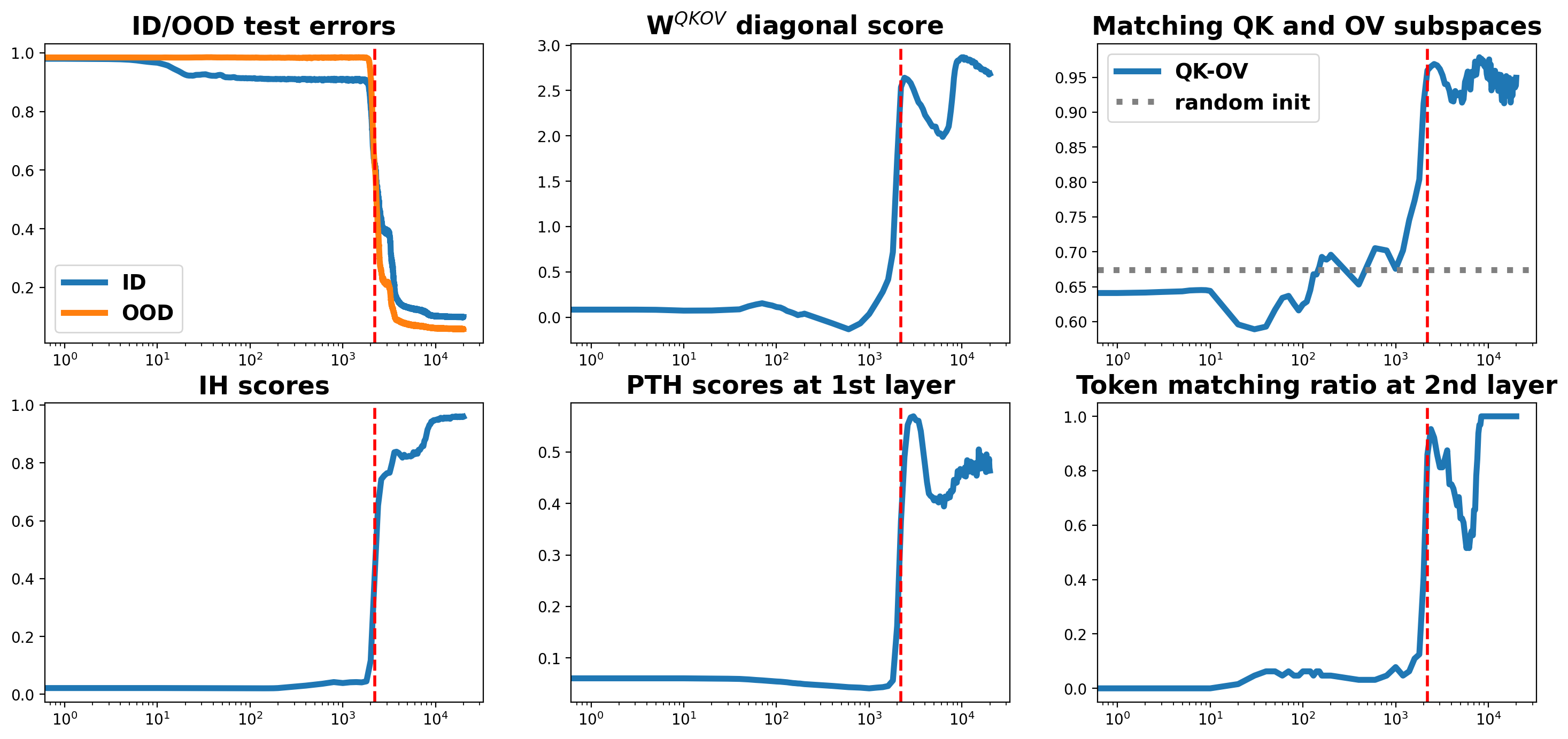}
\caption{Progress measures using $\mW_{\QK, \Delta t}$ where $\Delta t = 15$.
}\label{fig:progress-delta-15}
\end{figure}

\begin{figure}[ht!]
\centering
\includegraphics[width=0.95\textwidth]{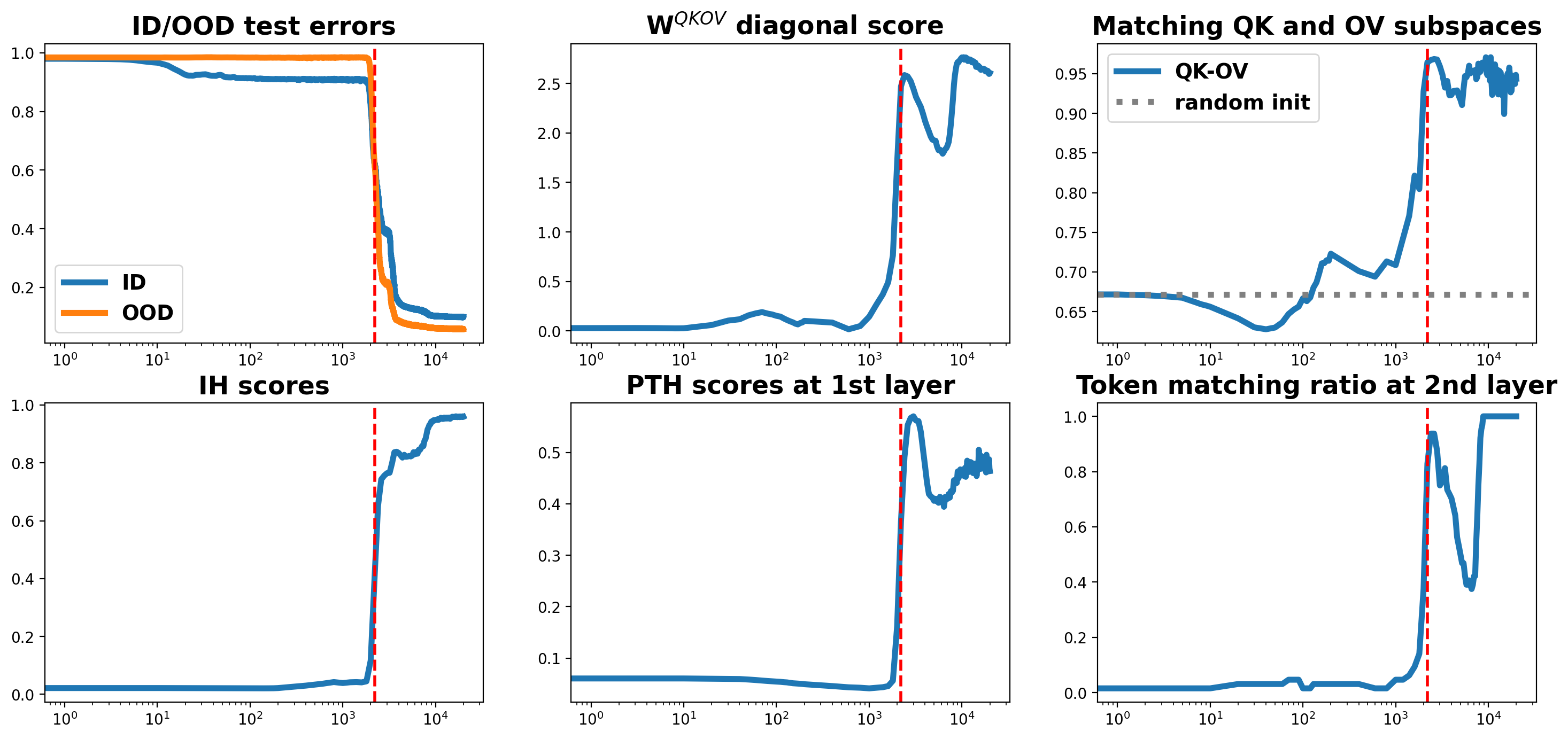}
\caption{Progress measures using $\mW_{\QK, \Delta t}$ where $\Delta t = 20$.
}\label{fig:progress-delta-20}
\end{figure}

We show that $\mW_{\QK, \Delta t}$ with nonzero $\Delta t$ yields similar results. This is demonstrated in Figure~\ref{fig:progress-delta-5}--\ref{fig:progress-delta-20} where $\Delta t \in \{5, 10, 15, 20\}$. Note that $\Delta t$ only affects the 2nd, 3rd, and 6th subplots in each figure.

%% file: src/appendix/spike.tex
\subsection{Connection to spiked matrices}\label{sec:append-spike}

We notice that common bridge subspace is connected to the statistical literature on spiked matrices \cite{spiked}.
Consider an ideal scenario, where the weight matrices are spiked with a shared spiked 
\begin{equation*}
    \mW_{OV}^j = \mV \mU^\top + \text{noise}, \qquad \mW_{QK}^j = \mU \mV^\top + \text{noise},
\end{equation*}
where $\mU, \mV \in \R^{d \times r}$ are orthonormal matrices and the \text{noise} matrices have much smaller magnitude. 
The observed properties are realized by this ideal case. Indeed, the QK and OV circuits are all matched through a common subspace $\mathrm{span}(\mV)$, QK circuits (or OV circuits) play exchangeable roles, and compositional information goes through $\mathrm{span}(\mV)$. Moreover, $\mW_{QK, j} \mW_{OV, j} \approx \mU \mU^\top$, so diagonal entries are generally larger for $\mU$ in a generic position.

%% file: src/appendix/implementation.tex
\subsection{Models}\label{sec:append-models}

We list the models in the experiments. 
\begin{itemize}
    \item \textbf{GPT2:} 12 layers, 12 attention heads, and a hidden size of 768.
    \item \textbf{GPT2-XL:} 48 layers, 25 attention heads, and a hidden size of 1600.
    \item \textbf{Llama2-7B:} 32 layers, 32 attention heads, and a hidden size of 4096.
    \item \textbf{Gemma-7B:} 36 layers, 32 attention heads, and a hidden size of 3072.
    \item \textbf{Gemma2-9B:} 28 layers, 16 attention heads, and a hidden size of 3072.
    \item \textbf{Falcon-7B:} 32 layers, 32 attention heads, and a hidden size of 4544.
    \item \textbf{Falcon2-11B:} 
    60 layers, 32 attention heads, and a hidden size of 4096. 
    \item \textbf{Mistral-7B:} 32 layers, 32 attention heads, and a hidden size of 4096.
    \item \textbf{Olmo-7B:} 32 layers, 32 attention heads, and a hidden size of 4096.
    \item \textbf{Llama3-8B:} 32 layers, 32 attention heads, and a hidden size of 4096.
    \item \textbf{Pythia-7B:} 32 layers, 32 attention heads, and a hidden size of 4096.
    \item \textbf{Llama2-70B:} 80 layers, 64 attention heads, and a hidden size of 8192.
    \item \textbf{Llama3-70B:} 80 layers, 64 attention heads, and a hidden size of 8192.
\end{itemize}
All models are downloadable from Huggingface.

\subsection{Tasks}\label{sec:append-task-details}

We provide details including setup, measurement and prompt examples for tasks introduced in our main paper: Fuzzy Copying, Indirect Object Identification (IOI), In-Context Learning (ICL), GSM8K, and GSM8K-rand.

\paragraph{Fuzzy copying} The task is evaluated on 100 samples, each containing 10 in-context examples. For each sample, the result is determined by whether the model correctly predicts the next two words. Two metrics are used for evaluation in IOI: accuracy and probability. For accuracy, each sample is evaluated based on whether the model correctly predict all the tokens in the answer. The mean accuracy across all samples is reported. For probability, we calculate the cumulative probability of the model predicting the correct tokens for each sample. The mean probability across all samples is then reported.

Here is an example prompt for the Fuzzy Copying task.

\begin{lstlisting}[backgroundcolor=\color{lightgray}, frame=single, rulecolor=\color{black}]
bear snake fox poppy plate butterfly caterpillar boy 
aquarium_fish motorcycle BEAR SNAKE FOX POPPY PLATE BUTTERFLY CATERPILLAR BOY
\end{lstlisting}

\paragraph{IOI} The task is evaluated on 100 samples. Two metrics are used for evaluation in IOI: accuracy and probability. For accuracy, each sample is evaluated based on whether the model assigns a higher probability to the correct name among the two options. The mean accuracy across all samples is reported. For probability, we calculate the cumulative probability of the model predicting the correct tokens for each sample. The mean probability across all samples is then reported.

Here are example prompts for the IOI task, presented in both the English and symbolized versions.

\begin{lstlisting}[backgroundcolor=\color{lightgray}, frame=single, rulecolor=\color{black}]
English Version
-------------------
Then, Anna and Matthew went to the station. Anna gave a basketball to

Symbolized Version
-------------------
Then, &^ and #$ went to the station. &^ gave a basketball to
    
\end{lstlisting}

\paragraph{ICL}
The task is evaluated on 100 samples, each containing 20 in-context examples. For each sample, the result is determined by whether the model correctly predicts the next category. Two metrics are used for evaluation in IOI: accuracy and probability. For accuracy, each sample is evaluated based on whether the model correctly predict all the tokens in the answer. The mean accuracy across all samples is reported. For probability, we calculate the cumulative probability of the model predicting the correct tokens for each sample. The mean probability across all samples is then reported.

Here are example prompts for the ICL task, presented in both the English and symbolized versions.

\begin{lstlisting}[backgroundcolor=\color{lightgray}, frame=single, rulecolor=\color{black}]
English Version
-------------------
baseball is sport, celery is plant, sheep is animal, 
volleyball is sport, rugby is sport, cycling is sport, 
camel is animal, llama is animal, hockey is sport, panda is animal, 
football is sport, onions is plant, cucumber is plant, zucchini is plant, 
zebra is animal, billiards is sport, golf is sport, horse is animal, 
kale is plant, volleyball is sport, lettuce is

Symbolized Version
-------------------
baseball is $#, celery is !%, sheep is &*, volleyball is $#, 
rugby is $#, cycling is $#, camel is &*, llama is &*, hockey is $#, 
panda is &*, football is $#, onions is !%, cucumber is !%, zucchini is !%, 
zebra is &*, billiards is $#, golf is $#, horse is &*, kale is !%, 
volleyball is $#, lettuce is
\end{lstlisting}

\paragraph{GSM8K} The task is evaluated on 100 samples, each containing 10 CoT examples of GSM questions. For each sample, the answer is extracted using regular expression matching for the pattern \#\#\#\# \{solution\}, and the result is determined by whether the model's solution exactly matches the extracted answer. We report the mean accuracy across all samples. Below is an example prompt. For demonstration purposes, we include only two in-context examples.

\begin{lstlisting}[backgroundcolor=\color{lightgray}, frame=single, rulecolor=\color{black}]
Question: Jack has a stack of books that is 12 inches thick. He knows from experience that 80 pages is one inch thick. If he has 6 books, how many pages is each one on average?
Answer: There are 960 pages because 80 x 12 = <<80*12=960>>960
Each book is 160 pages because 960 / 6 = <<960/6=160>>160
#### 160

Question: Joseph invested $1000 into a hedge fund. The fund promised a yearly interest rate of 10%. If he deposited an additional $100 every month into the account to add to his initial investment of $1000, how much money will he have in the fund after two years?
Answer: For the first year, Joseph will have invested $1000 + ($100 * 12) = $<<1000+100*12=2200>>2200.
The interest calculated for the first year will be $2200 * 10% = $<<2200*10*.01=220>>220.
The total value of the investment for the first year will be $2200 + $220 = $<<2200+220=2420>>2420.
For the second year, the total invested will be $2420 + ($100 * 12) = $<<2420+100*12=3620>>3620.
The interest calculated after the second year will be $3620 * 10% = $<<3620*10*.01=362>>362.
Therefore, Joseph's investment in the mutual fund will be worth $3620 + $362 = $<<3620+362=3982>>3982.
#### 3982

<<<<< EIGHT MORE IN-CONTEXT QUESTION-ANSWER PAIRS >>>>>

Question: Craig has 2 twenty dollar bills. He buys six squirt guns for $2 each.  He also buys 3 packs of water balloons for $3 each.  How much money does he have left?

\end{lstlisting}

\paragraph{GSM8K-rand} To construct the GSM8K-rand dataset, we sample 12 questions from GSM8K and use them to create 12 templates. Specifically, for each question, a template is generated by replacing numbers and names in the problem statement and CoT deduction process with references. These references are drawn from a pool of 40 symbols, making the problems unlikely to appear in the training data. Additionally, the sampled questions are deliberately chosen from those correctly predicted by the Llama2-7B model. This selection helps establish a higher baseline and allows for a clearer observation of accuracy reduction during the intervention.

Regarding the accuracy evaluation, each prompt consists of 10 unique template realizations followed by one question. All other details remain the same as those used for GSM8K. Below is an example prompt. For demonstration purposes, we include only two in-context examples.

\begin{lstlisting}[backgroundcolor=\color{lightgray}, frame=single, rulecolor=\color{black}]
Question: A boy has 15 #$. His brother has 2 fewer #$ than he has. How many #$ do they have together?
Answer:
His brother has 15 - 2 = <<15-2=13>>13 #$.
Together, they have 15 + 13 = <<15+13=28>>28 #$.
#### 28

Question: There is space for 12 &! in the box. If there are 6 &! missing from the box, how many pairs of &! are in the box?
Answer:
In the box there are 12 &! - 6 &! = <<12-6=6>>6 &!.
Dividing into pairs we have 6 &! / 2 &!/pair = 3 pairs of &!
#### 3

<<<<<  EIGHT MORE IN-CONTEXT QUESTION-ANSWER PAIRS >>>>>

Question: &* is cutting up wood for his wood-burning stove. Each $@ tree makes 90 logs, each @$ tree makes 63 logs, and each @^ tree makes 70 logs. If &* cuts up 7 $@ trees, 4 @$ trees, and 4 @^ trees, how many logs does he get?

\end{lstlisting}

%% file: src/appendix/induction_head_details.tex
\subsection{Details about PTHs and IHs in LLMs}\label{sec:append-list}

We find that PTHs and IHs are distributed across many layers. In particular, Figure~\ref{fig:CBRH}(c)(d) shows that the top-10 PTHs and IHs are strongly aligned compared with random pairs.

For the shuffling experiments and projection experiments in Section~\ref{sec:llm}, we selected PTHs and IHs from top-scoring attention heads. Here we detail the procedure and [layer, head] pairs for each model.

\paragraph{Selecting relevant IHs and PTHs.} Here we show the list of PTHs and IHs for each models in Table~\ref{tab:ih-pth-shuffle-proj}.


The screening process for IHs and PTHs in the shuffling and projection experiments in Section~\ref{sec:cbr} is as follows. For each LLM, we first sort the IH/PTH pairs by their diagonal score and then filter out pairs with scores below the cutoff of $\delta=2.3$,
$$
\mathrm{diagonal}(\textrm{IH}_1, \textrm{PTH}_1) \ge \mathrm{diagonal}(\textrm{IH}_2, \textrm{PTH}_2)\ge \ldots \ge \mathrm{diagonal}(\textrm{IH}_{K'}, \textrm{PTH}_{K'}) \ge \delta
$$
We obtain an ordered list $\textrm{IH}_\delta$ as the deduplicated IH heads appearing in the top-$K'$ pairs above. Similarly, the ordered list $\textrm{PTH}_\delta$ contains the deduplicated PTH heads appearing in the top-$K'$ pairs above.
To avoid too many heads, we select at most 10 heads from $\textrm{PTH}_\delta$ and $\textrm{IH}_\delta$.



\begin{table}[ht]
\centering
\begin{tabular}{>{\small}p{6em} >{\small}p{5cm} >{\small}p{5cm} }
\toprule
\textbf{Model} & \textbf{IH} & \textbf{PTH} \\
\midrule
\textbf{GPT2} &
[[5, 1], [6, 9], [5, 5], [7, 2], [7, 10], [5, 8], [5, 0], [8, 1], [7, 11], [9, 6]] 
& [[4, 11], [5, 6], [8, 7], [6, 8], [6, 0], [9, 3], [3, 3], [7, 0], [5, 2], [1, 0]]  \\
\midrule

\textbf{GPT2-XL} &
[[17, 6], [16, 21], [16, 3], [13, 0], [18, 0], [17, 14], [20, 0], [19, 18], [22, 20], [21, 3]] &
[[15, 19], [12, 21], [13, 20], [14, 12], [16, 5], [11, 2], [9, 7], [14, 20], [10, 15], [13, 12]] \\
\midrule

\textbf{Llama2-7B} & [[6, 9], [6, 30], [7, 4], [8, 26], [7, 12], [7, 13], [6, 11], [8, 31], [6, 16], [11, 15]] &
[[5, 15], [6, 5], [10, 3], [15, 11]] \\
\midrule
\textbf{Gemma-7B} & [[5, 0], [14, 15], [20, 1], [20, 13], [18, 13], [21, 1], [16, 1]] & [[3, 9], [13, 3], [19, 13], [3, 15], [2, 4], [17, 0], [1, 0], [13, 2], [15, 3]] \\
\midrule
\textbf{Gemma2-9B} & [[28, 6], [17, 5], [7, 1], [25, 13], [28, 2], [11, 2], [15, 2], [5, 1], [15, 3], [34, 14]] &
[[27, 2], [16, 0], [6, 9], [24, 2], [10, 4], [14, 2], [10, 1], [3, 14], [3, 15], [14, 11]] \\
\midrule
\textbf{Falcon-7B} & [[5, 65], [5, 18], [5, 13], [5, 10], [5, 2], [5, 1], [5, 69], [5, 43], [5, 41], [5, 14]] & [[3, 38]] \\
\midrule
\textbf{Mistral-7B} & [[12, 6], [12, 4], [12, 7], [18, 2], [18, 1], [18, 3]] & [[11, 17], [17, 22]] \\
\midrule
\textbf{Olmo-7B} & [[27, 14], [15, 15], [24, 7], [2, 28], [2, 10], [26, 17]] & [[26, 25], [14, 30]], [14, 18], [23, 24], [1, 7], [24, 3], [25, 6], [1, 15], [26, 30]] \\
\midrule
\textbf{Llama3-8B} & [[15, 28], [15, 30], [8, 1], [15, 1], [5, 11], [5, 8], [5, 10], [5, 9], [16, 20], [16, 23]] &
[[14, 26], [7, 2], [4, 13], [7, 1], [4, 12], [1, 20], [9, 11], [25, 20]] \\
\midrule
\textbf{Pythia-7B} & [[7, 26], [7, 2], [7, 1], [6, 30], [8, 11], [8, 17], [4, 18], [8, 4], [6, 13], [7, 20]] &
[[6, 9], [5, 10], [3, 7], [3, 23], [14, 3], [6, 23], [11, 19]] \\
\bottomrule
\end{tabular}
\caption{Table of selected IHs/PTHs in the shuffling experiments and projection experiments}
\label{tab:ih-pth-shuffle-proj}
\end{table}

%% file: src/appendix/cbrh.tex
\subsection{Shuffle intervention experiment}\label{sec:append-shuffle}

We reported in Figure~5(e) the probability of predicting correct tokens for the copying task under two edits. We detail the calculations used in the figure.

\paragraph{Evaluation.} First, we uniformly sample tokens from the vocabulary $\gA$ (i.e., $|\gA| = 50257$) $L$ times, forming the segment $\vs^\#$. Then we repeat this segment using two replicas, yielding the input sequence $(\vs^\#, \vs^\#, \vs^\#)$. We repeat this $N_0 = 50$ times and obtain a batch of input sequences. 

Then, using the edited models as well as the original model, we calculate the probability of predicting the correct token
\begin{equation}\label{eq:aveprob}
        p_{2L+t} \big( s_t^\# \big| (\vs^\#, \vs^\#, \vs_{<t} \big).
\end{equation}
For GPT-2, we gather the above probability for $t\in\{6,7,8,\ldots,L\}$ and every input sequence. Finally, in the left plot of Figure~5(e) in our main paper, we compare the histogram of the prediction probabilities of the original model, the edited model with shuffling, and the edited random baseline. 

We also summarize the result of each LLM by averaging the probability in Eq.~\ref{eq:aveprob} over $i\in\{6,7,8,\ldots,L\}$ and input sequences. We obtained three averaged probabilities: the original LLM, the edited model with shuffling, and the edited random baseline with random replacement. Then we measure how much probability is reduced from the original probability to the two edited models. Finally we report the result in the right plot of Figure~5(e).

\subsection{Projection intervention experiment}\label{sec:append-project}

\paragraph{Calculating bridge subspace.} As mentioned in our main paper, we calculated the bridge subspace of rank $r$ based on the top-$K$ $\mW_{\QK}$ from the list of IHs.
We chose to edit $25\%$ of all heads using the projection matrices $\mV \mV^\top$ or $\mI_d - \mV \mV^\top$ because we believe that $25\%$ of all heads are likely to be relevant to the copying task. These heads are selected as the $25\%$ top-ranking heads according to the IH attention score.

\paragraph{Evaluation.}

Similar to the shuffling experiment, we use $N_0=50$ input sequences $(\vs^\#, \vs^\#, \vs^\#)$ where the segment $\vs^\#$ consists of uniformly sampled tokens. Here we consider the prediction accuracy. Namely, we use $0$-$1$ loss to measure whether a predicted token matches the target token
    \begin{equation*}
        \mathbf{1} \Big( \argmax_{s} p_{2L + t}(s \big| (\vs^\#, \vs^\#, \vs_{<t} \big) = s_t^\# \Big)
    \end{equation*}
For GPT2, we measure the change of the average prediction accuracy under projection edits $\mV \mV^\top$ and $\mI_d - \mV \mV^\top$ for varying rank parameter $r$, which we report in the left plot of Figure~5(f). 

\begin{table}
\centering
\begin{tabular}{c|c|c|c|c|c|c|c|c|c}
\footnotesize{GPT2} & \footnotesize{GPT2-XL} & \footnotesize{Gemma-7b} & \footnotesize{Gemma2-9b} & \footnotesize{Llama2-7b} & \footnotesize{Llama3-8b} & \footnotesize{Mistral-7b} & \footnotesize{Olmo-7b} & \footnotesize{Pythia-7b} & \footnotesize{Falcon-7b} \\ \hline
50 & 80 & 150 & 170 & 200 & 200 & 200 & 200 & 200 & 220
\end{tabular}\caption{The rank $r$ of $\mV$ used in the projection calculation.}\label{tab:append-rank}
\end{table}

We also measure each LLM and summarize the result by selecting a fixed rank parameter. The rank is selected to be $5\%$ of the model dimension $d$. Table~\ref{tab:append-rank} lists the value of $r$ for each LLM.

%% file: src/appendix/histogram.tex
\subsection{Histograms for PTH/IH matching}\label{sec:append-llm-match}

\begin{figure*}
        \centering
        \begin{subfigure}[b]{0.31\textwidth}
            \centering
            \includegraphics[width=\textwidth]{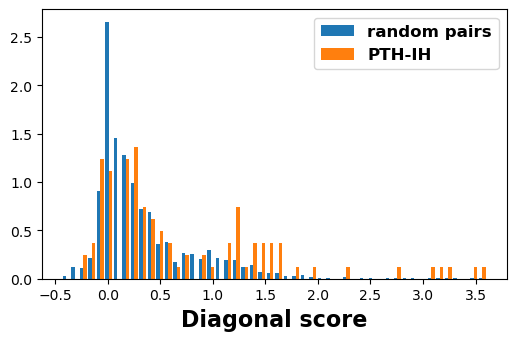}
            \caption%
            {{\small Falcon-7b}}    
        \end{subfigure}
        \hfill
        \begin{subfigure}[b]{0.31\textwidth}  
            \centering 
            \includegraphics[width=\textwidth]{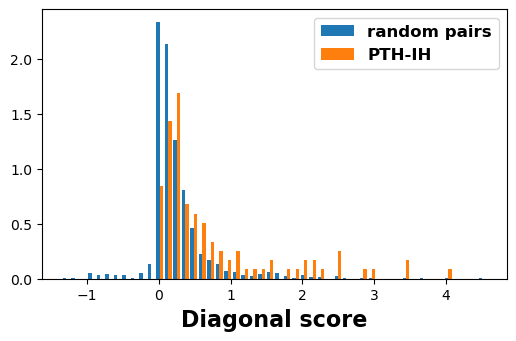}
            \caption[]%
            {{\small Gemma-7b}}    
        \end{subfigure}
        \hfill
        \begin{subfigure}[b]{0.31\textwidth}  
            \centering 
            \includegraphics[width=\textwidth]{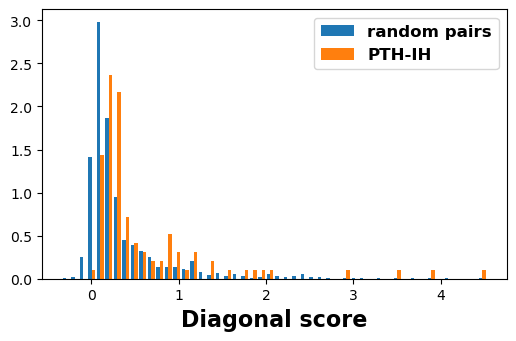}
            \caption[]%
            {{\small Gemma2-7b}}    
        \end{subfigure}
        \vskip\baselineskip
        \begin{subfigure}[b]{0.31\textwidth}  
            \centering 
            \includegraphics[width=\textwidth]{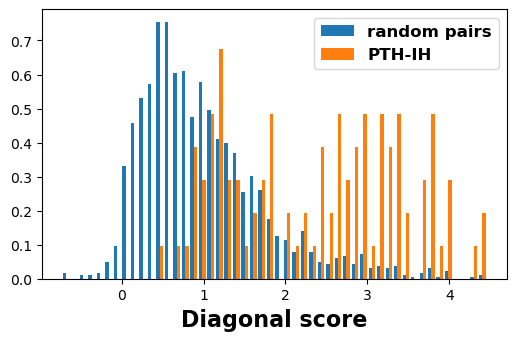}
            \caption[]%
            {{\small GPT2-XL}}    
        \end{subfigure}
        \hfill
        \begin{subfigure}[b]{0.31\textwidth}  
            \centering 
            \includegraphics[width=\textwidth]{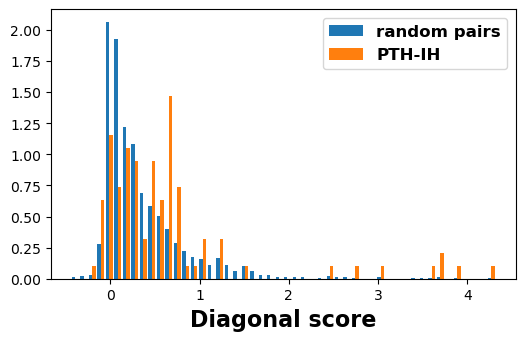}
            \caption[]%
            {{\small Llama2-7b}}    
        \end{subfigure}
        \hfill
        \begin{subfigure}[b]{0.31\textwidth}  
            \centering 
            \includegraphics[width=\textwidth]{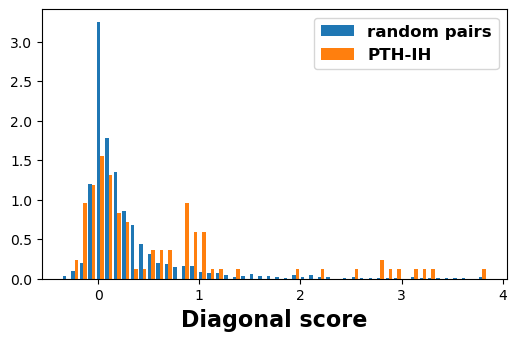}
            \caption[]%
            {{\small Llama3-8b}}    
        \end{subfigure}
        \vskip\baselineskip
        \begin{subfigure}[b]{0.31\textwidth}  
            \centering 
            \includegraphics[width=\textwidth]{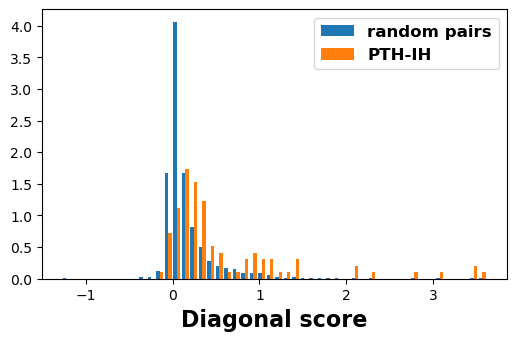}
            \caption[]%
            {{\small Mistral-7b}}    
        \end{subfigure}
        \hfill
        \begin{subfigure}[b]{0.31\textwidth}  
            \centering 
            \includegraphics[width=\textwidth]{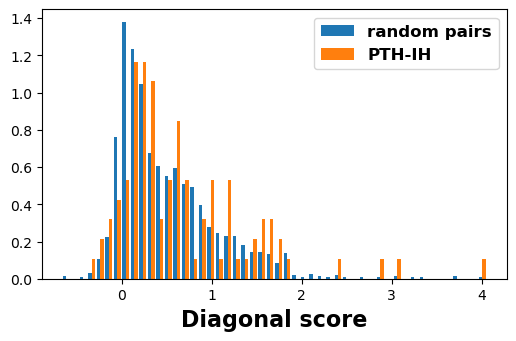}
            \caption[]%
            {{\small Olmo-7b}}    
        \end{subfigure}
        \hfill
        \begin{subfigure}[b]{0.31\textwidth}  
            \centering 
            \includegraphics[width=\textwidth]{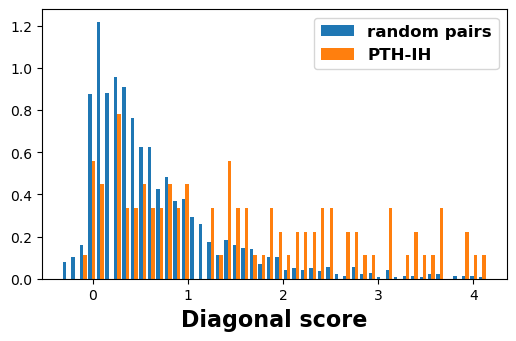}
            \caption[]%
            {{\small Pythia-7b}}    
        \end{subfigure}
        \caption{Diagonal scores: histograms showing matching of pairs from PTH/IH.} 
        \label{fig:pth-ih-matching}
    \end{figure*}

\begin{figure*}
        \centering
        \begin{subfigure}[b]{0.31\textwidth}
            \centering
            \includegraphics[width=\textwidth]{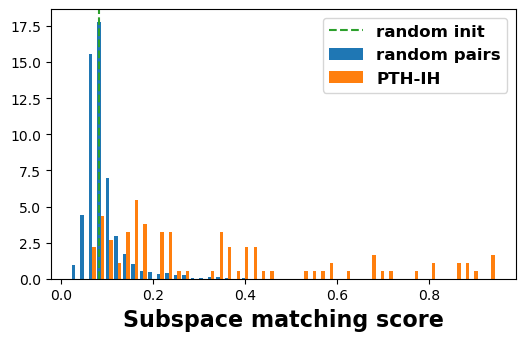}
            \caption%
            {{\small Falcon-7b}}    
        \end{subfigure}
        \hfill
        \begin{subfigure}[b]{0.31\textwidth}  
            \centering 
            \includegraphics[width=\textwidth]{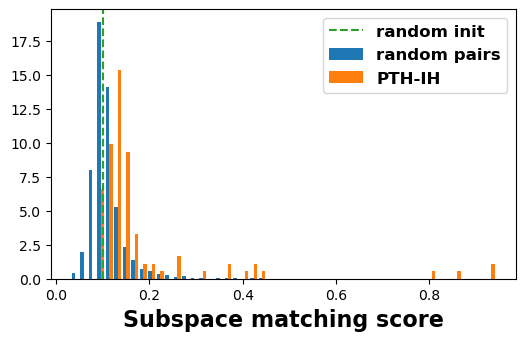}
            \caption[]%
            {{\small Gemma-7b}}    
        \end{subfigure}
        \hfill
        \begin{subfigure}[b]{0.31\textwidth}  
            \centering 
            \includegraphics[width=\textwidth]{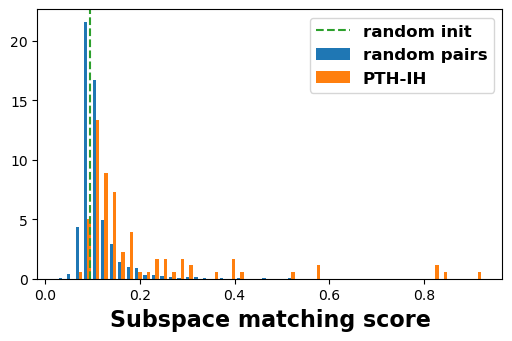}
            \caption[]%
            {{\small Gemma2-7b}}    
        \end{subfigure}
        \vskip\baselineskip
        \begin{subfigure}[b]{0.31\textwidth}  
            \centering 
            \includegraphics[width=\textwidth]{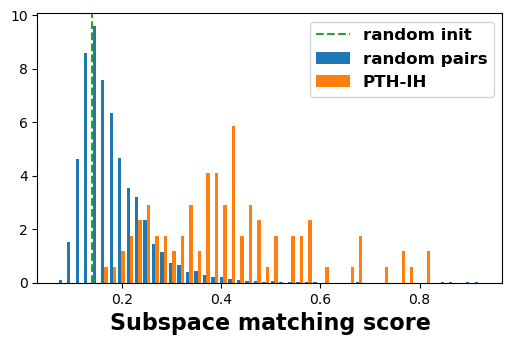}
            \caption[]%
            {{\small GPT2-XL}}    
        \end{subfigure}
        \hfill
        \begin{subfigure}[b]{0.31\textwidth}  
            \centering 
            \includegraphics[width=\textwidth]{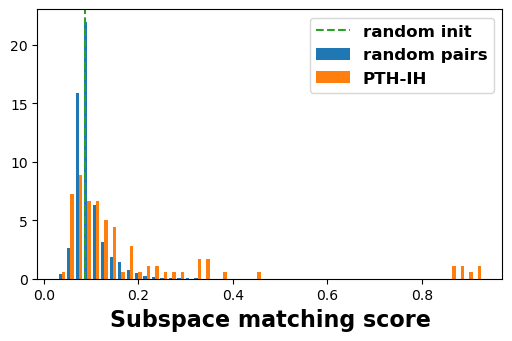}
            \caption[]%
            {{\small Llama2-7b}}    
        \end{subfigure}
        \hfill
        \begin{subfigure}[b]{0.31\textwidth}  
            \centering 
            \includegraphics[width=\textwidth]{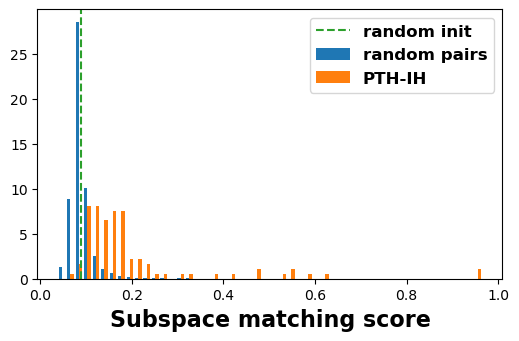}
            \caption[]%
            {{\small Llama3-8b}}    
        \end{subfigure}
        \vskip\baselineskip
        \begin{subfigure}[b]{0.31\textwidth}  
            \centering 
            \includegraphics[width=\textwidth]{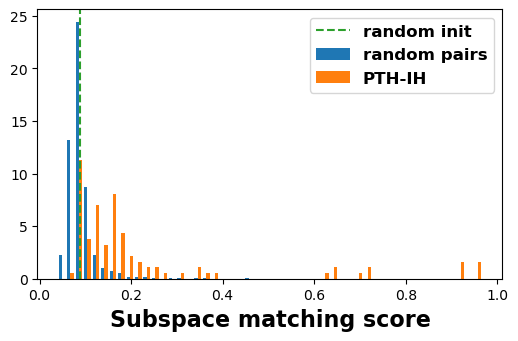}
            \caption[]%
            {{\small Mistral-7b}}    
        \end{subfigure}
        \hfill
        \begin{subfigure}[b]{0.31\textwidth}  
            \centering 
            \includegraphics[width=\textwidth]{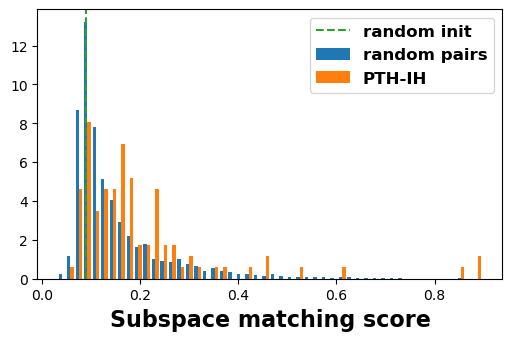}
            \caption[]%
            {{\small Olmo-7b}}    
        \end{subfigure}
        \hfill
        \begin{subfigure}[b]{0.31\textwidth}  
            \centering 
            \includegraphics[width=\textwidth]{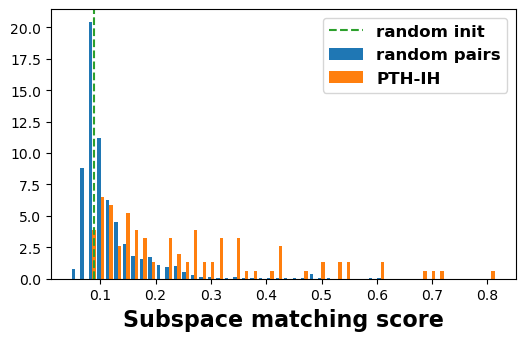}
            \caption[]%
            {{\small Pythia-7b}}    
        \end{subfigure}
        \caption{Subspace matching scores: histograms showing matching of pairs from PTH/IH.} 
        \label{fig:pth-ih-matching2}
    \end{figure*}

In Figure~\ref{fig:CBRH}(c)(d), we only presented the histogram plots based on GPT2. We show that results from other LLMs are consistent with our findings. In Figure~\ref{fig:pth-ih-matching} and~\ref{fig:pth-ih-matching2}, we present the histograms of diagonal scores and the histograms of subspace matching scores.

%% file: src/appendix/shuffle-histogram.tex
\subsection{Histograms for shuffling experiments}\label{sec:append-shuffle-hist}

\begin{figure*}
        \centering
        \begin{subfigure}[b]{0.31\textwidth}
            \centering
            \includegraphics[width=\textwidth]{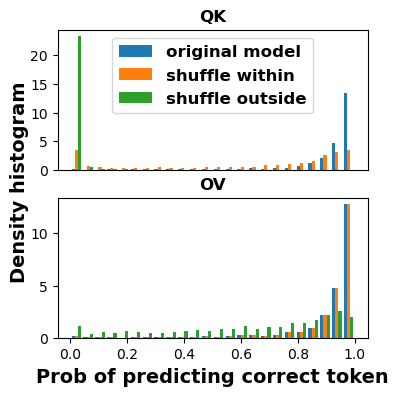}
            \caption%
            {{\small Falcon-7b}}    
        \end{subfigure}
        \hfill
        \begin{subfigure}[b]{0.31\textwidth}  
            \centering 
            \includegraphics[width=\textwidth]{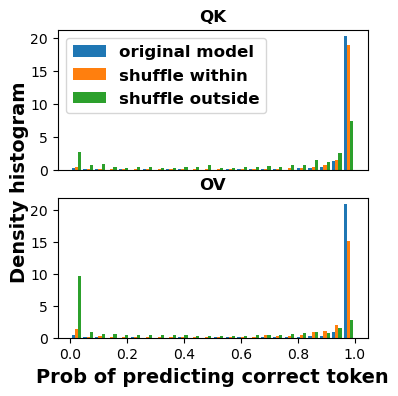}
            \caption[]%
            {{\small Gemma-7b}}    
        \end{subfigure}
        \hfill
        \begin{subfigure}[b]{0.31\textwidth}  
            \centering 
            \includegraphics[width=\textwidth]{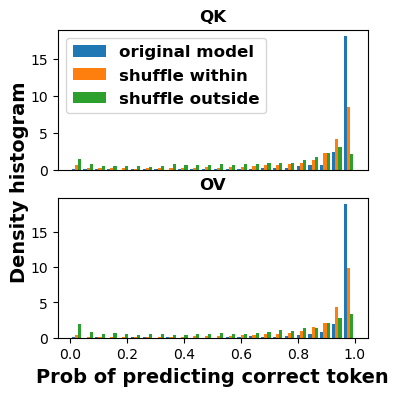}
            \caption[]%
            {{\small Gemma2-7b}}    
        \end{subfigure}
        \vskip\baselineskip
        \begin{subfigure}[b]{0.31\textwidth}  
            \centering 
            \includegraphics[width=\textwidth]{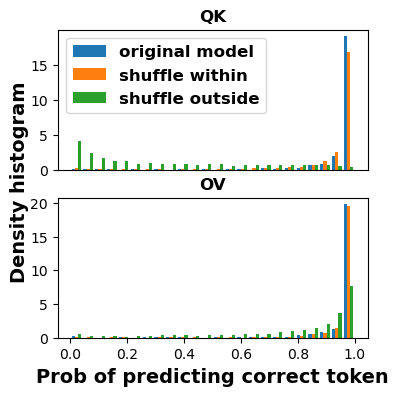}
            \caption[]%
            {{\small GPT2-XL}}    
        \end{subfigure}
        \hfill
        \begin{subfigure}[b]{0.31\textwidth}  
            \centering 
            \includegraphics[width=\textwidth]{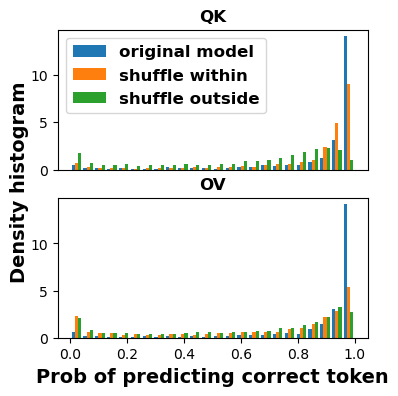}
            \caption[]%
            {{\small Llama2-7b}}    
        \end{subfigure}
        \hfill
        \begin{subfigure}[b]{0.31\textwidth}  
            \centering 
            \includegraphics[width=\textwidth]{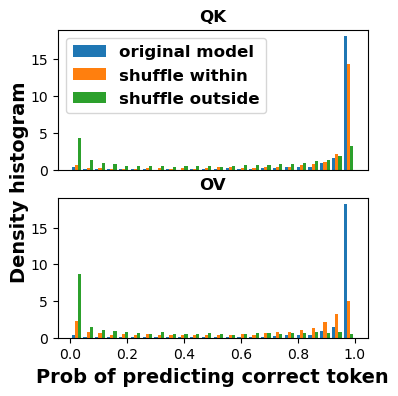}
            \caption[]%
            {{\small Llama3-8b}}    
        \end{subfigure}
        \vskip\baselineskip
        \begin{subfigure}[b]{0.31\textwidth}  
            \centering 
            \includegraphics[width=\textwidth]{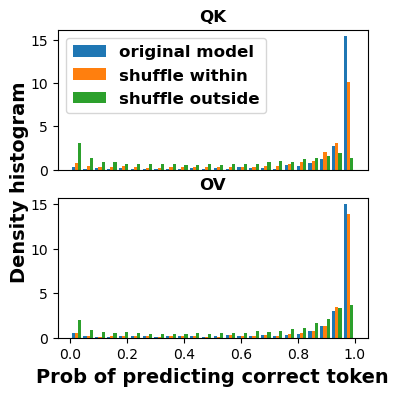}
            \caption[]%
            {{\small Mistral-7b}}    
        \end{subfigure}
        \hfill
        \begin{subfigure}[b]{0.31\textwidth}  
            \centering 
            \includegraphics[width=\textwidth]{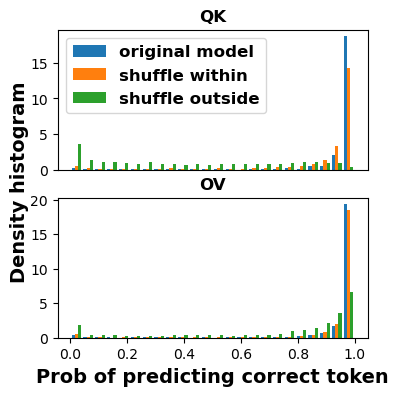}
            \caption[]%
            {{\small Olmo-7b}}    
        \end{subfigure}
        \hfill
        \begin{subfigure}[b]{0.31\textwidth}  
            \centering 
            \includegraphics[width=\textwidth]{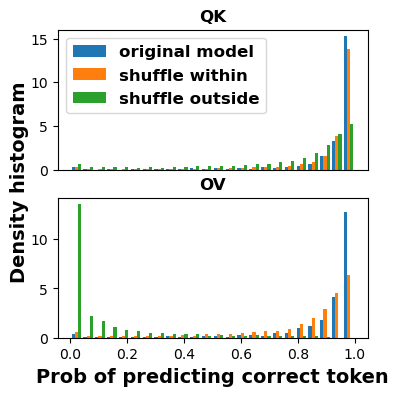}
            \caption[]%
            {{\small Pythia-7b}}    
        \end{subfigure}
        \caption{Shuffling experiments: histograms showing the effects of editing in various models.} 
        \label{fig:shuffle-hist}
    \end{figure*}

In Figure~\ref{fig:CBRH}(e), we only presented the histogram plot based on GPT2. We show in Figure~\ref{fig:shuffle-hist} that results from other LLMs are consistent with our findings.

%% file: src/appendix/projection-curve.tex
\subsection{Histograms for projection experiments}\label{sec:append-project-curve}

\begin{figure*}
        \centering
        \begin{subfigure}[b]{0.31\textwidth}
            \centering
            \includegraphics[width=\textwidth]{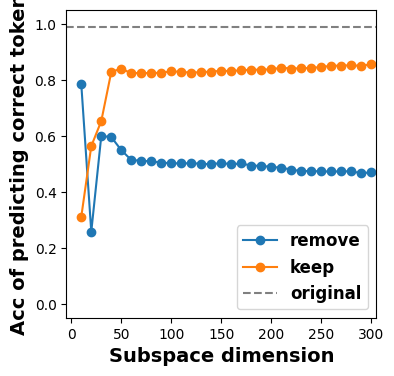}
            \caption%
            {{\small Falcon-7b (outlier)}}    
        \end{subfigure}
        \hfill
        \begin{subfigure}[b]{0.31\textwidth}  
            \centering 
            \includegraphics[width=\textwidth]{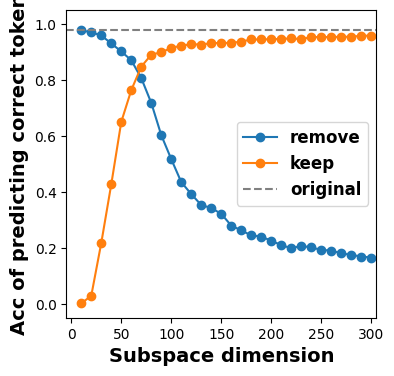}
            \caption[]%
            {{\small Gemma-7b}}    
        \end{subfigure}
        \hfill
        \begin{subfigure}[b]{0.31\textwidth}  
            \centering 
            \includegraphics[width=\textwidth]{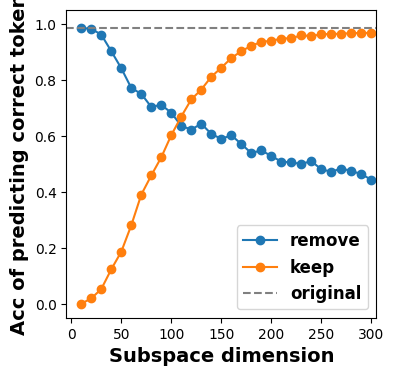}
            \caption[]%
            {{\small Gemma2-7b}}    
        \end{subfigure}
        \vskip\baselineskip
        \begin{subfigure}[b]{0.31\textwidth}  
            \centering 
            \includegraphics[width=\textwidth]{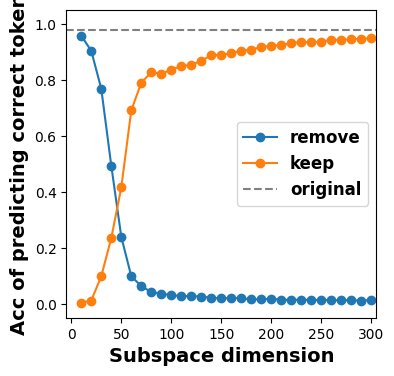}
            \caption[]%
            {{\small GPT2-XL}}    
        \end{subfigure}
        \hfill
        \begin{subfigure}[b]{0.31\textwidth}  
            \centering 
            \includegraphics[width=\textwidth]{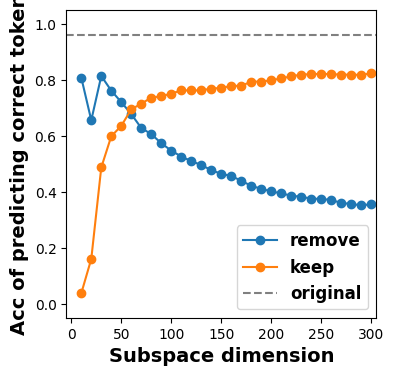}
            \caption[]%
            {{\small Llama2-7b}}    
        \end{subfigure}
        \hfill
        \begin{subfigure}[b]{0.31\textwidth}  
            \centering 
            \includegraphics[width=\textwidth]{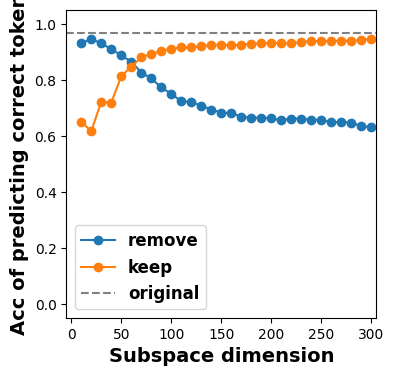}
            \caption[]%
            {{\small Llama3-8b}}    
        \end{subfigure}
        \vskip\baselineskip
        \begin{subfigure}[b]{0.31\textwidth}  
            \centering 
            \includegraphics[width=\textwidth]{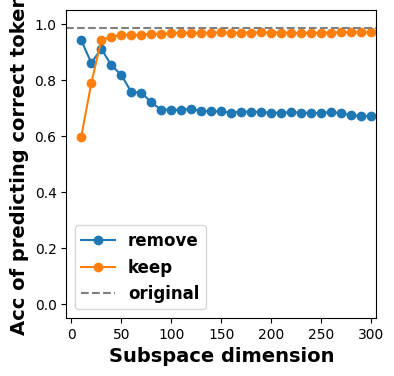}
            \caption[]%
            {{\small Mistral-7b}}    
        \end{subfigure}
        \hfill
        \begin{subfigure}[b]{0.31\textwidth}  
            \centering 
            \includegraphics[width=\textwidth]{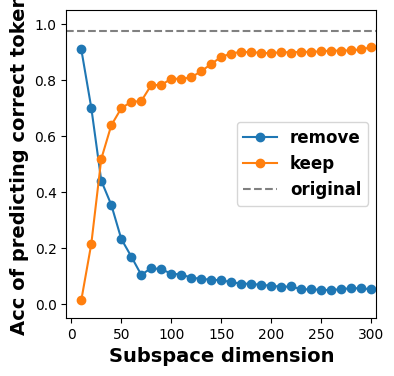}
            \caption[]%
            {{\small Olmo-7b}}    
        \end{subfigure}
        \hfill
        \begin{subfigure}[b]{0.31\textwidth}  
            \centering 
            \includegraphics[width=\textwidth]{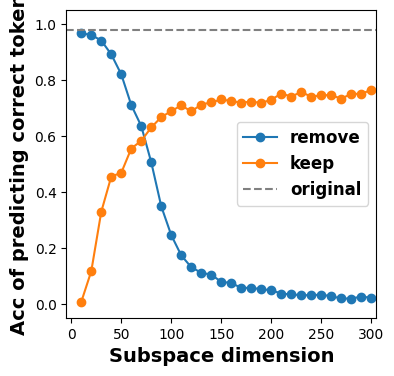}
            \caption[]%
            {{\small Pythia-7b}}    
        \end{subfigure}
        \caption{Projection experiments: histograms showing the effects of editing in various models.} 
        \label{fig:project-curve}
    \end{figure*}

In Figure~\ref{fig:CBRH}(f), we only presented the histogram plots based on GPT2. We show in Figure~\ref{fig:project-curve} that results from other LLMs are similar to the plot from GPT2.

%% file: src/appendix/add-cbrh.tex
\begin{figure*}
\centering
\begin{subfigure}[b]{0.31\textwidth}
\centering
\includegraphics[width=\textwidth]{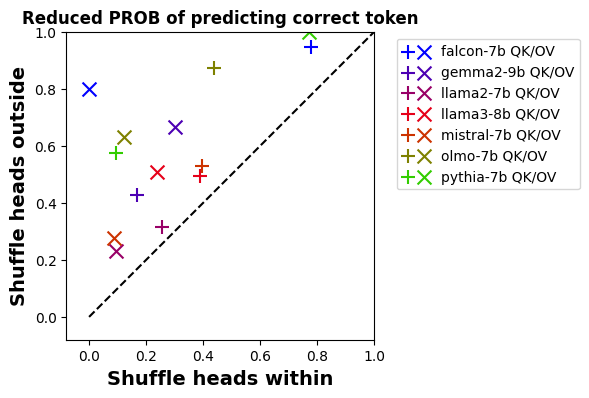}
\caption{Shuffle experiment: reduced probability on Fuzzy Copying task.}\label{fig:fuzzycopy-shuffle1}
\end{subfigure}
\hfill
\begin{subfigure}[b]{0.31\textwidth}
\centering
\includegraphics[width=\textwidth]{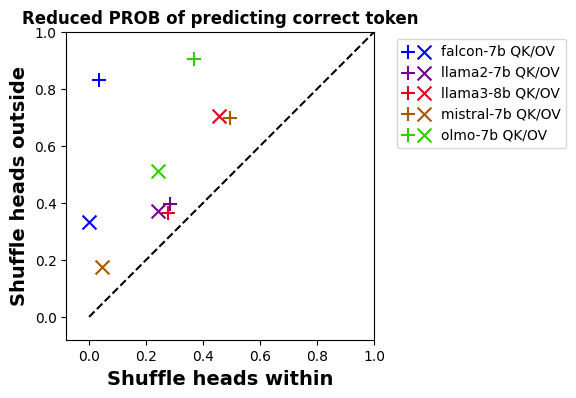}
\caption{Shuffle experiment: reduced probability on IOI task.}\label{fig:ioi-shuffle}
\end{subfigure}
\hfill
\begin{subfigure}[b]{0.31\textwidth}
\centering
\includegraphics[width=\textwidth]{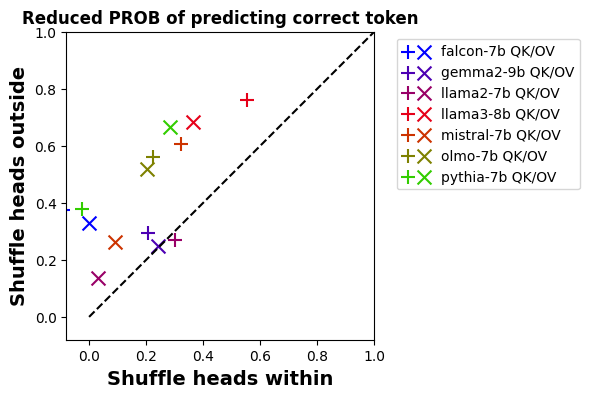}
\caption{Shuffle experiment: reduced probability on ICL task.}\label{fig:icl-shuffle}
\end{subfigure}

\vskip\baselineskip 

\begin{subfigure}[b]{0.31\textwidth}
\centering
\includegraphics[width=\textwidth]{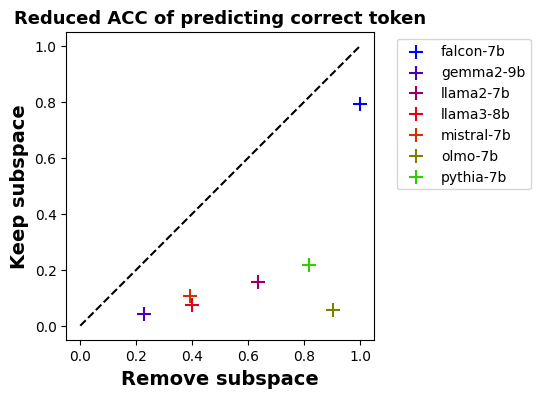}
\caption{Projection experiment: reduced accuracy on IOI task.}\label{fig:fuzzycopy-project}
\end{subfigure}
\hfill
\begin{subfigure}[b]{0.31\textwidth}
\centering
\includegraphics[width=\textwidth]{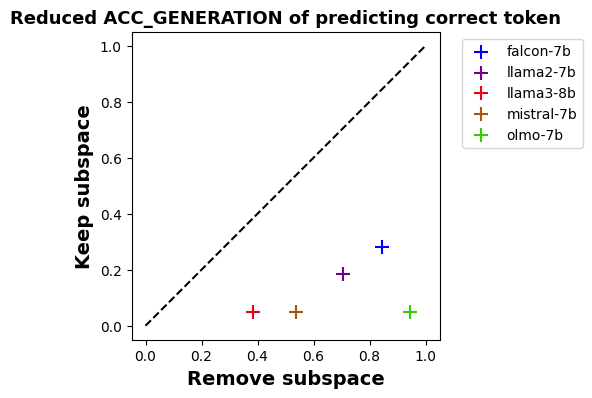}
\caption{Projection experiment: reduced accuracy on IOI task.}\label{fig:ioi-project}
\end{subfigure}
\hfill
\begin{subfigure}[b]{0.31\textwidth}
\centering
\includegraphics[width=\textwidth]{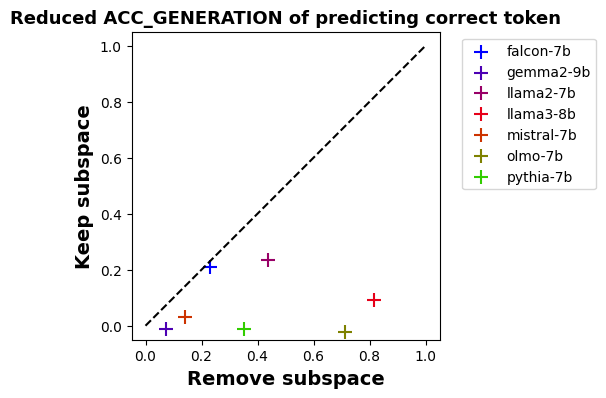}
\caption{Projection experiment: reduced accuracy on ICL task.}\label{fig:icl-project}
\end{subfigure}

\caption{A set of shuffle experiments: reduced probability on the Fuzzy Copying task.}
\label{fig:append-add-cbrh}
\end{figure*}

In Figure 7(e), 7(f), we only report the results of shuffle and projection intervention experiments on the Copying task. For the other three tasks, Fuzzy Copying, IOI, and ICL, we follow the measurement in \ref{sec:append-task-details} and the intervention method in \ref{sec:append-shuffle}, \ref{sec:append-project}. We show in in Figure \ref{fig:append-add-cbrh} the results of experiments. The results are similar to those presented in Figure 7.







%% file: src/appendix/scaling.tex
\paragraph{Model details.}

We list the models in the experiments. 
\begin{itemize}
    \item \textbf{Pythia-36M} 6 layers, 8 attention heads, and a hidden size of 256.
    \item \textbf{Pythia-70M} 6 layers, 8 attention heads, and a hidden size of 512.
    \item \textbf{Pythia-160M} 12 layers, 12 attention heads, and a hidden size of 768.
    \item \textbf{Pythia-410M} 24 layers, 16 attention heads, and a hidden size of 1024.
    \item \textbf{Pythia-1B} 16 layers, 8 attention heads, and a hidden size of 2048.
    \item \textbf{Pythia-1.4B} 24 layers, 16 attention heads, and a hidden size of 2048.
    \item \textbf{Pythia-2.8B} 32 layers, 32 attention heads, and a hidden size of 2560.
    \item \textbf{Pythia-7B} 32 layers, 32 attention heads, and a hidden size of 4096.
\end{itemize}

%% file: main.bbl
\begin{thebibliography}{10}

\bibitem{abbe2023generalization}
Emmanuel Abbe, Samy Bengio, Aryo Lotfi, and Kevin Rizk.
\newblock Generalization on the unseen, logic reasoning and degree curriculum.
\newblock In {\em International Conference on Machine Learning}, pages 31--60. PMLR, 2023.

\bibitem{agarwal2024manyshot}
Rishabh Agarwal, Avi Singh, Lei~M Zhang, Bernd Bohnet, Luis Rosias, Stephanie~C.Y. Chan, Biao Zhang, Aleksandra Faust, and Hugo Larochelle.
\newblock Many-shot in-context learning.
\newblock In {\em ICML 2024 Workshop on In-Context Learning}, 2024.

\bibitem{Akyrek2024InContextLL}
Ekin Aky{\"u}rek, Bailin Wang, Yoon Kim, and Jacob Andreas.
\newblock In-context language learning: Architectures and algorithms.
\newblock {\em ArXiv}, abs/2401.12973, 2024.

\bibitem{akyurek2024context}
Ekin Aky{\"u}rek, Bailin Wang, Yoon Kim, and Jacob Andreas.
\newblock In-context language learning: Arhitectures and algorithms.
\newblock {\em arXiv preprint arXiv:2401.12973}, 2024.

\bibitem{almazrouei2023falcon}
Ebtesam Almazrouei, Hamza Alobeidli, Abdulaziz Alshamsi, Alessandro Cappelli, Ruxandra Cojocaru, M{\'e}rouane Debbah, {\'E}tienne Goffinet, Daniel Hesslow, Julien Launay, Quentin Malartic, et~al.
\newblock The falcon series of open language models.
\newblock {\em arXiv preprint arXiv:2311.16867}, 2023.

\bibitem{anil2022exploring}
Cem Anil, Yuhuai Wu, Anders Andreassen, Aitor Lewkowycz, Vedant Misra, Vinay Ramasesh, Ambrose Slone, Guy Gur-Ari, Ethan Dyer, and Behnam Neyshabur.
\newblock Exploring length generalization in large language models.
\newblock {\em Advances in Neural Information Processing Systems}, 35:38546--38556, 2022.

\bibitem{anwar2024foundational}
Usman Anwar, Abulhair Saparov, Javier Rando, Daniel Paleka, Miles Turpin, Peter Hase, Ekdeep~Singh Lubana, Erik Jenner, Stephen Casper, Oliver Sourbut, et~al.
\newblock Foundational challenges in assuring alignment and safety of large language models.
\newblock {\em arXiv preprint arXiv:2404.09932}, 2024.

\bibitem{arora2023theory}
Sanjeev Arora and Anirudh Goyal.
\newblock A theory for emergence of complex skills in language models.
\newblock {\em arXiv preprint arXiv:2307.15936}, 2023.

\bibitem{arora-etal-2018-linear}
Sanjeev Arora, Yuanzhi Li, Yingyu Liang, Tengyu Ma, and Andrej Risteski.
\newblock Linear algebraic structure of word senses, with applications to polysemy.
\newblock {\em Transactions of the Association for Computational Linguistics}, 6:483--495, 2018.

\bibitem{bahdanau2014neural}
Dzmitry Bahdanau.
\newblock Neural machine translation by jointly learning to align and translate.
\newblock {\em arXiv preprint arXiv:1409.0473}, 2014.

\bibitem{bai2024transformers}
Yu~Bai, Fan Chen, Huan Wang, Caiming Xiong, and Song Mei.
\newblock Transformers as statisticians: Provable in-context learning with in-context algorithm selection.
\newblock {\em Advances in neural information processing systems}, 36, 2024.

\bibitem{bengio2024machine}
Yoshua Bengio and Nikolay Malkin.
\newblock Machine learning and information theory concepts towards an ai mathematician.
\newblock {\em Bulletin of the American Mathematical Society}, 61(3):457--469, 2024.

\bibitem{biderman2023pythia}
Stella Biderman, Hailey Schoelkopf, Quentin~Gregory Anthony, Herbie Bradley, Kyle O’Brien, Eric Hallahan, Mohammad~Aflah Khan, Shivanshu Purohit, USVSN~Sai Prashanth, Edward Raff, et~al.
\newblock Pythia: A suite for analyzing large language models across training and scaling.
\newblock In {\em International Conference on Machine Learning}, pages 2397--2430. PMLR, 2023.

\bibitem{boix-adsera2024when}
Enric Boix-Adser{\`a}, Omid Saremi, Emmanuel Abbe, Samy Bengio, Etai Littwin, and Joshua~M. Susskind.
\newblock When can transformers reason with abstract symbols?
\newblock In {\em The Twelfth International Conference on Learning Representations}, 2024.

\bibitem{brants2007large}
Thorsten Brants, Ashok Popat, Peng Xu, Franz~Josef Och, and Jeffrey Dean.
\newblock Large language models in machine translation.
\newblock In {\em Proceedings of the 2007 Joint Conference on Empirical Methods in Natural Language Processing and Computational Natural Language Learning (EMNLP-CoNLL)}, pages 858--867, 2007.

\bibitem{brown2020language}
Tom Brown, Benjamin Mann, Nick Ryder, Melanie Subbiah, Jared~D Kaplan, Prafulla Dhariwal, Arvind Neelakantan, Pranav Shyam, Girish Sastry, Amanda Askell, et~al.
\newblock Language models are few-shot learners.
\newblock {\em Advances in neural information processing systems}, 33:1877--1901, 2020.

\bibitem{bubeck2023sparks}
S{\'e}bastien Bubeck, Varun Chandrasekaran, Ronen Eldan, Johannes Gehrke, Eric Horvitz, Ece Kamar, Peter Lee, Yin~Tat Lee, Yuanzhi Li, Scott Lundberg, et~al.
\newblock Sparks of artificial general intelligence: Early experiments with gpt-4.
\newblock {\em arXiv preprint arXiv:2303.12712}, 2023.

\bibitem{chen2024can}
Xingwu Chen and Difan Zou.
\newblock What can transformer learn with varying depth? case studies on sequence learning tasks.
\newblock {\em arXiv preprint arXiv:2404.01601}, 2024.

\bibitem{cobbe2021gsm8k}
Karl Cobbe, Vineet Kosaraju, Mohammad Bavarian, Mark Chen, Heewoo Jun, Lukasz Kaiser, Matthias Plappert, Jerry Tworek, Jacob Hilton, Reiichiro Nakano, Christopher Hesse, and John Schulman.
\newblock Training verifiers to solve math word problems.
\newblock {\em arXiv preprint arXiv:2110.14168}, 2021.

\bibitem{dong2022survey}
Qingxiu Dong, Lei Li, Damai Dai, Ce~Zheng, Zhiyong Wu, Baobao Chang, Xu~Sun, Jingjing Xu, and Zhifang Sui.
\newblock A survey for in-context learning.
\newblock {\em arXiv preprint arXiv:2301.00234}, 2022.

\bibitem{Donoho2024Data}
David Donoho.
\newblock {Data} {Science} at the {Singularity}.
\newblock {\em Harvard Data Science Review}, 6(1), jan 29 2024.
\newblock https://hdsr.mitpress.mit.edu/pub/g9mau4m0.

\bibitem{dubey2024llama}
Abhimanyu Dubey, Abhinav Jauhri, Abhinav Pandey, Abhishek Kadian, Ahmad Al-Dahle, Aiesha Letman, Akhil Mathur, Alan Schelten, Amy Yang, Angela Fan, et~al.
\newblock The llama 3 herd of models.
\newblock {\em arXiv preprint arXiv:2407.21783}, 2024.

\bibitem{dziri2024faith}
Nouha Dziri, Ximing Lu, Melanie Sclar, Xiang~Lorraine Li, Liwei Jiang, Bill~Yuchen Lin, Sean Welleck, Peter West, Chandra Bhagavatula, Ronan Le~Bras, et~al.
\newblock Faith and fate: Limits of transformers on compositionality.
\newblock {\em Advances in Neural Information Processing Systems}, 36, 2024.

\bibitem{elhage2022superposition}
Nelson Elhage, Tristan Hume, Catherine Olsson, Nicholas Schiefer, Tom Henighan, Shauna Kravec, Zac Hatfield-Dodds, Robert Lasenby, Dawn Drain, Carol Chen, Roger Grosse, Sam McCandlish, Jared Kaplan, Dario Amodei, Martin Wattenberg, and Christopher Olah.
\newblock Toy models of superposition.
\newblock {\em Transformer Circuits Thread}, 2022.
\newblock https://transformer-circuits.pub/2022/toy\_model/index.html.

\bibitem{elhage2021mathematical}
Nelson Elhage, Neel Nanda, Catherine Olsson, Tom Henighan, Nicholas Joseph, Ben Mann, Amanda Askell, Yuntao Bai, Anna Chen, Tom Conerly, Nova DasSarma, Dawn Drain, Deep Ganguli, Zac Hatfield-Dodds, Danny Hernandez, Andy Jones, Jackson Kernion, Liane Lovitt, Kamal Ndousse, Dario Amodei, Tom Brown, Jack Clark, Jared Kaplan, Sam McCandlish, and Chris Olah.
\newblock A mathematical framework for transformer circuits.
\newblock {\em Transformer Circuits Thread}, 2021.
\newblock https://transformer-circuits.pub/2021/framework/index.html.

\bibitem{ferrando2024primer}
Javier Ferrando, Gabriele Sarti, Arianna Bisazza, and Marta~R Costa-juss{\`a}.
\newblock A primer on the inner workings of transformer-based language models.
\newblock {\em arXiv preprint arXiv:2405.00208}, 2024.

\bibitem{gao2024scaling}
Leo Gao, Tom~Dupr{\'e} la~Tour, Henk Tillman, Gabriel Goh, Rajan Troll, Alec Radford, Ilya Sutskever, Jan Leike, and Jeffrey Wu.
\newblock Scaling and evaluating sparse autoencoders.
\newblock {\em arXiv preprint arXiv:2406.04093}, 2024.

\bibitem{geva2020transformer}
Mor Geva, Roei Schuster, Jonathan Berant, and Omer Levy.
\newblock Transformer feed-forward layers are key-value memories.
\newblock {\em arXiv preprint arXiv:2012.14913}, 2020.

\bibitem{gould2024successor}
Rhys Gould, Euan Ong, George Ogden, and Arthur Conmy.
\newblock Successor heads: Recurring, interpretable attention heads in the wild.
\newblock In {\em The Twelfth International Conference on Learning Representations}, 2024.

\bibitem{groeneveld2024olmo}
Dirk Groeneveld, Iz~Beltagy, Pete Walsh, Akshita Bhagia, Rodney Kinney, Oyvind Tafjord, Ananya~Harsh Jha, Hamish Ivison, Ian Magnusson, Yizhong Wang, et~al.
\newblock Olmo: Accelerating the science of language models.
\newblock {\em arXiv preprint arXiv:2402.00838}, 2024.

\bibitem{hahn2023theory}
Michael Hahn and Navin Goyal.
\newblock A theory of emergent in-context learning as implicit structure induction.
\newblock {\em arXiv preprint arXiv:2303.07971}, 2023.

\bibitem{halawi2024overthinking}
Danny Halawi, Jean-Stanislas Denain, and Jacob Steinhardt.
\newblock Overthinking the truth: Understanding how language models process false demonstrations.
\newblock In {\em The Twelfth International Conference on Learning Representations}, 2024.

\bibitem{hsu2024mechanistic}
Aliyah~R Hsu, Yeshwanth Cherapanamjeri, Anobel~Y Odisho, Peter~R Carroll, and Bin Yu.
\newblock Mechanistic interpretation through contextual decomposition in transformers.
\newblock {\em arXiv preprint arXiv:2407.00886}, 2024.

\bibitem{ilharco2022editing}
Gabriel Ilharco, Marco~Tulio Ribeiro, Mitchell Wortsman, Suchin Gururangan, Ludwig Schmidt, Hannaneh Hajishirzi, and Ali Farhadi.
\newblock Editing models with task arithmetic.
\newblock {\em arXiv preprint arXiv:2212.04089}, 2022.

\bibitem{jiang2023mistral}
Albert~Q Jiang, Alexandre Sablayrolles, Arthur Mensch, Chris Bamford, Devendra~Singh Chaplot, Diego de~las Casas, Florian Bressand, Gianna Lengyel, Guillaume Lample, Lucile Saulnier, et~al.
\newblock Mistral 7b.
\newblock {\em arXiv preprint arXiv:2310.06825}, 2023.

\bibitem{spiked}
Iain~M. Johnstone.
\newblock On the distribution of the largest eigenvalue in principal components analysis.
\newblock {\em The Annals of Statistics}, 29(2):295--327, 2001.

\bibitem{kazemnejad2023the}
Amirhossein Kazemnejad, Inkit Padhi, Karthikeyan Natesan, Payel Das, and Siva Reddy.
\newblock The impact of positional encoding on length generalization in transformers.
\newblock In {\em Thirty-seventh Conference on Neural Information Processing Systems}, 2023.

\bibitem{kazemnejad2024impact}
Amirhossein Kazemnejad, Inkit Padhi, Karthikeyan Natesan~Ramamurthy, Payel Das, and Siva Reddy.
\newblock The impact of positional encoding on length generalization in transformers.
\newblock {\em Advances in Neural Information Processing Systems}, 36, 2024.

\bibitem{kojima2022large}
Takeshi Kojima, Shixiang~Shane Gu, Machel Reid, Yutaka Matsuo, and Yusuke Iwasawa.
\newblock Large language models are zero-shot reasoners.
\newblock {\em Advances in neural information processing systems}, 35:22199--22213, 2022.

\bibitem{kudo2023deep}
Keito Kudo, Yoichi Aoki, Tatsuki Kuribayashi, Ana Brassard, Masashi Yoshikawa, Keisuke Sakaguchi, and Kentaro Inui.
\newblock Do deep neural networks capture compositionality in arithmetic reasoning?
\newblock {\em arXiv preprint arXiv:2302.07866}, 2023.

\bibitem{lee2023teaching}
Nayoung Lee, Kartik Sreenivasan, Jason~D Lee, Kangwook Lee, and Dimitris Papailiopoulos.
\newblock Teaching arithmetic to small transformers.
\newblock {\em arXiv preprint arXiv:2307.03381}, 2023.

\bibitem{li2024inference}
Kenneth Li, Oam Patel, Fernanda Vi{\'e}gas, Hanspeter Pfister, and Martin Wattenberg.
\newblock Inference-time intervention: Eliciting truthful answers from a language model.
\newblock {\em Advances in Neural Information Processing Systems}, 36, 2024.

\bibitem{liang2023holistic}
Percy Liang, Rishi Bommasani, Tony Lee, Dimitris Tsipras, Dilara Soylu, Michihiro Yasunaga, Yian Zhang, Deepak Narayanan, Yuhuai Wu, Ananya Kumar, Benjamin Newman, Binhang Yuan, Bobby Yan, Ce~Zhang, Christian~Alexander Cosgrove, Christopher~D Manning, Christopher Re, Diana Acosta-Navas, Drew~Arad Hudson, Eric Zelikman, Esin Durmus, Faisal Ladhak, Frieda Rong, Hongyu Ren, Huaxiu Yao, Jue WANG, Keshav Santhanam, Laurel Orr, Lucia Zheng, Mert Yuksekgonul, Mirac Suzgun, Nathan Kim, Neel Guha, Niladri~S. Chatterji, Omar Khattab, Peter Henderson, Qian Huang, Ryan~Andrew Chi, Sang~Michael Xie, Shibani Santurkar, Surya Ganguli, Tatsunori Hashimoto, Thomas Icard, Tianyi Zhang, Vishrav Chaudhary, William Wang, Xuechen Li, Yifan Mai, Yuhui Zhang, and Yuta Koreeda.
\newblock Holistic evaluation of language models.
\newblock {\em Transactions on Machine Learning Research}, 2023.
\newblock Featured Certification, Expert Certification.

\bibitem{liu2023omnigrok}
Ziming Liu, Eric~J Michaud, and Max Tegmark.
\newblock Omnigrok: Grokking beyond algorithmic data.
\newblock In {\em The Eleventh International Conference on Learning Representations}, 2023.

\bibitem{lyu2023dichotomy}
Kaifeng Lyu, Jikai Jin, Zhiyuan Li, Simon~Shaolei Du, Jason~D Lee, and Wei Hu.
\newblock Dichotomy of early and late phase implicit biases can provably induce grokking.
\newblock In {\em The Twelfth International Conference on Learning Representations}, 2023.

\bibitem{malartic2024falcon2}
Quentin Malartic, Nilabhra~Roy Chowdhury, Ruxandra Cojocaru, Mugariya Farooq, Giulia Campesan, Yasser Abdelaziz~Dahou Djilali, Sanath Narayan, Ankit Singh, Maksim Velikanov, Basma El~Amel Boussaha, et~al.
\newblock Falcon2-11b technical report.
\newblock {\em arXiv preprint arXiv:2407.14885}, 2024.

\bibitem{mallinar2024emergence}
Neil Mallinar, Daniel Beaglehole, Libin Zhu, Adityanarayanan Radhakrishnan, Parthe Pandit, and Mikhail Belkin.
\newblock Emergence in non-neural models: grokking modular arithmetic via average gradient outer product.
\newblock {\em arXiv preprint arXiv:2407.20199}, 2024.

\bibitem{merullo2023circuit}
Jack Merullo, Carsten Eickhoff, and Ellie Pavlick.
\newblock Circuit component reuse across tasks in transformer language models.
\newblock {\em arXiv preprint arXiv:2310.08744}, 2023.

\bibitem{Mikolov2013EfficientEO}
Tomas Mikolov, Kai Chen, Gregory~S. Corrado, and Jeffrey Dean.
\newblock Efficient estimation of word representations in vector space.
\newblock In {\em International Conference on Learning Representations}, 2013.

\bibitem{mikolov2013linguistic}
Tom{\'a}{\v{s}} Mikolov, Wen-tau Yih, and Geoffrey Zweig.
\newblock Linguistic regularities in continuous space word representations.
\newblock In {\em Proceedings of the 2013 conference of the north american chapter of the association for computational linguistics: Human language technologies}, pages 746--751, 2013.

\bibitem{mirzadeh2024gsm}
Iman Mirzadeh, Keivan Alizadeh, Hooman Shahrokhi, Oncel Tuzel, Samy Bengio, and Mehrdad Farajtabar.
\newblock Gsm-symbolic: Understanding the limitations of mathematical reasoning in large language models.
\newblock {\em arXiv preprint arXiv:2410.05229}, 2024.

\bibitem{nanda2023progress}
Neel Nanda, Lawrence Chan, Tom Lieberum, Jess Smith, and Jacob Steinhardt.
\newblock Progress measures for grokking via mechanistic interpretability.
\newblock {\em arXiv preprint arXiv:2301.05217}, 2023.

\bibitem{nichani2024transformers}
Eshaan Nichani, Alex Damian, and Jason~D Lee.
\newblock How transformers learn causal structure with gradient descent.
\newblock {\em arXiv preprint arXiv:2402.14735}, 2024.

\bibitem{olah2020zoom}
Chris Olah, Nick Cammarata, Ludwig Schubert, Gabriel Goh, Michael Petrov, and Shan Carter.
\newblock Zoom in: An introduction to circuits.
\newblock {\em Distill}, 5(3):e00024--001, 2020.

\bibitem{inductionhead22}
Catherine Olsson, Nelson Elhage, Neel Nanda, Nicholas Joseph, Nova DasSarma, Tom Henighan, Ben Mann, Amanda Askell, Yuntao Bai, Anna Chen, Tom Conerly, Dawn Drain, Deep Ganguli, Zac Hatfield-Dodds, Danny Hernandez, Scott Johnston, Andy Jones, Jackson Kernion, Liane Lovitt, Kamal Ndousse, Dario Amodei, Tom Brown, Jack Clark, Jared Kaplan, Sam McCandlish, and Chris Olah.
\newblock In-context learning and induction heads.
\newblock {\em Transformer Circuits Thread}, 2022.
\newblock https://transformer-circuits.pub/2022/in-context-learning-and-induction-heads/index.html.

\bibitem{pan-etal-2023-context}
Jane Pan, Tianyu Gao, Howard Chen, and Danqi Chen.
\newblock What in-context learning {``}learns{''} in-context: Disentangling task recognition and task learning.
\newblock In Anna Rogers, Jordan Boyd-Graber, and Naoaki Okazaki, editors, {\em Findings of the Association for Computational Linguistics: ACL 2023}, pages 8298--8319, Toronto, Canada, July 2023. Association for Computational Linguistics.

\bibitem{Park2023TheLR}
Kiho Park, Yo~Joong Choe, and Victor Veitch.
\newblock The linear representation hypothesis and the geometry of large language models.
\newblock {\em ArXiv preprint arXiv:2311.03658}, 2023.

\bibitem{pennington-etal-2014-glove}
Jeffrey Pennington, Richard Socher, and Christopher Manning.
\newblock {G}lo{V}e: Global vectors for word representation.
\newblock In Alessandro Moschitti, Bo~Pang, and Walter Daelemans, editors, {\em Proceedings of the 2014 Conference on Empirical Methods in Natural Language Processing ({EMNLP})}, pages 1532--1543, Doha, Qatar, October 2014. Association for Computational Linguistics.

\bibitem{pennington2014glove}
Jeffrey Pennington, Richard Socher, and Christopher~D Manning.
\newblock Glove: Global vectors for word representation.
\newblock In {\em Proceedings of the 2014 conference on empirical methods in natural language processing (EMNLP)}, pages 1532--1543, 2014.

\bibitem{power2022grokking}
Alethea Power, Yuri Burda, Harri Edwards, Igor Babuschkin, and Vedant Misra.
\newblock Grokking: Generalization beyond overfitting on small algorithmic datasets.
\newblock {\em arXiv preprint arXiv:2201.02177}, 2022.

\bibitem{reddy2023mechanistic}
Gautam Reddy.
\newblock The mechanistic basis of data dependence and abrupt learning in an in-context classification task.
\newblock {\em arXiv preprint arXiv:2312.03002}, 2023.

\bibitem{reizinger2024understanding}
Patrik Reizinger, Szilvia Ujv{\'a}ry, Anna M{\'e}sz{\'a}ros, Anna Kerekes, Wieland Brendel, and Ferenc Husz{\'a}r.
\newblock Understanding llms requires more than statistical generalization.
\newblock {\em arXiv preprint arXiv:2405.01964}, 2024.

\bibitem{reynolds2021prompt}
Laria Reynolds and Kyle McDonell.
\newblock Prompt programming for large language models: Beyond the few-shot paradigm.
\newblock In {\em Extended Abstracts of the 2021 CHI Conference on Human Factors in Computing Systems}, pages 1--7, 2021.

\bibitem{rong2021extrapolating}
Frieda Rong.
\newblock Extrapolating to unnatural language processing with gpt-3’s in-context learning: The good, the bad, and the mysterious, 2021.

\bibitem{ood-language-23}
Abulhair Saparov, Richard~Yuanzhe Pang, Vishakh Padmakumar, Nitish Joshi, Mehran Kazemi, Najoung Kim, and He~He.
\newblock Testing the general deductive reasoning capacity of large language models using ood examples.
\newblock {\em Advances in Neural Information Processing Systems}, 36, 2024.

\bibitem{saparov2024testing}
Abulhair Saparov, Richard~Yuanzhe Pang, Vishakh Padmakumar, Nitish Joshi, Mehran Kazemi, Najoung Kim, and He~He.
\newblock Testing the general deductive reasoning capacity of large language models using ood examples.
\newblock {\em Advances in Neural Information Processing Systems}, 36, 2024.

\bibitem{shi2024why}
Zhenmei Shi, Junyi Wei, Zhuoyan Xu, and Yingyu Liang.
\newblock Why larger language models do in-context learning differently?
\newblock In {\em Forty-first International Conference on Machine Learning}, 2024.

\bibitem{singh2024needs}
Aaditya~K Singh, Ted Moskovitz, Felix Hill, Stephanie~CY Chan, and Andrew~M Saxe.
\newblock What needs to go right for an induction head? a mechanistic study of in-context learning circuits and their formation.
\newblock {\em arXiv preprint arXiv:2404.07129}, 2024.

\bibitem{song2023uncovering}
Jiajun Song and Yiqiao Zhong.
\newblock Uncovering hidden geometry in transformers via disentangling position and context.
\newblock {\em arXiv preprint arXiv:2310.04861}, 2023.

\bibitem{su2024roformer}
Jianlin Su, Murtadha Ahmed, Yu~Lu, Shengfeng Pan, Wen Bo, and Yunfeng Liu.
\newblock Roformer: Enhanced transformer with rotary position embedding.
\newblock {\em Neurocomputing}, 568:127063, 2024.

\bibitem{tang2023large}
Xiaojuan Tang, Zilong Zheng, Jiaqi Li, Fanxu Meng, Song-Chun Zhu, Yitao Liang, and Muhan Zhang.
\newblock Large language models are in-context semantic reasoners rather than symbolic reasoners.
\newblock {\em arXiv preprint arXiv:2305.14825}, 2023.

\bibitem{team2024gemma}
Gemma Team, Thomas Mesnard, Cassidy Hardin, Robert Dadashi, Surya Bhupatiraju, Shreya Pathak, Laurent Sifre, Morgane Rivi{\`e}re, Mihir~Sanjay Kale, Juliette Love, et~al.
\newblock Gemma: Open models based on gemini research and technology.
\newblock {\em arXiv preprint arXiv:2403.08295}, 2024.

\bibitem{team2024gemma2}
Gemma Team, Morgane Riviere, Shreya Pathak, Pier~Giuseppe Sessa, Cassidy Hardin, Surya Bhupatiraju, L{\'e}onard Hussenot, Thomas Mesnard, Bobak Shahriari, Alexandre Ram{\'e}, et~al.
\newblock Gemma 2: Improving open language models at a practical size.
\newblock {\em arXiv preprint arXiv:2408.00118}, 2024.

\bibitem{templeton2024scaling}
A~Templeton, T~Conerly, J~Marcus, J~Lindsey, T~Bricken, B~Chen, A~Pearce, C~Citro, E~Ameisen, A~Jones, et~al.
\newblock Scaling monosemanticity: Extracting interpretable features from claude 3 sonnet.
\newblock {\em Transformer Circuits Thread}, 2024.

\bibitem{touvron2023llama}
Hugo Touvron, Louis Martin, Kevin Stone, Peter Albert, Amjad Almahairi, Yasmine Babaei, Nikolay Bashlykov, Soumya Batra, Prajjwal Bhargava, Shruti Bhosale, et~al.
\newblock Llama 2: Open foundation and fine-tuned chat models.
\newblock {\em arXiv preprint arXiv:2307.09288}, 2023.

\bibitem{vaswani2017attention}
Ashish Vaswani, Noam Shazeer, Niki Parmar, Jakob Uszkoreit, Llion Jones, Aidan~N Gomez, {\L}ukasz Kaiser, and Illia Polosukhin.
\newblock Attention is all you need.
\newblock {\em Advances in neural information processing systems}, 30, 2017.

\bibitem{vinyals2015show}
Oriol Vinyals, Alexander Toshev, Samy Bengio, and Dumitru Erhan.
\newblock Show and tell: A neural image caption generator.
\newblock In {\em Proceedings of the IEEE conference on computer vision and pattern recognition}, pages 3156--3164, 2015.

\bibitem{wang2024grokked}
Boshi Wang, Xiang Yue, Yu~Su, and Huan Sun.
\newblock Grokked transformers are implicit reasoners: A mechanistic journey to the edge of generalization.
\newblock {\em arXiv preprint arXiv:2405.15071}, 2024.

\bibitem{Grokked}
Boshi Wang, Xiang Yue, Yu~Su, and Huan Sun.
\newblock Grokked transformers are implicit reasoners: A mechanistic journey to the edge of generalization.
\newblock {\em arXiv preprint arXiv:2405.15071}, 2024.

\bibitem{wang2023interpretability}
Kevin~Ro Wang, Alexandre Variengien, Arthur Conmy, Buck Shlegeris, and Jacob Steinhardt.
\newblock Interpretability in the wild: a circuit for indirect object identification in {GPT}-2 small.
\newblock In {\em The Eleventh International Conference on Learning Representations}, 2023.

\bibitem{2405-15302}
Zhiwei Wang, Yunji Wang, Zhongwang Zhang, Zhangchen Zhou, Hui Jin, Tianyang Hu, Jiacheng Sun, Zhenguo Li, Yaoyu Zhang, and Zhi-Qin~John Xu.
\newblock The buffer mechanism for multi-step information reasoning in language models, 2024.

\bibitem{wei2022emergent}
Jason Wei, Yi~Tay, Rishi Bommasani, Colin Raffel, Barret Zoph, Sebastian Borgeaud, Dani Yogatama, Maarten Bosma, Denny Zhou, Donald Metzler, et~al.
\newblock Emergent abilities of large language models.
\newblock {\em arXiv preprint arXiv:2206.07682}, 2022.

\bibitem{wei2022chain}
Jason Wei, Xuezhi Wang, Dale Schuurmans, Maarten Bosma, Fei Xia, Ed~Chi, Quoc~V Le, Denny Zhou, et~al.
\newblock Chain-of-thought prompting elicits reasoning in large language models.
\newblock {\em Advances in neural information processing systems}, 35:24824--24837, 2022.

\bibitem{wei2024larger}
Jerry Wei, Jason Wei, Yi~Tay, Dustin Tran, Albert Webson, Yifeng Lu, Xinyun Chen, Hanxiao Liu, Da~Huang, Denny Zhou, and Tengyu Ma.
\newblock Larger language models do in-context learning differently, 2024.

\bibitem{wu2023reasoning}
Zhaofeng Wu, Linlu Qiu, Alexis Ross, Ekin Aky{\"u}rek, Boyuan Chen, Bailin Wang, Najoung Kim, Jacob Andreas, and Yoon Kim.
\newblock Reasoning or reciting? exploring the capabilities and limitations of language models through counterfactual tasks.
\newblock {\em arXiv preprint arXiv:2307.02477}, 2023.

\bibitem{wu-etal-2024-reasoning}
Zhaofeng Wu, Linlu Qiu, Alexis Ross, Ekin Aky{\"u}rek, Boyuan Chen, Bailin Wang, Najoung Kim, Jacob Andreas, and Yoon Kim.
\newblock Reasoning or reciting? exploring the capabilities and limitations of language models through counterfactual tasks.
\newblock In Kevin Duh, Helena Gomez, and Steven Bethard, editors, {\em Proceedings of the 2024 Conference of the North American Chapter of the Association for Computational Linguistics: Human Language Technologies (Volume 1: Long Papers)}, pages 1819--1862, Mexico City, Mexico, June 2024. Association for Computational Linguistics.

\bibitem{xu2024do}
Zhuoyan Xu, Zhenmei Shi, and Yingyu Liang.
\newblock Do large language models have compositional ability? an investigation into limitations and scalability.
\newblock In {\em ICLR 2024 Workshop on Mathematical and Empirical Understanding of Foundation Models}, 2024.

\bibitem{yun2021transformer}
Zeyu Yun, Yubei Chen, Bruno~A Olshausen, and Yann LeCun.
\newblock Transformer visualization via dictionary learning: contextualized embedding as a linear superposition of transformer factors.
\newblock {\em arXiv preprint arXiv:2103.15949}, 2021.

\bibitem{zhang2024trained}
Ruiqi Zhang, Spencer Frei, and Peter~L Bartlett.
\newblock Trained transformers learn linear models in-context.
\newblock {\em Journal of Machine Learning Research}, 25(49):1--55, 2024.

\bibitem{zhang2022unveiling}
Yi~Zhang, Arturs Backurs, S{\'e}bastien Bubeck, Ronen Eldan, Suriya Gunasekar, and Tal Wagner.
\newblock Unveiling transformers with lego: a synthetic reasoning task.
\newblock {\em arXiv preprint arXiv:2206.04301}, 2022.

\bibitem{zhong2024clock}
Ziqian Zhong, Ziming Liu, Max Tegmark, and Jacob Andreas.
\newblock The clock and the pizza: Two stories in mechanistic explanation of neural networks.
\newblock {\em Advances in Neural Information Processing Systems}, 36, 2024.

\bibitem{zhou2023algorithms}
Hattie Zhou, Arwen Bradley, Etai Littwin, Noam Razin, Omid Saremi, Josh Susskind, Samy Bengio, and Preetum Nakkiran.
\newblock What algorithms can transformers learn? a study in length generalization.
\newblock {\em arXiv preprint arXiv:2310.16028}, 2023.

\end{thebibliography}
